\documentclass[twoside]{article}

\usepackage[accepted]{aistats2024}
\usepackage{apalike}
\usepackage{graphicx}
\usepackage{xcolor}
\usepackage{macro}
\usepackage{wrapfig}
\usepackage[T1]{fontenc}

\newcommand{\instfede}{UMMISCO, IRD, Sorbonne Université}
\newcommand{\instmarco}{New York University}
\newcommand{\instpablo}{International Laboratory on Learning Systems (ILLS), Quebec AI Institute (MILA)\\CNRS, CentraleSupélec - Université Paris-Saclay}

\makeatletter
\def\blfootnote{\xdef\@thefnmark{}\@footnotetext}
\makeatother
\begin{document}

\twocolumn[

\aistatstitle{Optimal Zero-Shot Detector for Multi-Armed Attacks}

\aistatsauthor{ Federica Granese$^*$ \And Marco Romanelli$^*$}

\aistatsaddress{ \instfede \And \instmarco }
\aistatsauthor{ Pablo Piantanida }
\aistatsaddress{ \instpablo }

]

\begin{abstract}
This paper explores a scenario in which a malicious actor employs a multi-armed attack strategy to manipulate data samples, offering them various avenues to introduce noise into the dataset. Our central objective is to protect the data by detecting any alterations to the input. We approach this defensive strategy with utmost caution, operating in an environment where the defender possesses significantly less information compared to the attacker. Specifically, the defender is unable to utilize any data samples for training a defense model or verifying the integrity of the channel. Instead, the defender relies exclusively on a set of pre-existing detectors readily available ``off the shelf''. To tackle this challenge, we derive an innovative information-theoretic defense approach that optimally aggregates the decisions made by these detectors, eliminating the need for any training data. We further explore a practical use-case scenario for empirical evaluation, where the attacker possesses a pre-trained classifier and launches well-known adversarial attacks against it. Our experiments highlight the effectiveness of our proposed solution, even in scenarios that deviate from the optimal setup.


\end{abstract}

\section{INTRODUCTION}
\label{sec:intro}
Defending signal communication from attackers is a fundamental problem in information theory~\cite{KarlofW2003IEEEIWSSNPA,PerrigSW2004ACMComm}. Notably, some attacks are aimed at the physical layer of the communication channel, which is responsible for transmitting the signal. 
The goal of such attacks is to generate a denial of service (DoS), 
which involves disrupting legitimate
communication by causing intentional malfunction of the communication channel~\cite{GroverLY2014IJAHUC}.
In a typical input perturbation scenario, a malicious actor is allowed to detect and alter the signal before it reaches the communication channel~\cite{SadeghiL2019IEEEWCL,TianZSML2022DCN}. The interest in such attacks has been exacerbated by the growing popularity of machine learning (ML) models, which are known to be vulnerable to adversarial attacks~\cite{goodfellowSS2014ICLR}.

We consider a scenario in which the attacker cannot change the internal parameters of the channel, i.e. it cannot modify the way signals are transmitted. However, it possesses two key advantages: \textit{i)} it has white-box access to the channel, and \textit{ii)} it can benefit from a \textit{multi-armed attack} scheme. The former implies that the attacker has full knowledge of the channel's internal parameters, while the latter means that the attacker can mount multiple attacks at the same time, choosing from a set of available attack strategies~\cite{GranesePRMP2022ECMLPKDD}.
\blfootnote{$^*$Equal contribution.}
From the defense standpoint, we consider a zero-shot detection framework, and we propose an optimal way to aggregate the individual decisions of multiple ``off-the-shelf'' detectors using a minimax approach. Our solution is a zero-shot one since it does not require training samples. Moreover, it is optimal in the sense that, assuming the availability of one effective\footnote{A detector is effective if and only if it is able to distinguish with high accuracy between clean inputs and corrupted inputs generated according to at least one of the possible strategies available to the multi-armed attacker.} detector for each attack strategy available in the multi-arm scenario, it aggregates the detectors' decisions in a way that minimizes the success of the strongest multi-armed attacker. As we shall see, our proposed solution is optimal within the framework described in \cref{sec:math_framework}, i.e., when we want to minimize the average worst-case regret \cite{BarronRY1998TInfT} of detecting adversarial examples when the attacker can mount multi-armed attacks.
Interestingly, when the aforementioned assumption for the optimum is not met, we provide an upper bound on the detection error of our solution, which is linked to a notion of ``statistical distance'' between the attack domain known at the level of the defender, and the new attack domain that may arise at evaluation time and is unknown to the defender.
Our proposed solution is highly flexible, allowing for the aggregation of any existing or future supervised or unsupervised detector as long as its output can be interpreted as a probability distribution over two categories, with no additional training data.

The main contributions of this work are three-fold:

\noindent 1. We formalize the problem of detecting multi-armed attacks in a zero-shot setting as a minimax problem (cf.~\cref{optimal-loss}). We suppose the defender has access to a set of ``off-the-shelf'' detectors, but no direct access to the channel or new data points. 

\noindent 2. Based on this formulation, we characterize the optimal soft-detector in~\cref{eq:minimaxProb4}, leading to our proposed solution. 
Furthermore, we propose an upper bound on the detection error of the proposed solution in the case where the assumptions for the optimum are not met (cf.~\cref{sec:optimal_objective}).

\noindent 3. Finally, we consider a practical use case for the empirical evaluation, where the channel is a pre-trained neural network classifier, and a multi-armed attacker, which has white-box access to it, mounts well-known adversarial attacks against it. We empirically evaluate the proposed solution on popular computer vision datasets, such as CIFAR10 and SVHN\footnote{Code available at \url{https://github.com/fgranese/Optimal-Zero-Shot-Detector-for-Multi-Armed-Attacks}.}. The results show that the proposed method leads to higher and more consistent performance compared to the state-of-the-art (SOTA) in the multi-armed attack setup, including in settings that deviate from the optimum (cf.~\cref{sec:experiments}).

\section{THREAT MODEL}
\label{sec:math_framework}
Let $X$ be the random variable (r.v.) for which we have realizations $\x\in\mathcal{X}\subseteq\mathbb{R}^d$. Let $Y$ be the r.v. representing the class label, taking values $y\in\mathcal{Y}=\{1,\dots,C\}$. Samples $(\x,y)$ are i.i.d. from the generative distribution $P_{XY}$.
The target channel is a classifier described by the function $h_{\theta}:\mathcal{X}\rightarrow\mathbb{R}^{|\mathcal{Y}|}$. Generally, the output of $h_{\theta}$ is normalized to represent the posterior distribution of $p_{\widehat{Y}|X}(\cdot|\x; \theta)$.
The inference induced by the channel $h_{\theta}$ is defined as $g_{\theta}:\mathcal{X}\rightarrow\mathcal{Y}$
s.t. ${g_{\theta}(\x)\myeq \arg\max_{y\in\mathcal{Y}} p_{\widehat{Y}|X}(y|\x; \theta)}$.
We consider a scenario where the attacker is a malicious actor that intercepts the input signal $X$ before it reaches the channel and decides whether to perturb it or not according to a Bernoulli distribution $\mathcal{B}er(p)$, where $p\in[0,1]$ is the perturbation probability.
For a given signal $\x$, if it is perturbed the channel $h_{\theta}$ will receive a corrupted input $\x'$, otherwise the original natural input $\x$.
        
\subsection{The Attacker}
A corrupted input $\x'$ is created according to a\textit{ multi-armed attack} scheme. In particular, the attacker has access to a set of different white-box attack strategies on the target classifier.
Let $\mathcal{K}$ be the countable set of indexes, each corresponding to one attack strategy. 
For the sake of simplicity, we assume that the distribution over $\mathcal{K}$ is uniform. Thus, we define the $k$-th attack strategy as $\mathcal{A}^{(k)}_{\varepsilon, p}: \mathcal{X}\rightarrow\mathbb{R}^d_{\varepsilon,p}$, where, with abuse of notation, $\mathbb{R}^d_{\varepsilon,p}$ represents the subset of $\mathbb{R}^d$ where the corrupted samples lie, according to $\mathcal{A}^{(k)}_{\varepsilon, p}$. Therefore, $\mathcal{A}^{(k)}_{\varepsilon, p}$ can generate a corrupted sample 
\begin{align}
    \label{eq:adv_problem}
    \x'&\myeq\underset{||\x'-\x||_p<\varepsilon}{\arg\max}\,\ell(\x',\x; \theta),
\end{align}
where $\ell(\cdot)$
represents a loss function on the classifier, for inputs $\x'$ and $\x$.
Moreover, let  $\mathcal{M}=\big\{P^{(k)}_{XZ}\,  : \, k\in\mathcal{K} \big\}$ be the set of joint probability distributions on  $\calX\times\calZ$ which are indexed with ${k,~\forall k\in\calK}$, where $\calX$ is the input (feature) space and $\calZ=\{0, 1\}$ indicates a binary space label for the adversarial example detection task.
The attacker selects an arbitrary strategy $k\in \mathcal{K}$ and then samples an input according to $p^{(k)}_{X|Z}(\x|z=1)$ which corresponds to the probability density function induced by the chosen attack $k$. According to the Bernoulli threat model described above, $p^{(k)}_{X|Z}(\x|z=0)= p_X(\x) $ \emph{almost surely} corresponds to the probability distribution of the natural samples, i.e. the case in which the attacker does not perturb the input signal. 
Notice that, for the sake of simplicity, with a slight abuse of notation, in the rest of the paper, we will use $\x$ to mean a generic input. The fact that it is a clean sample or a corrupted one is understood from the context, given the presence of the variable $Z$.
\subsection{The Defender}
We assume that the defender has no access to the target classifier, or to any distribution for data sampling, and therefore it cannot change, robustly re-train, or certify it. Finally, the defender has access only to pre-trained detectors that are given by a third party and are also not re-trainable, due to the impossibility of collecting new samples\footnote{Notice that this is a crucial set of assumptions: it drastically limits the amount of information directly available to the defender, and therefore its capabilities. On the other hand, it is also a realistic one since the literature on this topic proposes a rich body of detectors that can be used off the shelf, and are effective at least against one attack strategy among those available to the multi-armed attacker (cf.~\cite{GranesePRMP2022ECMLPKDD}).}. 

Formally, the defender is given a set of \textit{soft-detectors} models:
$$
\Q = \left\{q_{\widehat{Z}|\logit}^{(k)} \, :\, \calU \mapsto [0,1]^2  \right\}_{k\in \calK},
$$
which have possibly been trained to detect attacks according to each strategy $k\in\mathcal{K}$, e.g.,  ${q_{\widehat{Z}|\logit}^{(k)} \equiv  p_{\widehat{Z}|U}(z|\logit; \psi_k)}$ with parameters $\psi_k$ and $\logit\in\calU = \{\classifier(\x)~|~\x\in \R^d\}$ denotes the space of logits. 
The set of possible detectors $\Q$ is available to the defender, however, the specific attack chosen by the attacker at the test time is unknown. 

The following section will go over how to best aggregate detector decisions.

\section{AN OPTIMAL OBJECTIVE FOR DETECTION UNDER MULTI-ARMED ATTACKS}
\label{sec:optimal_objective}
We start by considering the case in which for each attack in $\mathcal{M}$, there exists at least a detector in $\mathcal{Q}$ that has been trained to detect it. We formally devise an optimal solution that exploits full knowledge of $\Q$. 

Consider a fixed input sample $\mathbf{x_0}$ and let $\mathbf{u_0}=\classifier(\mathbf{x_0})$. Clearly, the problem at hand consists in finding an optimal soft-detector  ${q}^{\star}_{\widehat{Z}|\mathbf{u_0}}$ that performs well simultaneously over all possible attacks in  $\calK$. This can be formalized as the solution to the following minimax problem: 
\begin{equation}
\mathcal{L}(\Q, \mathbf{x_0}) = \min_{{q}_{\widehat{Z}|\mathbf{u_0}}} \max_{k\in\calK}\, \EE_{q^{(k)}_{\widehat{Z}| \mathbf{u_0}}}\left[ -\log {q}_{\widehat{Z}|\mathbf{u_0}}  \right], \label{optimal-loss}
\end{equation}
which requires solving \eqref{optimal-loss} for $\Q$ and for each given input sample $\mathbf{x_0}$. 
It is important to note that the minimization is performed over all distributions ${q}_{\widehat{Z}|\mathbf{u_0}}$, including elements that are not part of the set $\mathcal{Q}$. 
The Cross-Entropy term in \cref{optimal-loss} represents the measure of the agreement between our target detector $q_{\widehat{Z}|\mathbf{u}_0}$ and one of the possible $q_{\widehat{Z}|\mathbf{u}_0}^{(k)}$. Overall,  as the defender remains unaware of whether $\mathbf{u}_0$ originates from an adversarially perturbed sample or which of the $|\calK|$ possible strategies the attacker has utilized, the minimax problem in \cref{optimal-loss} formalizes the attempt to minimize the average worst-case regret of detecting adversarial examples, when the attacker is allowed to mount multi-armed attacks.

That being said, the objective in~\cref{optimal-loss} is not tractable computationally. To overcome this issue, we derive a surrogate (an upper bound) that can be computationally optimized. For any   arbitrary choice of  ${q}_{\widehat{Z}|\mathbf{u_0}}$, we have 
\begin{align}
  \max_{k\in\calK}\,  \EE_{q^{(k)}_{\widehat{Z}| \mathbf{u_0}}} &\left[ -\log {q}_{\widehat{Z}|\mathbf{u_0}}  \right] \leq \nonumber\\ 
  &\underbrace{\max_{k\in\calK}\, \EE_{q^{(k)}_{\widehat{Z}| \mathbf{u_0}}}\left[ -\log q^{(k)}_{\widehat{Z}| \mathbf{u_0}} \right]}_{=\textrm{constant term}} + \nonumber\\ &+
  \max_{k\in\calK}\, \EE_{q^{(k)}_{\widehat{Z}| \mathbf{u_0}}}\left[\log\left(\frac{q^{(k)}_{\widehat{Z}|\mathbf{u_0}}}{{q}_{\widehat{Z}|\mathbf{u_0}}}\right)\right].  
  \label{eq-missing}
\end{align}
See \cref{app:proof1} for the proof.

Notably, the first term of the upper bound in \eqref{eq-missing} is constant w.r.t. the choice of  ${q}_{\widehat{Z}|\mathbf{u_0}}$ and the second term is well-known to be equivalent to the \emph{average worst-case regret}~\cite{BarronRY1998TInfT}. This upper bound provides a surrogate to our intractable objective in \eqref{optimal-loss} that can be minimized overall ${q}_{\widehat{Z}|\mathbf{u_0}}$.  We can formally state our problem as follows: 
\begin{align}
  \label{eq:minimaxProb1}
\tilde{\mathcal{L}}(\Q,\mathbf{x_0}) = &    \min_{{q}_{\widehat{Z}|\mathbf{u_0}}}\max_{k\in\calK}\, \EE_{q^{(k)}_{\widehat{Z}| \mathbf{u_0}}}\left[\log\left(\frac{q^{(k)}_{\widehat{Z}|\mathbf{u_0}}}{{q}_{\widehat{Z}|\mathbf{u_0}}}\right)\right]\nonumber\\=&\min_{{q}_{\widehat{Z}|\mathbf{u_0}}}\max_{\textcolor{black}{P_{\Omega}}}\, \EE_{\Omega}\left[D_{\textrm{KL}}\left(q^{(\Omega)}_{\widehat{Z}|\mathbf{u_0}}\big \| {q}_{\widehat{Z}|\mathbf{u_0}}\right)\right],
\end{align}
\textcolor{black}{where the $\min$ is taken over all the possible distributions ${q}_{\widehat{Z}|\mathbf{u_0}}$;  
and $\Omega$ is a discrete random variable with $P_{\Omega}$ denoting a generic  probability  distribution whose probabilities are  $(\omega_1,\dots,\omega_{|\calK|})$, i.e., $P_{\Omega}(k)=\omega_k$;}
and $D_{\textrm{KL}}(\cdot\|\cdot)$ is the Kullback–Leibler divergence, representing the expected value of regret of ${q}_{\widehat{Z}|U}$ w.r.t. the worst-case distribution in   $\Q$. 
See \cref{app:proof2} for the proof.

The convexity of the KL-divergence allows us to rewrite~\cref{eq:minimaxProb1} as follows:
\begin{align}
    \label{eq:minimaxProb3}
    &\min_{{q}_{\widehat{Z}|\mathbf{u_0}}}\max_{\textcolor{black}{P_\Omega}}\EE_{\Omega}\left[D_{\textrm{KL}}\left(q^{(\Omega)}_{\widehat{Z}|\mathbf{u_0}}\big \| {q}_{\widehat{Z}|\mathbf{u_0}}\right)\right]=\nonumber\\&=  \max_{\textcolor{black}{P_\Omega}}\min_{\widehat{q}_{\widehat{Z}|\mathbf{u_0}}}\EE_{\Omega}\left[D_{\textrm{KL}}\left(q^{(\Omega)}_{\widehat{Z}|\mathbf{u_0}}\big \| {q}_{\widehat{Z}|\mathbf{u_0}}\right)\right]. \end{align}
   See \cref{app:proof3} for the proof.


The solution to~\cref{eq:minimaxProb3} provides the optimal distribution \textcolor{black}{$P_{\Omega}^\star$}, i.e. the collection of weights $\{w_k^{\star}\}$, which leads to our soft-detector~\cite{BarronRY1998TInfT}: 
\begin{align}
    \label{eq:minimaxProb4}
 \widehat{q}^{~\star}_{\widehat{Z}|\mathbf{u_0}} = \sum_{k\in\calK}w^{\star}_k \cdot q^{(k)}_{\widehat{Z}|\mathbf{u_0}}  , \ \ \text{ with } \ \ 
 \textcolor{black}{ P_{\Omega}^\star=\arg\max_{\{\omega_k\}}I_{\mathbf{u_0}}(\Omega;\widehat{Z}), }
\end{align}
where $I_{\mathbf{u_0}}(\cdot;\cdot)$ denotes the Shannon mutual information between the random variable \textcolor{black}{$\Omega$}, distributed according to \textcolor{black}{$\{\omega_k\}$}, and the binary soft-prediction variable $\widehat{Z}$, distributed according to $q^{(k)}_{\widehat{Z}|\mathbf{u_0}} $ and  conditioned on the particular test example $\mathbf{u_0}$. 
See \cref{app:proof4} for the proof.

\noindent\textbf{From Theory to Our Practical Detector.} According to our derivation in~\cref{eq:minimaxProb4}, the optimal detector turns out to be given by a  mixture of the $|\calK|$ detectors belonging to the class $\Q$, with weights carefully optimized to maximize the mutual information between $\Omega$ and the predicted variable $\widehat{Z}$ for each detector in the class $\Q$. Using this key ingredient, it is straightforward to devise our optimal detector. 

\begin{definition}
\label{def:salad}
For any $0\leq\gamma\leq1$ and a given $\mathbf{x_0} \in\calX $, let us define the following detector $\textsc{D}:\mathbb{R}^d\rightarrow\{0, 1\}$:  
\begin{align}
\textsc{d}(\x_0)\myeq \mathds{1} \left[ {q}^{~\star}_{\widehat{Z}|\mathbf{u_0}}(1|\classifier(\mathbf{x_0}))>\gamma \right],
\label{eq:salad}
\end{align}
where $\mathds{1} \left[\cdot\right]$ is the indicator function.
\end{definition}

\noindent\textbf{On the Optimization of~\cref{eq:minimaxProb4}}. We implement the well-known \textit{Blahut–Arimoto algorithm}~\cite{Arimoto72}, an iterative algorithm for finding the capacity of a channel. Further details can be found in~\cref{app:optimization}.

\subsection{On the Consequences of Domain Shift}
\label{sec:domain_shift}
So far we have described an optimal solution to aggregate $|\mathcal{K}|$ detectors such that each of them has been trained to effectively detect natural samples and samples that have been corrupted according to one of the $k\in\mathcal K$ attack strategies, i.e. the \textit{source domain}. Let us now suppose that a new strategy $k^* \notin \mathcal{K}$ is introduced. Clearly, none of the detectors we aggregated has ever seen corruptions produced by this strategy, thus one may wonder whether the aggregated detector will be able to detect samples corrupted according to $k^*$, i.e. the \textit{new domain}. In this section, we consider this problem and provide an upper bound on detection error for the new domain as a function of the detection error for the previous domain. 

Let us consider a detector \textsc{d}, like the one defined in \cref{eq:salad}. Let us also assume a function $f^{S}:\mathbb{R}^{d}\rightarrow\{0,1\}$, i.e. the source label function (oracle) which assigns a label to any input sample distributed according to the source domain. The source domain, defined as $P^{S}_{X}$ is the distribution of the input $X$ (natural or adversarial) over the input space $\mathbb{R}^{d}$, where the adversarial examples are generated according to the possible $|\mathcal K|$ strategies of which our detector is aware.

Similarly, we define $f^{\newdomain}:\mathbb{R}^{d}\rightarrow\{0,1\}$, i.e. the  label function relative to the new domain. The new (testing) domain, defined as $P^{\newdomain}_{X}$ is the distribution of the input $X$ (natural or adversarial), where the adversarial examples are generated according to the strategy $k^{\star}$, which is new to our detector. 

We can now define the source error:
\begin{align}
    P_e^{S}(\textsc{D})\myeq \EE_{\x\sim P^{S}_{X}}\left[\mathds{1}\left[\textsc{D}\left(\x\right)\neq f^{S}(\x)\right]\right],
\end{align}
and the error on the new domain:
\begin{align}
    P_e^{\newdomain}(\textsc{D}) \myeq  \EE_{\x\sim P^{\newdomain}_{X}}\left[\mathds{1}\left[\textsc{D}\left(\x\right)\neq f^{\newdomain}(\x)\right]\right]. 
\end{align}
Let 
\begin{align}
d\left(P^{S}_{X|Z=1},P^{\newdomain}_{X|Z=1}\right) \myeq 2\sup_{B\in\mathcal{X}} \big | {\Pr}_{S}(B) - {\Pr}_{\newdomain}(B) \big|,
\end{align}
where $\mathcal{X}$ is the set of measurable subsets under the noise distributions $P^{S}_{X|Z=1}$, and $P^{\newdomain}_{X|Z=1}$. Then, according to ~\cite{Ben-DavidBCKPV2010ML},
\begin{align}
    \nonumber
    P_e^{\newdomain} (\textsc{D}) &\leq P_e^{S}(\textsc{D}) + d\left(P^{S}_{X|Z=1},P^{\newdomain}_{X|Z=1}\right)
    \\
    &+ \min\big\{\EE_{
    \x\sim P^{S}_{X}}[|f^{S}(\x)-f^{\newdomain}(\x)|],\nonumber 
    \\
    &\EE_{\x\sim P^{\newdomain}_X}[|f^{S}(\x)-f^{\newdomain}(\x)|]\big \}.
\end{align}
Intuitively, as the detector has never seen samples from the new domain, it is expected to perform worse on it. Conversely, the above bound indicates that the loss in terms of performance is expected to be low proportionally to a small $d\left(P^{S}_{X|Z=1},P^{\newdomain}_{X|Z=1}\right)$ of the noises between the domains. The proof is provided in~\cref{app:proof_thm}.

\section{RELATED WORKS}
\label{sec:related}
We explore a multi-armed attack scheme, a variation of the well-known multi-armed bandit (MAB) problem where an agent selects attacks, i.e. arms, to maximize its cumulative reward~\cite{Slivkins2019introduction}. 
We can see the multi-armed attack scheme as the MAB problem, where the agent (the attacker) chooses all the arms (attacks) for each round (sample). 

In this sense, the work in~\cite{GranesePRMP2022ECMLPKDD} has developed the idea of multi-armed attacks in the context of adversarial examples.
These examples are crafted patterns specifically designed starting from `natural' or `clean' samples to fool a model into making incorrect predictions. Generally, to combat this issue, there are two main strategies: robust training and adversarial detection. Robust training (e.g.,~\cite{MadryMSTV2018ICLR,Zimmermann2022,Kang2019testing}) aims to make a model more resistant to adversarial examples, while adversarial detection (e.g.,~\cite{aldahdooh2022adversarial,PangZHD000L22,RaghuramCJB21}) attempts to identify such examples. 
According to the {\mead} framework in~\cite{GranesePRMP2022ECMLPKDD}, a target classifier is simultaneously attacked using multiple strategies, with no extra detector-specific information. The detector is then evaluated on all crafted adversarial examples, achieving success only if it correctly identifies all attacks.

The ``cat-and-mouse'' game of adversarial attacks against target classifiers has motivated the community to find different solutions, such as robustness certification, 
~\cite{RaghunathanSL2018ICLR,CohenRK2019ICML,Levine2021ICML}. In general, a certificate is a trainable function that is optimized at training time to ensure that the decision boundary of the classifier is guaranteed not to change within a perturbation radius.
Other than limitations intrinsic to the nature of certified classifiers, such as a trade-off between certified radius and accuracy and training complexity~\cite{VaishnaviER2022NeurIPS}, it is important to mention that in general, these techniques require \emph{large amount of data} (e.g. randomized smoothing~\cite{CohenRK2019ICML}) replacing the channel with a certified one. Note that in our scenario \emph{the impossibility of replacing the channel, and the lack of additional samples is a key assumption}. Some efforts to achieve zero-shot certifiability have been recently made, but crucially, the assumption is to be able to aggregate several diffusion models as denoisers with
the channel~\cite{CarliniTDRSK2023ICLR} or rely on networks with small Lipschitz constant as channels~\cite{DelattreABA2023CoRR}. Finally, to the best of our understanding, certification, although more general than detection, is hardly universal, and has not been deployed against multi-armed attacks~\cite{NandiARB2023CVPR_WS,PalSV2020NeurIPS}. \cite{PalV2020NeurIPS} proposes a game-theoretic analysis of the adversarial attack/defense problem, although it does not consider multi-armed attacks and its results are confined to a binary classification setting
with locally linear decision boundaries.

\section{EXPERIMENTAL RESULTS}
\label{sec:experiments}
In our empirical evaluation, we apply the provided theoretical framework within the domain of adversarial attack detection. In this context, multi-armed adversarial attack detection has previously been explored solely from an empirical standpoint in~\cite{GranesePRMP2022ECMLPKDD}, yet no solution to the problem is presented therein. 

\begin{figure}[!htbp]
    \centering
    \includegraphics[width=.9\columnwidth]{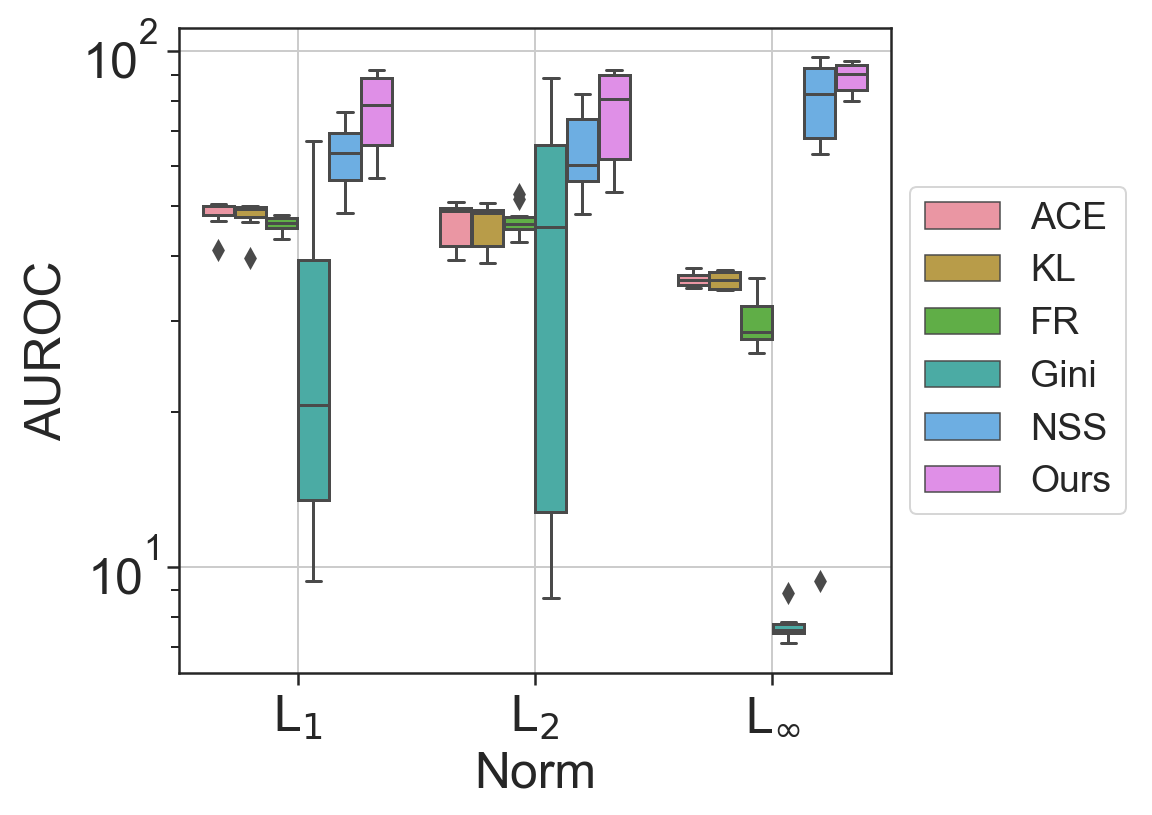}
    \caption{The \textit{shallow} detectors are named after the loss function used to craft the attacks they are trained to detect. Overall, NSS clearly outperforms all the individual shallow detectors. The aggregation we propose allows us to use the shallow models to attain a detector whose performance is consistently comparable and in many cases better than NSS.}
    \label{fig:box_plot}
\end{figure}

To align with the theoretical framework in~\cref{sec:math_framework}, we assume that the attacker has white-box access to the target classifier $h_\theta$, but has no information about the defense. Conversely, we assume that a third party provides the defender with four simple supervised detectors. Each of them is trained to effectively detect attacks crafted by a single specific attack strategy. 
This is a reasonable assumption, as many methods in the literature are able to successfully detect at least one type of attack and fail at detecting others.
In addition, to emphasize the role played by the proposed method, these detectors are merely shallow networks (3 fully connected layers with 256 nodes each), which are only allowed to observe the logits of the target classifier to distinguish between natural and adversarial samples. 
Due to their specifics, these individual shallow detectors are bound to perform very poorly, i.e. much worse than SOTA detectors, against attacks they have not been trained on, as shown in~\cref{fig:box_plot}. This aspect enhances the value of our solution, which attains favorable performance by aggregating detectors that individually exhibit subpar performance w.r.t. SOTA detection methods. 

\cref{sec:res_th_framework} reports results in the optimal setting introduced in~\cref{sec:optimal_objective}, whereas~\cref{sec:res_mead} reports results in the setting of~\cite{GranesePRMP2022ECMLPKDD}, when the assumptions for the optimal setting are not met, i.e. attacks may come from a new domain not known at the detectors' level (cf.~\cref{sec:domain_shift}).

\subsection{Evaluation Framework}
\label{sec:eval_framework}
\noindent\textbf{Datasets and Pre-Trained Classifiers}.
We run our experiments on CIFAR10~\cite{Cifar} and SVHN~\cite{SVHN}. For both, the pre-trained target classifier is a ResNet-18 model that has been trained for $100$ epochs, using SGD optimizer with a learning rate of $0.1$, weight decay equal to $10^{-5}$, and momentum equal to $0.9$. The accuracy achieved by the classifiers on the original clean data is 99\% for CIFAR10 and 100\% for SVHN over the train split; 93.3\% for CIFAR10 and 95.5\% for SVHN over the test split.

\noindent\textbf{Attack Strategies.} We consider both \textit{white-box} and, for completeness, \textit{black-box} adversarial attacks\footnote{We remind that the attacks are perpetrated over the target classifier $h_\theta$.} 
For the white-box scenario, i.e. when the attacker has complete knowledge of the channel's parameters, we consider \textit{Fast Gradient Sign Method} (FGSM)~\cite{FGSM}, \textit{Basic Iterative Method} (BIM)~\cite{BIM} and \textit{Projected Gradient Descent} (PGD)~\cite{MadryMSTV2018ICLR}, \textit{Carlini-Wagner} attack (CW)~\cite{CarliniW2017SP} and \textit{DeepFool} attack (DF)~\cite{DF}. 
In the case of \textit{black-box} attacks, the adversary has no access to the internals of the target model, hence it creates attacks by querying the model and monitoring the outputs of the model to attack. Within this category of attacks we contemplate \textit{Square Attack} (SA)~\cite{SA}, \textit{Hop Skip Jump} attack (HOP)~\cite{HOP} and \textit{Spatial Transformation Attack} (STA)~\cite{EngstromTTSM19}.
We refer to the survey in~\cite{aldahdooh2022adversarial} and references therein for an extensive discussion of this topic. 

\noindent{\textbf{\mead}~\cite{GranesePRMP2022ECMLPKDD}}. The pre-trained classifier is attacked simultaneously with multiple attack strategies without extra information on the specific detector. 
To create a set of simultaneous attacks (cf.~\cref{tab:attacks} in \cref{app:attacks}), multiple perturbed versions of the same natural input sample are created according to the set of attack strategies, objective loss $\ell$, perturbation magnitude $\varepsilon$, and norm
L$_p$. We consider $\ell$ to be either the \textit{Adversarial Cross Entropy loss} (ACE)~\cite{SzegedyZSBEGF13,MadryMSTV2018ICLR}, or the \textit{Kullback-Leibler divergence} (KL), or the \textit{Fisher-Rao objective} (FR)~\cite{PicotMBLBAP2022TPAMI} or the \textit{Gini Impurity score} (Gini)~\cite{GraneseRGPP2021NeurIPS}.
The perturbed samples unable to fool the target classifier are discarded. Finally, the detector is evaluated on all the crafted adversarial examples, and only if all the attacks are correctly identified the detection is successful. 

\noindent\textbf{Detectors}.
\begin{figure*}[t]
	\centering
	\begin{subfigure}[b]{ .5\columnwidth}
	    \centering
	    \includegraphics[width=\columnwidth]{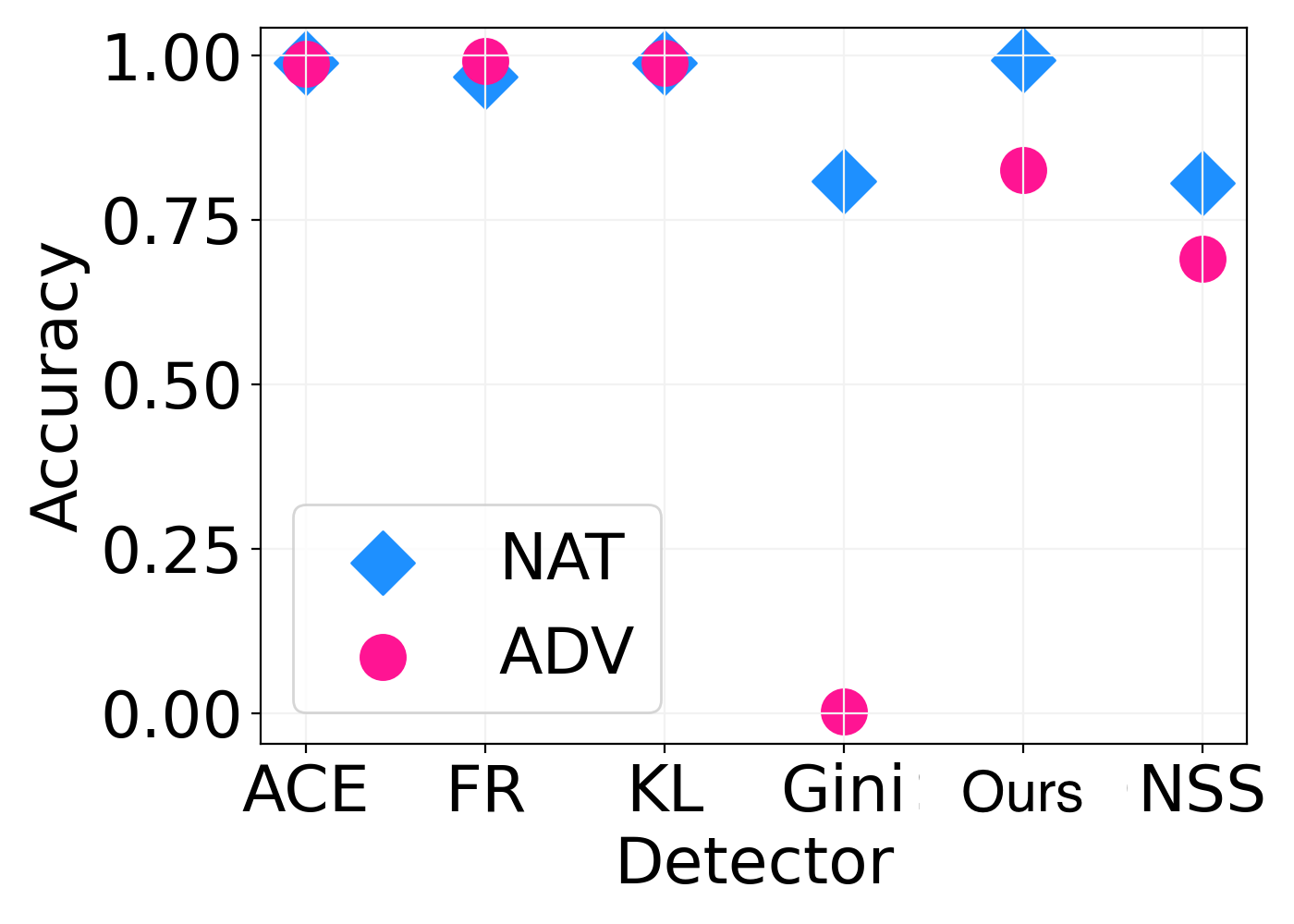}
	    \caption{Attacks with PGD algorithm, FR loss, $\varepsilon=40$, and norm L$_1$}
	    \label{fig:acc_pgd1}
	\end{subfigure}
	\hfill
		\begin{subfigure}[b]{ .5\columnwidth}
		\centering
		\includegraphics[width=\columnwidth]{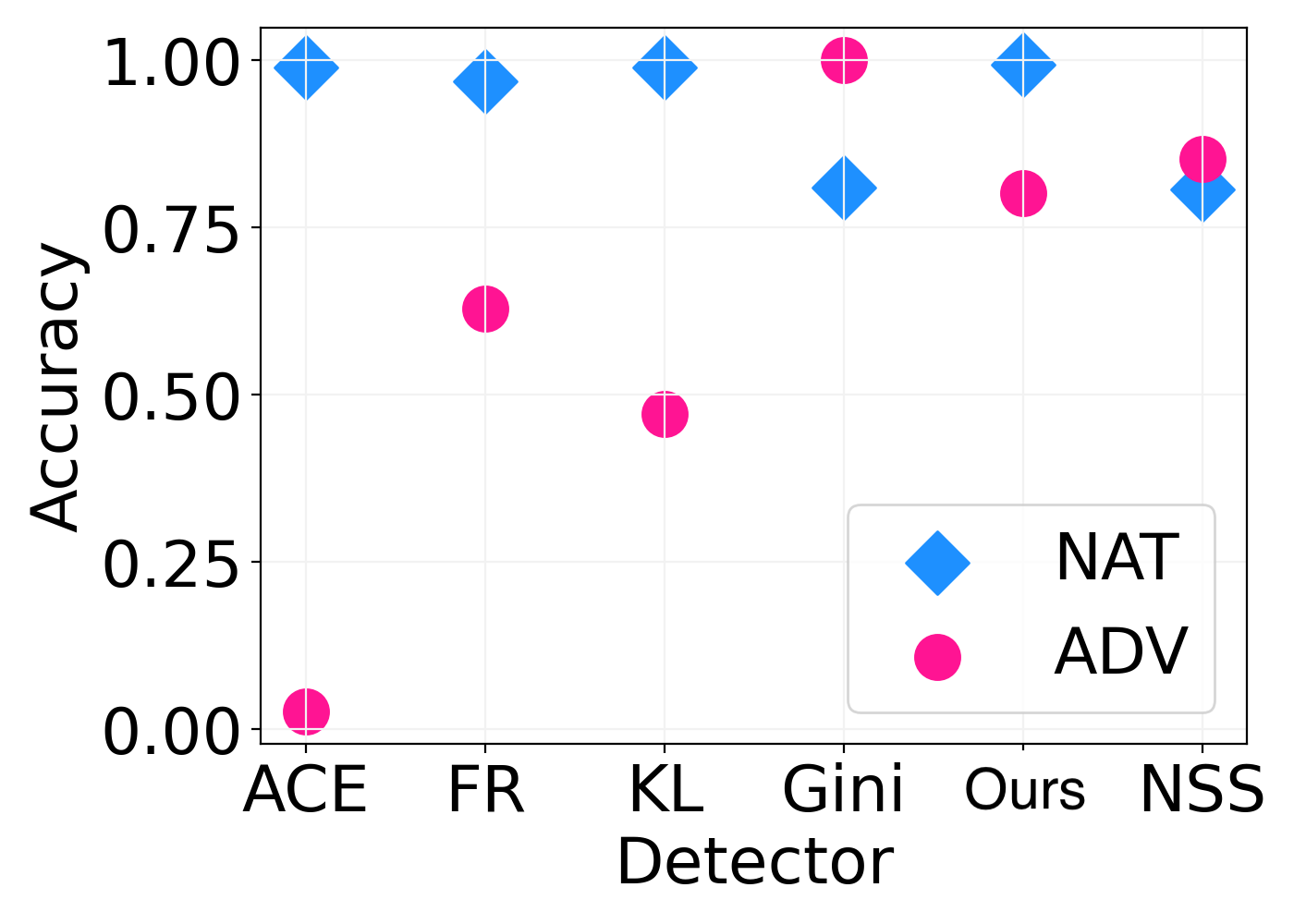}
		\caption{Attacks with FGSM algorithm, FR loss, $\varepsilon=0.5$, and norm  L$_\infty$}
		\label{fig:acc_fgsm}
	\end{subfigure}
\hfill
	\begin{subfigure}[b]{ .5\columnwidth}
		\centering
		\includegraphics[width=\columnwidth]{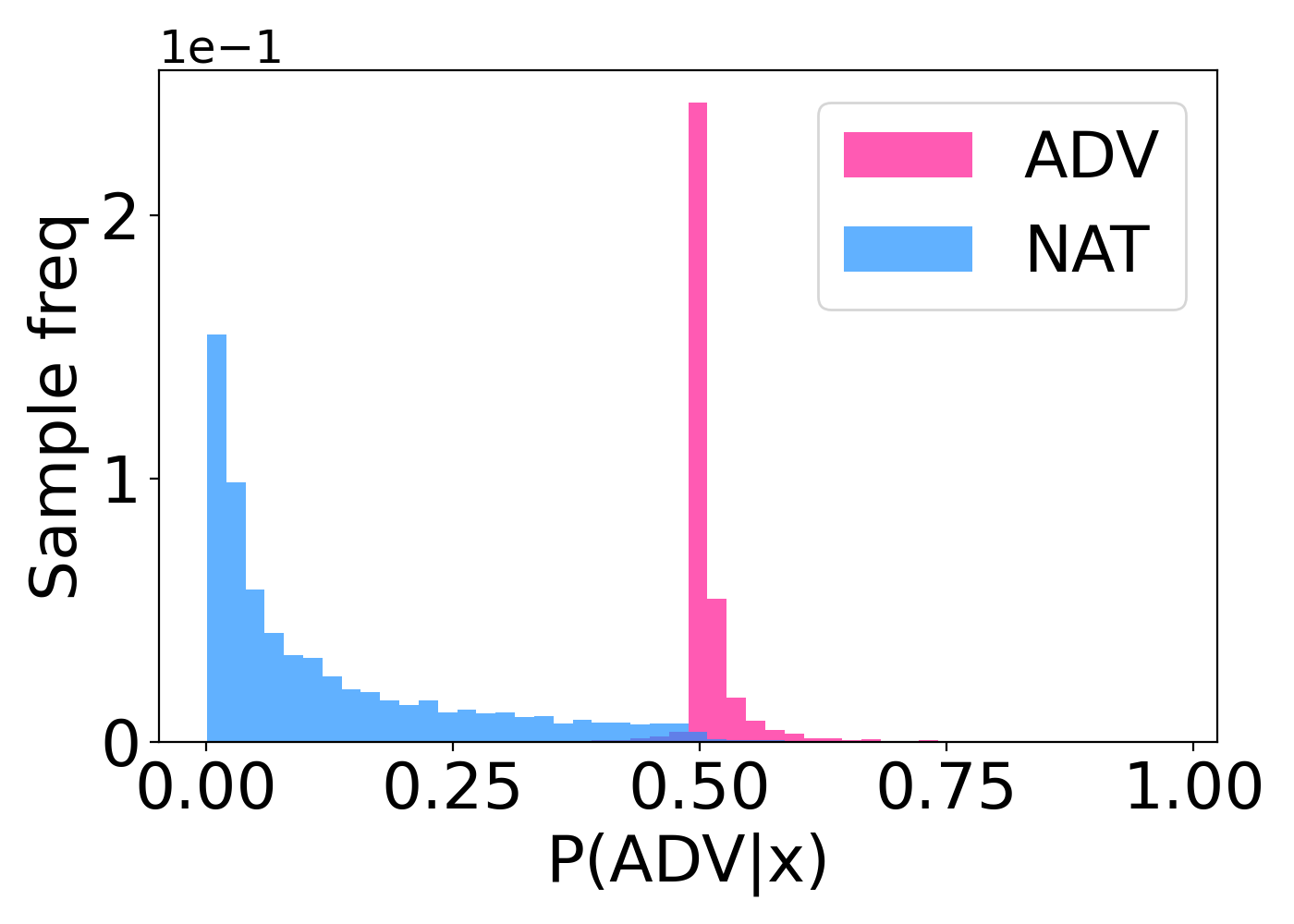}
		\caption{Ours against attacks with PGD algorithm, FR loss, $\varepsilon=40$, and norm  L$_1$}
		\label{fig:hist_salad}
	\end{subfigure}
\hfill
	\begin{subfigure}[b]{ .5\columnwidth}
		\centering
		\includegraphics[width=\columnwidth]{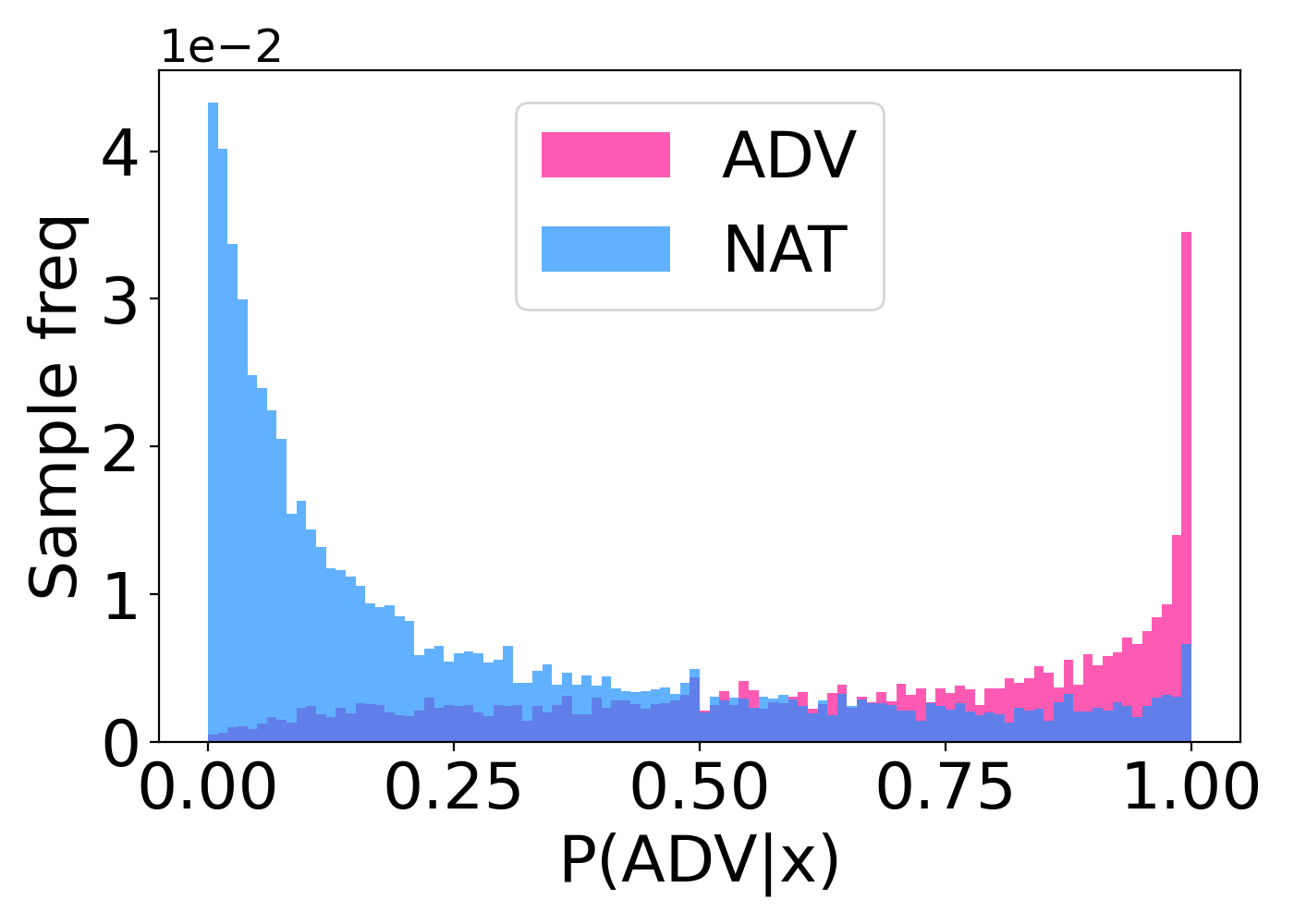}
		\caption{NSS against attacks with PGD algorithm, FR loss, $\varepsilon=40$, and norm  L$_1$}
		\label{fig:hist_nss}
	\end{subfigure}
	\caption{
	In~\cref{fig:acc_pgd1} and~\cref{fig:acc_fgsm}, the accuracies of the detectors on natural and adversarial examples; in~\cref{fig:hist_salad} and~\cref{fig:hist_nss} we show how the proposed method and NSS split the data samples. We report the results for the detection of adversarial examples in pink, and the results for the detection of the natural in blue.}
	\label{fig:evaluation}
\end{figure*} 
The proposed method aggregates four simple pre-trained detectors.
The detectors are four fully-connected neural networks, composed of 3 layers of 256 nodes each. All the detectors are trained for 100 epochs, using SGD optimizer with learning rate of 0.01 and weight decay 0.0005. They are trained to distinguish between natural and adversarial examples created according to the PGD algorithm, under L$_\infty$ norm constraint and perturbation magnitude $\varepsilon=0.125$ for CIFAR10 and $\varepsilon=0.25$ for SVHN. Each detector is trained on natural and adversarial examples generated using one of the loss
functions mentioned in~\cite{GranesePRMP2022ECMLPKDD} (i.e., ACE Eq.~(3), KL Eq.~(4), FR Eq.~(5), Gini Eq.~(6) of the corresponding paper) to craft its adversarial training samples. We want to point out that the purpose of this paper is not to create a new supervised detector but rather to show a method to aggregate a set of pre-trained detectors. Moreover, it is important to notice that either supervised or unsupervised methods can be added to or pool of detectors, provided that their output is the confidence on the input sample being or not an adversarial example. We further expand on the selection of the $\varepsilon$ parameter of the adversarial examples used at training time in~\cref{app:varius_eps} (cf.~\cref{tab:cifar10_salad,tab:svhn_salad}).

\noindent\textbf{NSS}~\cite{NSS}. We compare the proposed method with NSS, which is the best among the supervised SOTA methods against multi-armed adversarial attacks (cf.~\cite{GranesePRMP2022ECMLPKDD}).
NSS characterizes the adversarial perturbations using \textit{natural scene statistics}, i.e., statistical properties that the presence of adversarial perturbations can alter. 
NSS is trained by using PGD algorithm, L$_\infty$ norm constraint and perturbation magnitude $\varepsilon=0.03125$ for CIFAR10 and $\varepsilon=0.0625$ for SVHN. We further expand on the selection of the $\varepsilon$ parameter of the adversarial examples used at training time in~\cref{tab:cifar10_nss,tab:svhn_nss,app:varius_eps}. 

\noindent\textbf{Evaluation Metrics}. For each sample and for each group of attacks we consider a detection successful, i.e. a true positive, if and only if all the adversarial attacks are detected. Otherwise, we report a false negative. 
We use the classical definitions of \textit{true negative} and \textit{false positive} for the natural samples detection. This means that a true negative is a natural sample detected as natural, and a false positive is a natural sample detected as adversarial.
We measure the performance of the detectors in terms of $i)$ \underline{\auc}~\cite{davis2006relationship} (\textit{Area Under the Receiver Operating Characteristic curve}), i.e., the detector's ability to discriminate between adversarial and natural examples (higher is better); $ii)$ \underline{FPR at 95 \% TPR (\fpr)}, i.e., the percentage of natural examples detected as adversarial when 95 \% of the adversarial are detected (lower is better).

\subsection{Discussion}
We present the main experimental results to show the effectiveness of the proposed method for multi-armed adversarial attack detection.
Further discussion and additional results, can be found in~\cref{app:experiments}. 

\subsubsection{The \textit{Shallow} Detectors}
\label{sec:discussion}
\cref{fig:evaluation,fig:box_plot,fig:eps_norm} provides a graphical interpretation of the detection performance when the target classifier is ResNet18, trained on CIFAR10. The single detectors are named after the loss function used to craft the adversarial examples on which each detector is trained along with the natural samples. The main takeaway from~\cref{fig:box_plot} is the observation that, when considered individually, the shallow detectors are clearly subpar w.r.t. the SOTA adversarial attacks detection mechanism. On the contrary, the aggregation provided by our method results in detection performance that is comparable to SOTA performance and, in some cases, outperforms well-established detection mechanisms. 
\Cref{fig:eps_norm} sheds a light on the fact that the performance attained by our proposed method can consistently improve the detection of adversarial examples over several multi-armed attacks mounted using different norms and perturbation magnitudes.

A key insight of this paper is that when provided with generally non-robust detectors whose performance is good only against a limited amount of attacks (as it is confirmed by~\cref{fig:box_plot,fig:eps_norm}), we can successfully aggregate them through the proposed method to obtain a consistently better detection.

In~\cref{fig:evaluation} we consider attacks crafted according to the PGD algorithm, the FR loss, $\varepsilon=40$, and norm constraint L$_1$ (cf.~\cref{fig:acc_pgd1,fig:hist_salad,fig:hist_nss}), and attacks crafted according to the FGSM algorithm, FR loss, $\varepsilon=0.5$, and L$_\infty$ norm in~\cref{fig:acc_fgsm}.
We also report the performance of the considered detectors in terms of detection accuracy over the natural examples in blue and the adversarial examples in pink. 
As we can observe, the individual detectors, which are named after the loss functions ACE, FR, KL, and Gini, exhibit different behaviors for the specific attack. In~\cref {fig:acc_pgd1}, the Gini detector drastically fails at detecting the attack as its accuracy plummets to 0\% on the adversarial examples. 
In the same way, the FR and KL detectors but mostly the ACE detector, perform poorly against FGSM (cf.~\cref{fig:acc_fgsm}). On the contrary,
our method, benefiting from the aggregation, obtains favorable results in both cases, confirming what we had observed. 

The histograms in~\cref{fig:hist_salad,fig:hist_nss} show how the method we propose and NSS separate natural (blue) and adversarial examples (pink), respectively. The values along the horizontal axis represent the probability of being classified as adversarial, and the vertical axis represents the frequency of the samples within the bins.
The detection error is proportional to the area of overlap between the blue and the pink histograms.~\cref{fig:hist_salad} and~\cref{fig:hist_nss} suggest that the proposed method achieves lower detection error on the considered attack, as it is confirmed in~\cref{tab:final_table} where our proposed method attains 92.1 {\auc}, while NSS only achieves 76.1 {\auc}. Additional plots are provided in~\cref{app:additional_plot}.


\subsubsection{Evaluation of the Proposed Solution in the Optimal Setting}
\begin{figure*}[!htbp]
	\centering
    	\centering
		\begin{subfigure}[b]{0.5\columnwidth}
		\centering
		\includegraphics[width=\columnwidth]{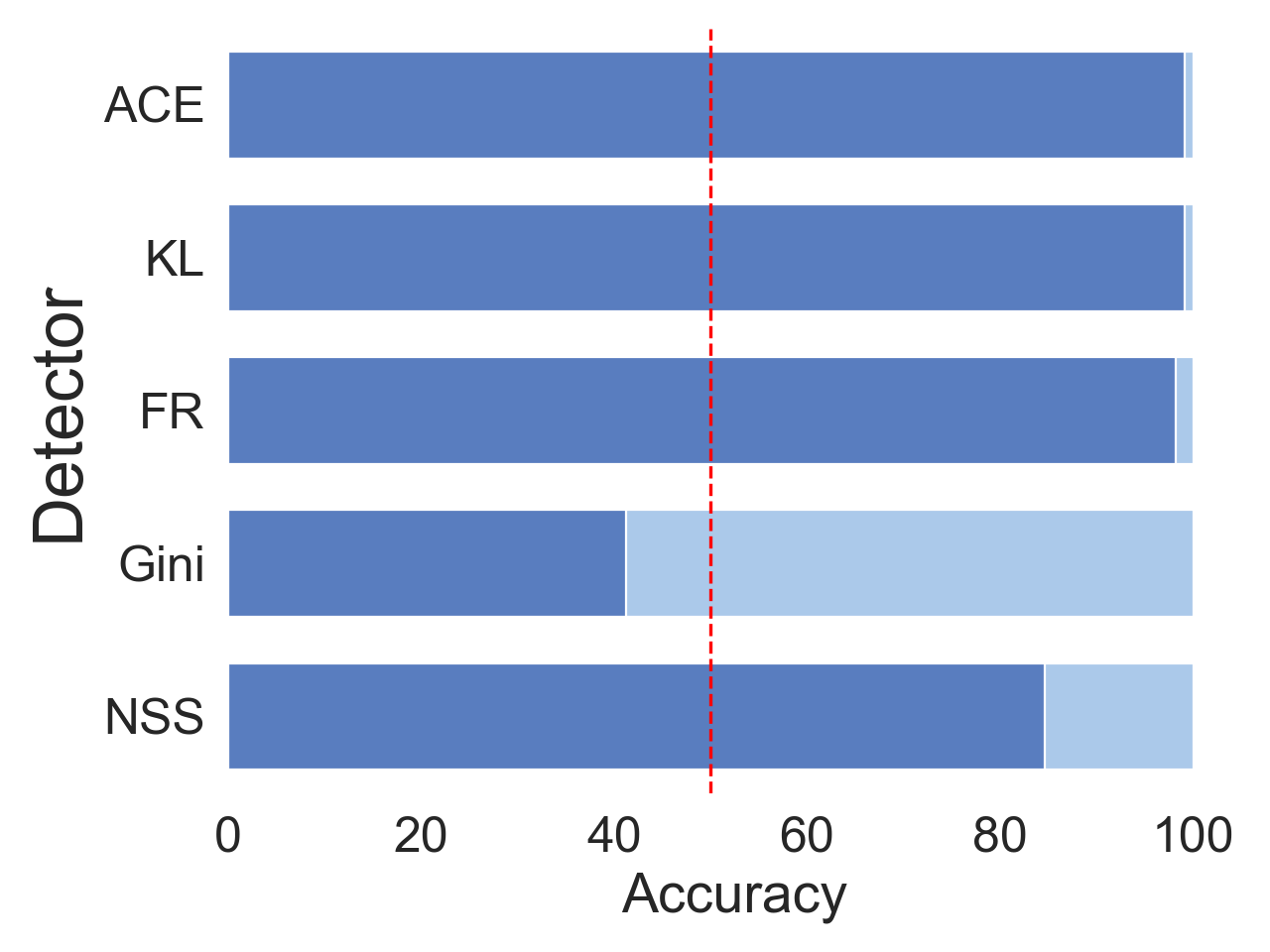}
		\vspace{-1.5\baselineskip}
		\caption{Attack loss: ACE}
		\label{fig:acc_ACE_frm}
	\end{subfigure}
	\hfill
	\begin{subfigure}[b]{0.5\columnwidth}
		\centering
		\includegraphics[width=\columnwidth]{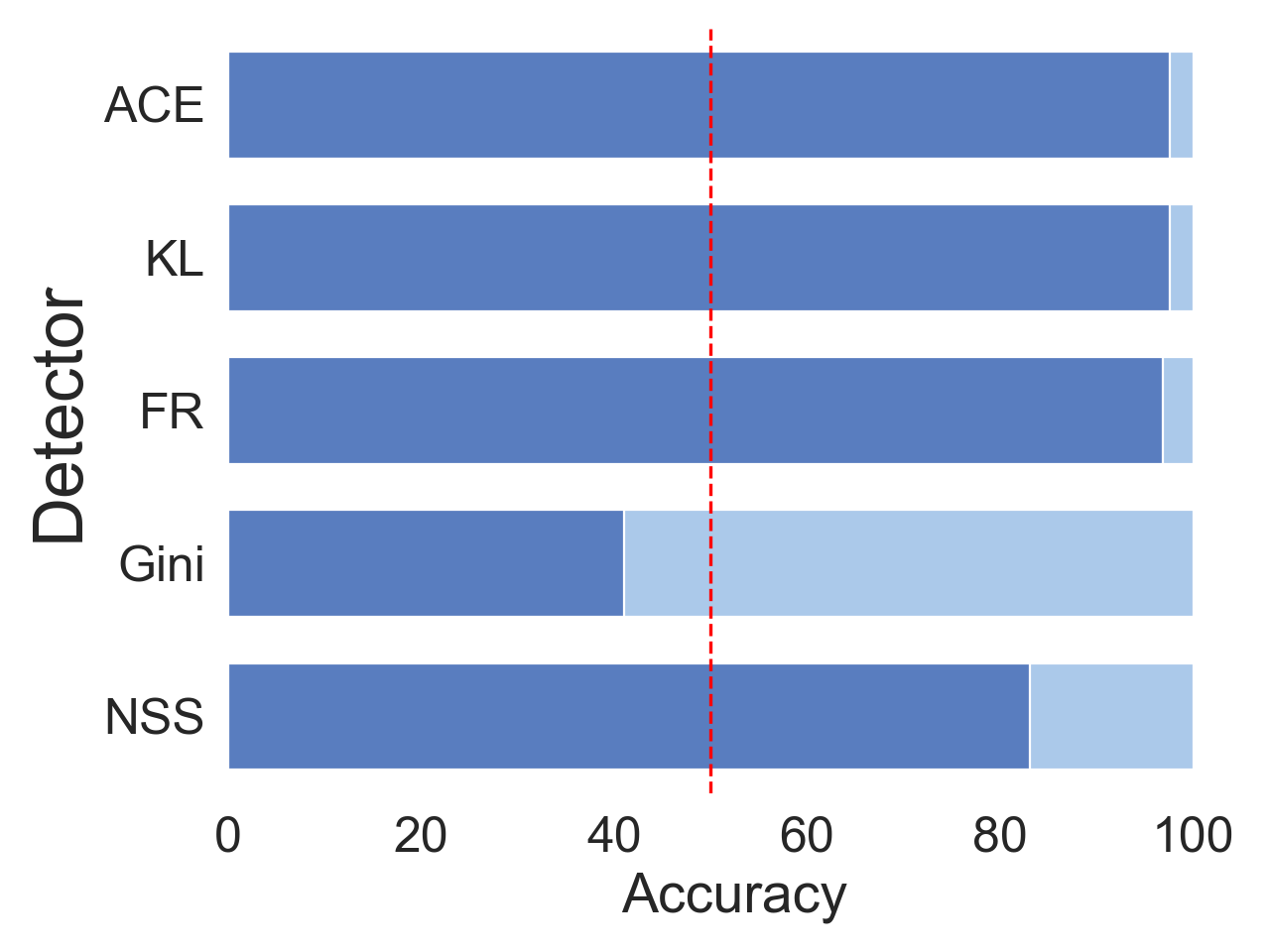}
		\vspace{-1.5\baselineskip}
		\caption{Attack loss: KL}
		\label{fig:acc_KL_frm}
	\end{subfigure}
        \hfill
	\begin{subfigure}[b]{0.5\columnwidth}
		\centering
		\includegraphics[width=\columnwidth]{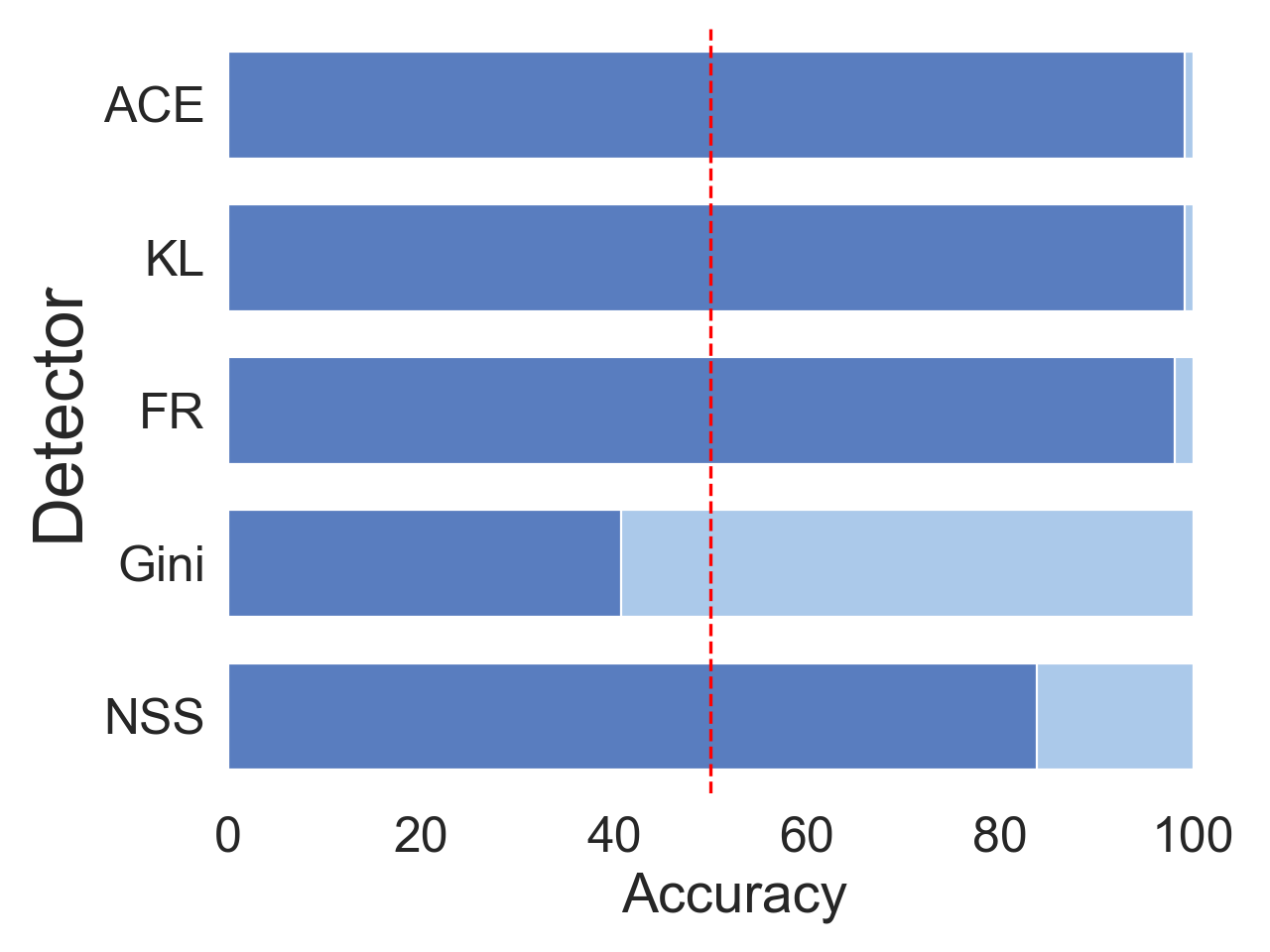}
		\vspace{-1.5\baselineskip}
		\caption{Attack loss: FR}
		\label{fig:acc_FR_frm}
	\end{subfigure}
        \hfill
	\begin{subfigure}[b]{0.5\columnwidth}
		\centering
		\includegraphics[width=\columnwidth]{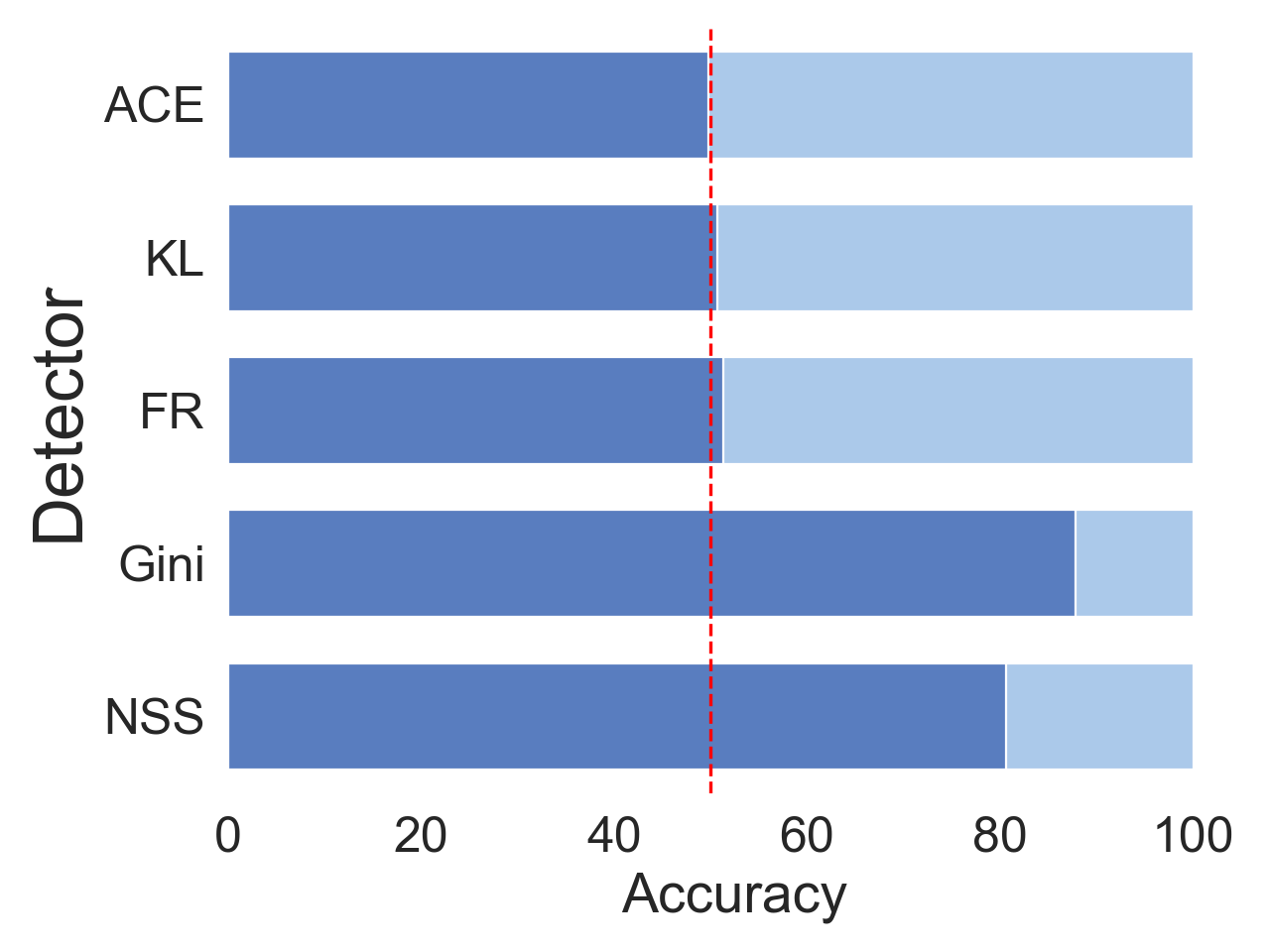}
		\vspace{-1.5\baselineskip}
		\caption{Attack loss: Gini}
		\label{fig:acc_Gini_frm}
	\end{subfigure}
        \hfill
 	\begin{subfigure}[b]{0.5\columnwidth}
		\centering
		\includegraphics[width=\columnwidth]{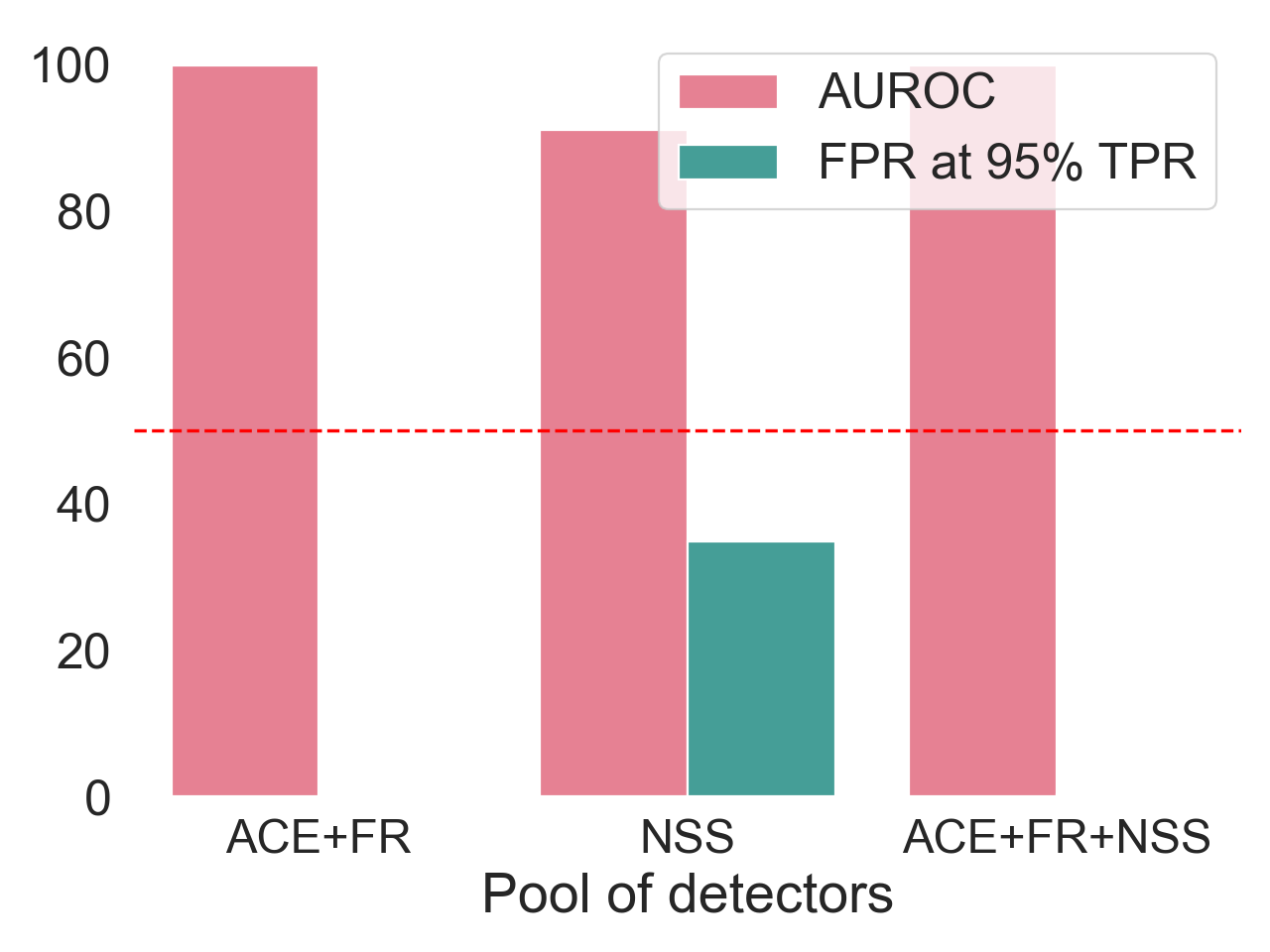}
		\vspace{-1.5\baselineskip}
		\caption{Attack losses: ACE and FR}
		\label{fig:metric_1}
	\end{subfigure}
        \hfill
 	\begin{subfigure}[b]{0.5\columnwidth}
		\centering
		\includegraphics[width=\columnwidth]{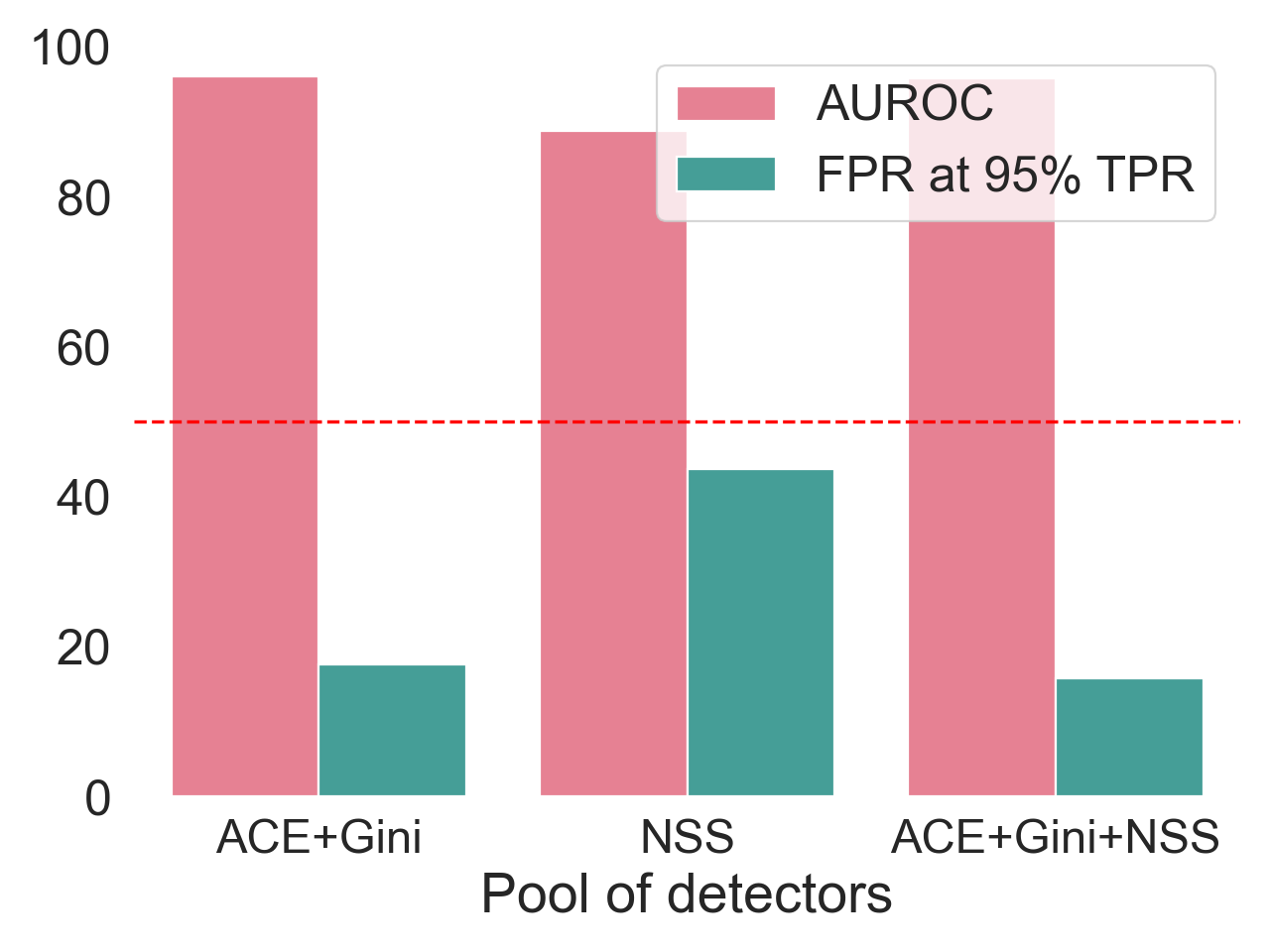}
		\vspace{-1.5\baselineskip}
		\caption{Attack losses: ACE and Gini}
		\label{fig:metric_2}
	\end{subfigure}
         \hfill
 	\begin{subfigure}[b]{0.5\columnwidth}
		\centering
		\includegraphics[width=\columnwidth]{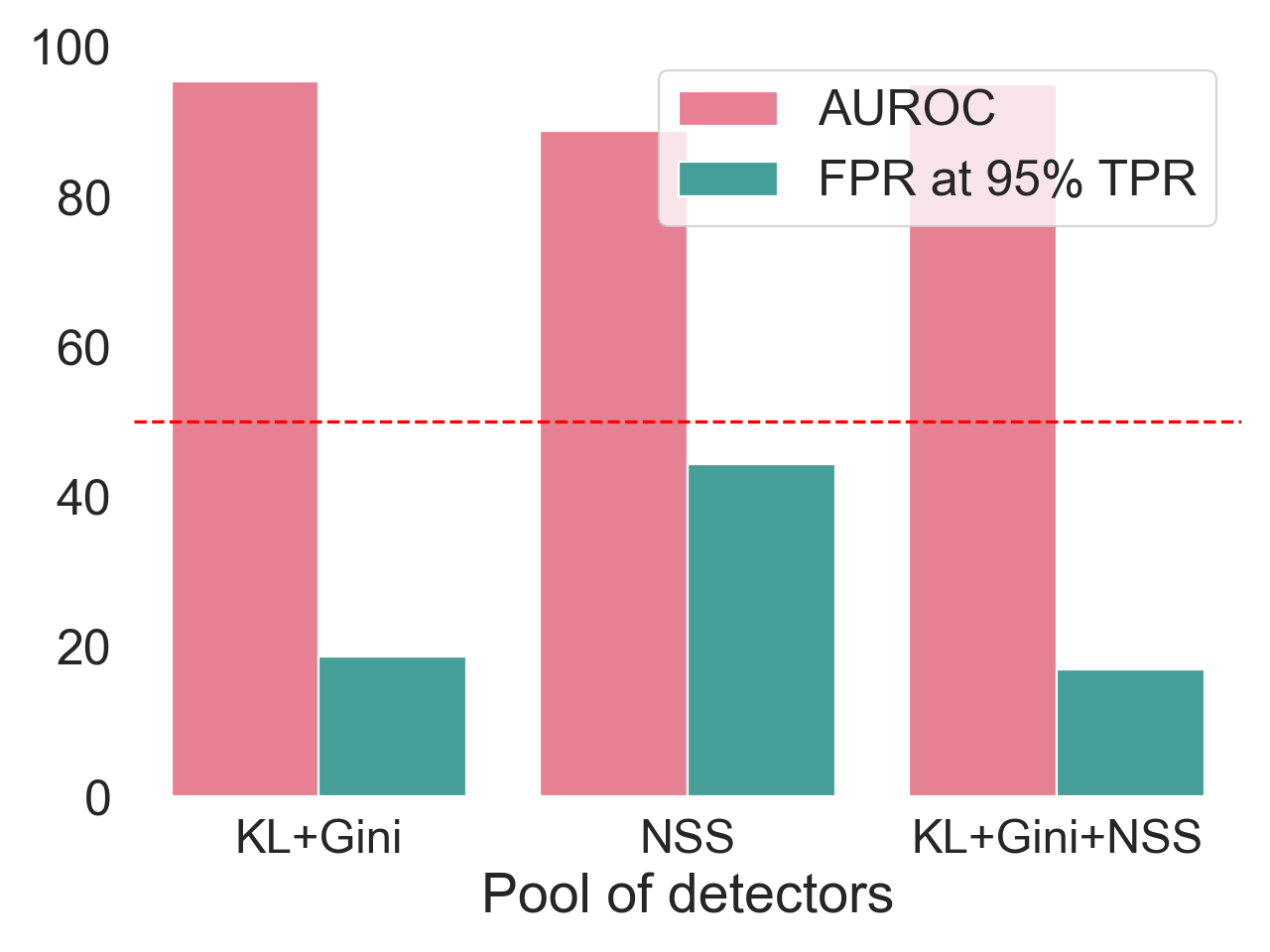}
		\vspace{-1.5\baselineskip}
		\caption{Attack losses: KL and Gini}
		\label{fig:metric_3}
	\end{subfigure}
         \hfill
 	\begin{subfigure}[b]{0.5\columnwidth}
		\centering
		\includegraphics[width=\columnwidth]{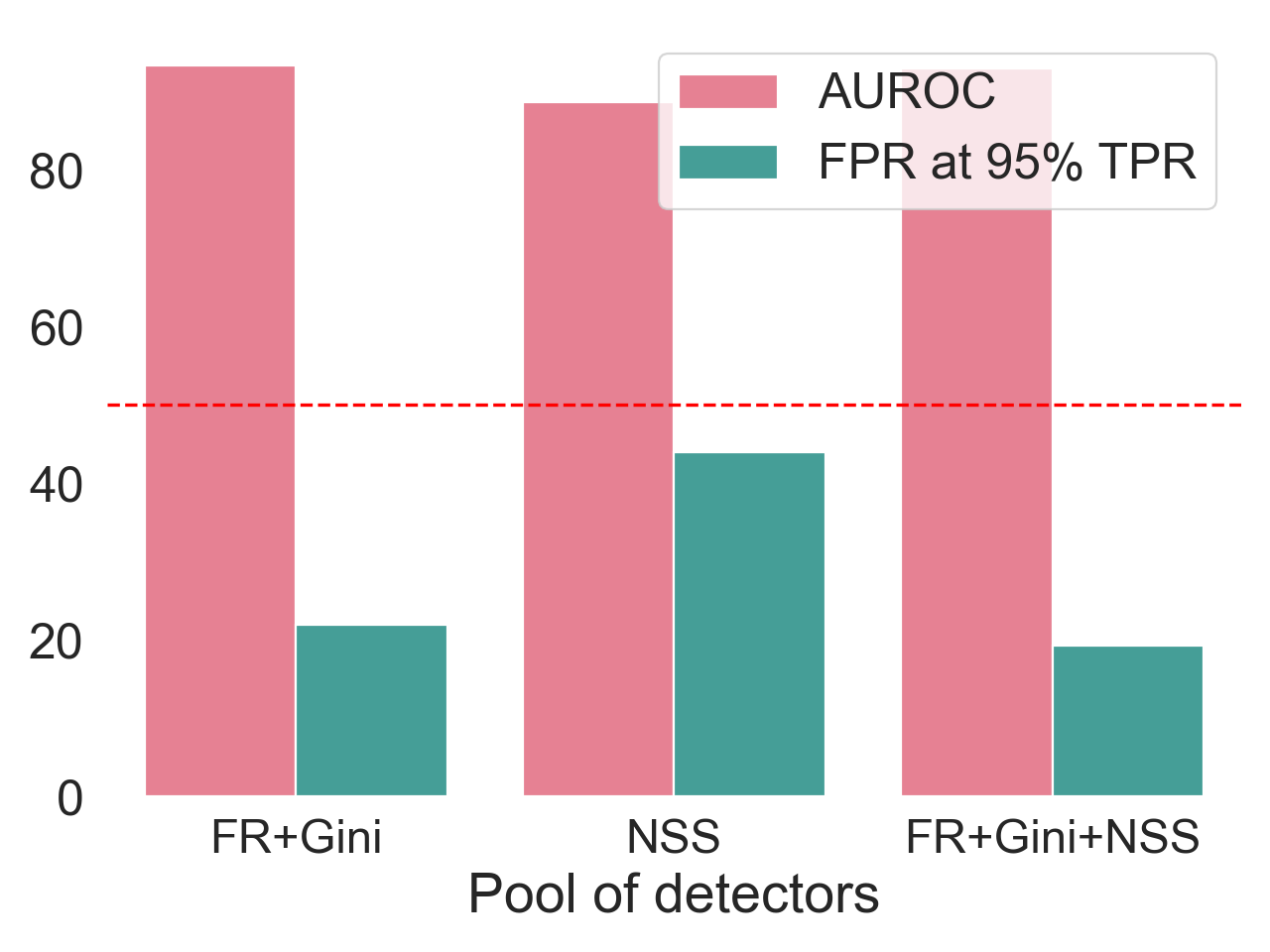}
		\vspace{-1.5\baselineskip}
		\caption{Attack losses: FR and Gini}
		\label{fig:metric_4}
	\end{subfigure}
 \caption{Attacks created with L$_\infty$ norm, PGD and $\varepsilon=0.03125$. The attack losses are given in the captions. The \textit{shallow} detectors are named after the loss function used to craft the attacks they are trained to detect.}
\label{fig:framework_th}
\end{figure*}
\label{sec:res_th_framework}
In~\cref{fig:framework_th} we present an illustrative example on \underline{CIFAR10} of the aggregator behavior in the optimal setting, i.e., when there each detector in the considered set is trained to detect at least one strategy in the multi-arm scheme. This example considers attacks created using the PGD algorithm with L$_\infty$ norm and $\varepsilon=0.03125$, and ACE loss in~\cref{fig:acc_ACE_frm}, KL loss in~\cref{fig:acc_KL_frm}, FR loss in~\cref{fig:acc_FR_frm}, and Gini loss in ~\cref{fig:acc_Gini_frm}. Note that, in the figures, the \textit{shallow} detectors are named after the loss function used to craft the attacks they are trained to detect.
As we operate within the optimal theoretical framework, in the figures when we mention the pool of detectors consisting of, for example, ACE+KL, it implies that multi-armed attacks are also created with ACE and KL losses. 
In~\cref{fig:metric_2,fig:metric_3,fig:metric_4},
even with just two detectors—one optimized for recognizing one specific type of attack and behaving poorly on the remaining one—we consistently observe an improvement in aggregator performance compared to the individual detector (i.e. NSS), which performs moderately well on both attacks. 
Moreover, in~\cref{fig:metric_1}, we consider the setting in which all the detectors in the pool are able to detect each other attacks. Not surprisingly, the combination of the decisions brings positive results.
These scenarios highlight the effectiveness of using a combination of specialized detectors, as opposed to relying on a single generalized detector for all attacks. 
Due to space limitations, the complete tables of results and those for \underline{SVHN}, are in~\cref{app:res_th_framework}.

\subsubsection{Evaluation of the Proposed Solution in the Setting of~\cite{GranesePRMP2022ECMLPKDD}}
\label{sec:res_mead}
We continue our evaluation by considering the \textsc{Mead} setup in~\cite{GranesePRMP2022ECMLPKDD} where the assumptions of the optimal setting are not met. The numerical results are provided in~\cref{app:experiments} (\cref{tab:final_table}) for space reasons.
On \underline{CIFAR10}, our aggregator achieves the maximum AUROC improvement of 79.5 percentage points compared to NSS. This improvement occurs for attacks under $L_{\infty}$-norm constraint, $\varepsilon = 0.125$ and PGD$^\star$, FGSM$^\star$, BIM$^\star$, SA, i.e. when as many as 13 different simultaneous adversarial attacks are mounted. 
Similarly, for our proposed method the maximum attained FPR at 95\% TPR improvement w.r.t. NSS is 90.3 percentage points and happens for attacks under $L_{\infty}$-norm constraint, $\varepsilon = 0.5$ and PGD$^\star$, FGSM$^\star$, BIM$^\star$, i.e. when as many as 12 different simultaneous adversarial attacks are mounted.
Our aggregator outperforms NSS in the case of the attacks with L$_1$ and L$_2$ norm, regardless of the algorithm or the perturbation magnitude, and in the case of L$_{\infty}$ norm with large perturbations.
However, for the attacks with L$_\infty$ norm and small $\varepsilon$, although the proposed method's performance is comparable to that of NSS, we notice a slight degradation. To shed light on this, we remind that individual detectors aggregated are based on the classifier's logits; NSS, on the other hand, extracts natural scene statistics from the inputs. This more sophisticated technique makes NSS perform well when tested on attacks with similar $\varepsilon$ and the same norm as the ones seen at training time. Additionally, it is essential to emphasize that the evaluation setting considered in the following analysis introduces additional challenges compared to the theoretical framework. In this evaluation, there may be cases where none of the detectors are optimal for the attacks under consideration. While this adds complexity, it reflects common real-life scenarios. Similar conclusions can be drawn for the results on \underline{SVHN} (cf.~\cref{tab:final_table}). 
In~\cref{app:res_mead_ablation} we conduct an in-depth analysis of the ablation study for this specific evaluation setup.

\section{CONCLUDING REMARKS}
\label{sec:conclusion}
We introduce a zero-shot method to detect attacks mounted against a target channel by a multi-armed malicious actor that can alter the input signal in an adversarial way. We characterized our theoretically optimal soft-detector which aggregates the decision of detection mechanisms, whose individual performance are poor in the multi-armed attack scenario, as the solution to a minimax problem where the defense minimizes the maximum risk due to the multi-armed attacker. Our empirical results, tailored to the domain of adversarial example detection, show that aggregating simple detectors using our method results in consistently improved detection performance. The achieved performance is comparable and in a large set of cases better than the best state-of-the-art (SOTA) method in the multi-armed attack scenarios. Our method has two key benefits: it is modular, allowing existing and future methods to be integrated (whether they are supervised or unsupervised), and it is general, able to recognize adversarial examples from various attack algorithms and loss functions. 
In general, adaptive attacks mounted on individual detectors are an argument in favor of other approaches to the problem (cf.~\cref{app:adaptive}).
We argue that, in our framework, we can rely on the effect redundancy to make sure that the task of mounting adaptive attacks on multiple detectors, possibly effective against the same attack strategy, becomes expensive or, at least, prohibitively expensive for an attacker. We leave the exploration of this aspect to future work.
Finally, our work to be applicable to other problems, such as intrusion, anomaly, out-of-distribution, or backdoor detection.


\subsubsection*{Acknowledgments}
    The work of Federica Granese was supported by the ANR-20-CE17-0022 DeepECG4U funding from the French National Research Agency.

\bibliographystyle{apalike}
\bibliography{biblio}
\clearpage
\clearpage
\section*{Checklist}



 \begin{enumerate}

 \item For all models and algorithms presented, check if you include:
 \begin{enumerate}
   \item A clear description of the mathematical setting, assumptions, algorithm, and/or model. Yes (\cref{sec:math_framework,sec:optimal_objective})
   \item An analysis of the properties and complexity (time, space, sample size) of any algorithm. Yes (\cref{app:enviroment})
   \item (Optional) Anonymized source code, with specification of all dependencies, including external libraries. Yes (Supplementary materials)
 \end{enumerate}

 \item For any theoretical claim, check if you include:
 \begin{enumerate}
   \item Statements of the full set of assumptions of all theoretical results. Yes (\cref{sec:math_framework,sec:optimal_objective})
   \item Complete proofs of all theoretical results. Yes (\cref{app:proofs})
   \item Clear explanations of any assumptions. Yes (\cref{sec:math_framework,sec:optimal_objective})   
 \end{enumerate}

 \item For all figures and tables that present empirical results, check if you include:
 \begin{enumerate}
   \item The code, data, and instructions needed to reproduce the main experimental results (either in the supplemental material or as a URL). Code available at \url{https://github.com/fgranese/Optimal-Zero-Shot-Detector-for-Multi-Armed-Attacks}.
   \item All the training details (e.g., data splits, hyperparameters, how they were chosen). Yes (\cref{sec:eval_framework,app:experiments})
         \item A clear definition of the specific measure or statistics and error bars (e.g., with respect to the random seed after running experiments multiple times). Not Applicable
         \item A description of the computing infrastructure used. (e.g., type of GPUs, internal cluster, or cloud provider). Yes (\cref{app:enviroment})
 \end{enumerate}

 \item If you are using existing assets (e.g., code, data, models) or curating/releasing new assets, check if you include:
 \begin{enumerate}
   \item Citations of the creator If your work uses existing assets. Yes (for CIFAR10 and SVHN datasets,~\cref{sec:eval_framework})
   \item The license information of the assets, if applicable. Not Applicable (Open source)
   \item New assets either in the supplemental material or as a URL, if applicable. Not Applicable
   \item Information about consent from data providers/curators. Not Applicable
   \item Discussion of sensible content if applicable, e.g., personally identifiable information or offensive content. Not Applicable
 \end{enumerate}

 \item If you used crowdsourcing or conducted research with human subjects, check if you include:
 \begin{enumerate}
   \item The full text of instructions given to participants and screenshots. Not Applicable
   \item Descriptions of potential participant risks, with links to Institutional Review Board (IRB) approvals if applicable. Not Applicable
   \item The estimated hourly wage paid to participants and the total amount spent on participant compensation. Not Applicable
 \end{enumerate}

 \end{enumerate}
 \appendix

%
%





%

%

\onecolumn
\aistatstitle{Supplementary Materials}
\section{SUPPLEMENTARY DETAILS ON SEC.~\ref{sec:optimal_objective}}
\subsection{Proofs}
\label{app:proofs}
\subsubsection{Proof of~\cref{eq-missing}}
\label{app:proof1}
\begin{proof}{}
\begin{align*}
\max_{k\in\calK}\, \EE_{q^{(k)}_{\widehat{Z}| \mathbf{u_0}}}\left[ -\log {q}_{\widehat{Z}|\mathbf{u_0}}  \right] &=
\max_{k\in\calK}\, \left[\EE_{q^{(k)}_{\widehat{Z}| \mathbf{u_0}}}\left[ -\log q^{(k)}_{\widehat{Z}| \mathbf{u_0}} \right] + \EE_{q^{(k)}_{\widehat{Z}| \mathbf{u_0}}}\left[\log\left(\frac{q^{(k)}_{\widehat{Z}|\mathbf{u_0}}}{{q}_{\widehat{Z}|\mathbf{u_0}}}\right)\right]\right]\\
&\leq \max_{k\in\calK}\, \EE_{q^{(k)}_{\widehat{Z}| \mathbf{u_0}}}\left[-\log q^{(k)}_{\widehat{Z}| \mathbf{u_0}} \right] + \max_{k\in\calK}\, \EE_{q^{(k)}_{\widehat{Z}| \mathbf{u_0}}}\left[\log\left(\frac{q^{(k)}_{\widehat{Z}|\mathbf{u_0}}}{{q}_{\widehat{Z}|\mathbf{u_0}}}\right)\right].     
\end{align*}
\end{proof}

\subsubsection{Proof of~\cref{eq:minimaxProb1}}
\label{app:proof2}
\begin{proof}
The equality holds by noticing that 
\begin{align*}
  \max_{P_{\Omega}}&\, \EE_{\Omega}\left[D_{\textrm{KL}}\left(q^{(\Omega)}_{\widehat{Z}|\mathbf{u_0}}\big \| {q}_{\widehat{Z}|\mathbf{u_0}}\right)\right] 
  \leq
  \max_{k\in\calK}\, \EE_{q^{(k)}_{\widehat{Z}| \mathbf{u_0}}}\left[\log\left(\frac{q^{(k)}_{\widehat{Z}|\mathbf{u_0}}}{{q}_{\widehat{Z}|\mathbf{u_0}}}\right)\right],
\end{align*}
and moreover, 
\begin{align*}
    \max_{k\in\calK}\, \EE_{q^{(k)}_{\widehat{Z}| \mathbf{u_0}}}\left[\log\left(\frac{q^{(k)}_{\widehat{Z}|\mathbf{u_0}}}{{q}_{\widehat{Z}|\mathbf{u_0}}}\right)\right]& = \EE_{\bar{\Omega}}\left[D_{\textrm{KL}}\left(q^{(\bar{\Omega})}_{\widehat{Z}|\mathbf{u_0}}\big \| {q}_{\widehat{Z}|\mathbf{u_0}}\right)\right],
\end{align*}
by choosing the random variable $\bar{\Omega}$ with uniform probability over the set of maximizers  ${\overline{\mathcal{K}}=\arg\max_{k\in\calK}\, \EE_{q^{(k)}_{\widehat{Z}| \mathbf{u_0}}}\left[\log\left(\frac{q^{(k)}_{\widehat{Z}|\mathbf{u_0}}}{{q}_{\widehat{Z}|\mathbf{u_0}}}\right)\right]}$, zero otherwise. 
\end{proof}

\subsubsection{Proof of~\cref{eq:minimaxProb3}}
\label{app:proof3}
\begin{proof}
We consider a zero-sum game with a concave-convex mapping defined on a product of convex sets. The sets of all probability distributions ${q}_{\widehat{Z}|\mathbf{u_0}}$ and $P_\Omega$ are two nonempty convex sets, bounded and finite-dimensional. On the other hand, $\big(P_\Omega,{q}_{\widehat{Z}|\mathbf{u_0}} \big)\rightarrow  \EE_{\Omega}\left[D_{\textrm{KL}}\left(q^{(\Omega)}_{\widehat{Z}|\mathbf{u_0}}\big \| {q}_{\widehat{Z}|\mathbf{u_0}}\right)\right]$ is a concave-convex mapping, i.e., $P_\Omega\rightarrow  \EE_{\Omega}\left[D_{\textrm{KL}}\left(q^{(\Omega)}_{\widehat{Z}|\mathbf{u_0}}\big \| {q}_{\widehat{Z}|\mathbf{u_0}}\right)\right]$  is concave and ${q}_{\widehat{Z}|\mathbf{u_0}} \rightarrow  \EE_{\Omega}\left[D_{\textrm{KL}}\left(q^{(\Omega)}_{\widehat{Z}|\mathbf{u_0}}\big \| {q}_{\widehat{Z}|\mathbf{u_0}}\right)\right]$ is convex for every $\big(P_\Omega,{q}_{\widehat{Z}|\mathbf{u_0}} \big)$.  Then, by classical min-max theorem ~\cite{vonNeumann1928-VONZTD-2} we have that~\cref{eq:minimaxProb3} holds. 
\end{proof}

\subsubsection{Proof of~\cref{eq:minimaxProb4}}
\label{app:proof4}
\begin{proof}
It is enough to show that 
\begin{align}
\min_{\widehat{q}_{\widehat{Z}|\mathbf{u_0}}}\EE_{\Omega}\left[D_{\textrm{KL}}\left(q^{(\Omega)}_{\widehat{Z}|\mathbf{u_0}}\big \| {q}_{\widehat{Z}|\mathbf{u_0}}\right)\right] = I_{\mathbf{u_0}}(\Omega;\widehat{Z}), \label{eq-missing-appex}
    \end{align}
for every random variable $\Omega$ distributed according to an arbitrary probability distribution $P_{\Omega}$ and each distribution $q^{(\Omega)}_{\widehat{Z}|\mathbf{u_0}}$. We begin by showing that 
\begin{align*}
\EE_{\Omega}\left[D_{\textrm{KL}}\left(q^{(\Omega)}_{\widehat{Z}|\mathbf{u_0}}\big \| {q}_{\widehat{Z}|\mathbf{u_0}}\right)\right] & \geq  I_{\mathbf{u_0}}(\Omega;\widehat{Z}), 
    \end{align*}
for any arbitrary distributions $P_{\Omega}$ and   $q^{(\Omega)}_{\widehat{Z}|\mathbf{u_0}}$. To this end, we use the following identities: 
\begin{align}
\EE_{\Omega}\left[D_{\textrm{KL}}\left(q^{(\Omega)}_{\widehat{Z}|\mathbf{u_0}}\big \| {q}_{\widehat{Z}|\mathbf{u_0}}\right)\right]
&=\EE_{\Omega}\EE_{ q^{(\Omega)}_{\widehat{Z}|\mathbf{u_0}}}\left(\log\frac{q^{(\Omega)}_{\widehat{Z}|\mathbf{u_0}}}{ {q}_{\widehat{Z}|\mathbf{u_0}}}\right)\nonumber\\
&= \EE_{\Omega}\EE_{ q^{(\Omega)}_{\widehat{Z}|\mathbf{u_0}}}\left(\log\frac{q^{(\Omega)}_{\widehat{Z}|\mathbf{u_0}}}{P_{\widehat{Z}} }\right) + \nonumber D_{\textrm{KL}}\left( P_{\widehat{Z}}  \| {q}_{\widehat{Z}|\mathbf{u_0}} \right)\nonumber\\
&= I_{\mathbf{u_0}}(\Omega;\widehat{Z}) + D_{\textrm{KL}}\left( P_{\widehat{Z}}  \| {q}_{\widehat{Z}|\mathbf{u_0}} \right)\geq  I_{\mathbf{u_0}}(\Omega;\widehat{Z}), \label{eq-missing-appex-B}
\end{align}
where $P_{\widehat{Z}}$ denotes the marginal distribution of $q^{(\Omega)}_{\widehat{Z}|\mathbf{u_0}}$ w.r.t. $P_{\Omega}$ and the last inequality follows since the KL divergence is positive. Finally, it is easy to check that by selecting $ {q}_{\widehat{Z}|\mathbf{u_0}} = P_{\widehat{Z}} $ the lower bound in \eqref{eq-missing-appex-B} is achieved which proves the identity in expression \eqref{eq-missing-appex}. By taking the maximum overall probability distributions  $P_{\Omega}$ at both sides of expression \eqref{eq-missing-appex} the claim follows. 
\end{proof}

\subsubsection{Proof of the Result in~\cref{sec:domain_shift}}
\label{app:proof_thm}
This proof is adapted from~\cite{Ben-DavidBCKPV2010ML}.
\begin{proof}
    \begin{align}
        P_e^{\newdomain}(\textsc{D})&=  P_e^{\newdomain}(\textsc{D}) +P_e^{S}(\textsc{D}) -P_e^{S}(\textsc{D}) + \EE_{\x\sim P^{S}_{X}}\left[\mathds{1}\left[\textsc{D}\left(\x\right)\neq f^{T}(\x)\right]\right]-\EE_{\x\sim P^{S}_{X}}\left[\mathds{1}\left[\textsc{D}\left(\x\right)\neq f^{T}(\x)\right]\right]\\
        &\leq P_e^{S}(\textsc{D}) + \left|\EE_{\x\sim P^{S}_{X}}\left[\mathds{1}\left[\textsc{D}\left(\x\right)\neq f^{T}(\x)\right]\right]-P_e^{S}(\textsc{D})\right|+\left|P_e^{\newdomain}(\textsc{D})-\EE_{\x\sim P^{S}_{X}}\left[\mathds{1}\left[\textsc{D}\left(\x\right)\neq f^{T}(\x)\right]\right]\right|\\
        &\leq P_e^{S}(\textsc{D}) + \EE_{\x\sim P^{S}_{X}}\left|\mathds{1}\left[\textsc{D}\left(\x\right)\neq f^{T}(\x)\right]-\mathds{1}\left[\textsc{D}\left(\x\right)\neq f^{S}(\x)\right]\right|+\left|P_e^{\newdomain}(\textsc{D})-\EE_{\x\sim P^{S}_{X}}\left[\mathds{1}\left[\textsc{D}\left(\x\right)\neq f^{T}(\x)\right]\right]\right|\\
        &\leq P_e^{S}(\textsc{D}) + \EE_{\x\sim P^{S}_{X}}\left|f^{T}(\x)-f^{S}(\x)\right|+\left|P_e^{\newdomain}(\textsc{D})-\EE_{\x\sim P^{S}_{X}}\left[\mathds{1}\left[\textsc{D}\left(\x\right)\neq f^{T}(\x)\right]\right]\right|\\
        &\leq P_e^{S}(\textsc{D}) + \EE_{\x\sim P^{S}_{X}}\left|f^{T}(\x)-f^{S}(\x)\right|+d\left(P^{S}_{X|Z=1},P^{\newdomain}_{X|Z=1}\right).
    \end{align}
Notice that by choosing to add and subtract $\EE_{\x\sim P^{T}_{X}}\left[\mathds{1}\left[\textsc{D}\left(\x\right)\neq f^{T}(\x)\right]\right]$ instead of $\EE_{\x\sim P^{S}_{X}}\left[\mathds{1}\left[\textsc{D}\left(\x\right)\neq f^{T}(\x)\right]\right]$, we would get the term $\EE_{\x\sim P^{T}_{X}}\left|f^{T}(\x)-f^{S}(\x)\right|$, instead of $\EE_{\x\sim P^{S}_{X}}\left|f^{T}(\x)-f^{S}(\x)\right|$. Therefore, the final result holds true:
\begin{align}
    \nonumber
    P_e^{\newdomain} (\textsc{D}) &\leq P_e^{S}(\textsc{D}) + d\left(P^{S}_{X|Z=1},P^{\newdomain}_{X|Z=1}\right)
    \\
    &+ \min\big\{\EE_{
    \x\sim P^{S}_{X}}[|f^{S}(\x)-f^{\newdomain}(\x)|],\nonumber 
    \\
    &\EE_{\x\sim P^{\newdomain}_X}[|f^{S}(\x)-f^{\newdomain}(\x)|]\big \}.
\end{align}
\end{proof}

\subsection{On the Optimization of~\cref{eq:minimaxProb4}}
\label{app:optimization}
The maximization problem in~\cref{eq:minimaxProb4} is well-posed given that the mutual information is a concave function of $\omega\in\Omega$.
Although from the theoretical point of view,~\cref{eq:minimaxProb4} guarantees the optimal solution for the average regret minimization problem, in practice, we have to deal with some technical limitations.

For the optimization of ~\cref{eq:minimaxProb4}, we rely on the implementation of the \textit{Blahut-Arimoto algorithm}~\cite{Arimoto72}. This is an iterative algorithm for finding the capacity, $C$, of a channel that in our context corresponds to objective in~\cref{eq:minimaxProb4} i.e., $C = \underset{w_k}{\max}\,I_{\mathbf{u}_0}(\Omega; \widehat{Z})$. Therefore, at iteration $t+1$ the weights $\omega^{t+1}_k$ to associate with the decision made by the detector $k$ will be computed as follows
\begin{align}
    \Tilde{\omega}^{t+1}_k(z) &= \frac{\omega^t_k \cdot (q^{(k)}_{\widehat{Z}=z|\mathbf{u}_0})}{\sum_{k=0}^K \, \omega^t_k \cdot (q^{(k)}_{\widehat{Z}=z|\mathbf{u}_0})}\\
    \omega^{t+1}_k &= 
    \frac{\prod_{z\in\{0, 1\}} \Tilde{\omega}^{t+1}_k(z)^{(q^{(k)}_{\widehat{Z}=z|\mathbf{u}_0})}}
    {\sum_{k=0}^K \prod_{z\in\{0, 1\}} \Tilde{\omega}^{t+1}_k(z)^{(q^{(k)}_{\widehat{Z}=z|\mathbf{u}_0})}},
\end{align}
where $K$ denotes the number of detectors to aggregate.
Overall, we obtained the satisfactory results provided in the paper by assigning default values to all the parameters and by setting a uniform distribution $[\omega_1, \omega_2, \omega_3, \omega_4]=[.25, .25, .25, .25]$ as the initial point in the solutions space.


\section{ADDITIONAL EXPERIMENTS}
\label{app:experiments}
\subsection{Experimental Environment and Time Measurements}
\label{app:enviroment}
We run each experiment on a machine equipped with an Intel(R) Xeon(R) Gold 6226 CPU, 2.70GHz clock frequency, and a Tesla V100-SXM2-32GB GPU.
\begin{table}[!ht]
    \centering
\resizebox{.7\columnwidth}{!}{%
\begin{tabular}{r|c}
\toprule
     Training 1 single detector in our method &  \texttt{1h45m10s}\\
     Evaluating the optimization in our method & \texttt{5s} (for one attack)\\
     \midrule
     Training NSS & \texttt{3m30s}\\
     Evaluating NSS & \texttt{20s} (for one attack)\\
     \midrule
     \midrule

     On the largest set of simultaneous attacks (13
     attacks):\\
     
     Ours & \texttt{5s * 13 $\sim$ 65s}\\
     NSS & \texttt{20s * 13 $\sim$ 4m}\\
     \bottomrule
\end{tabular}
}
\label{tab:table_measurements}
\end{table}
\subsection{State-of-the-Art (SOTA) Detectors}
\label{app:sota}
\begin{wrapfigure}{r}{9cm}
\vspace{-1.7cm}
    \centering
    \includegraphics[width=.45\columnwidth]{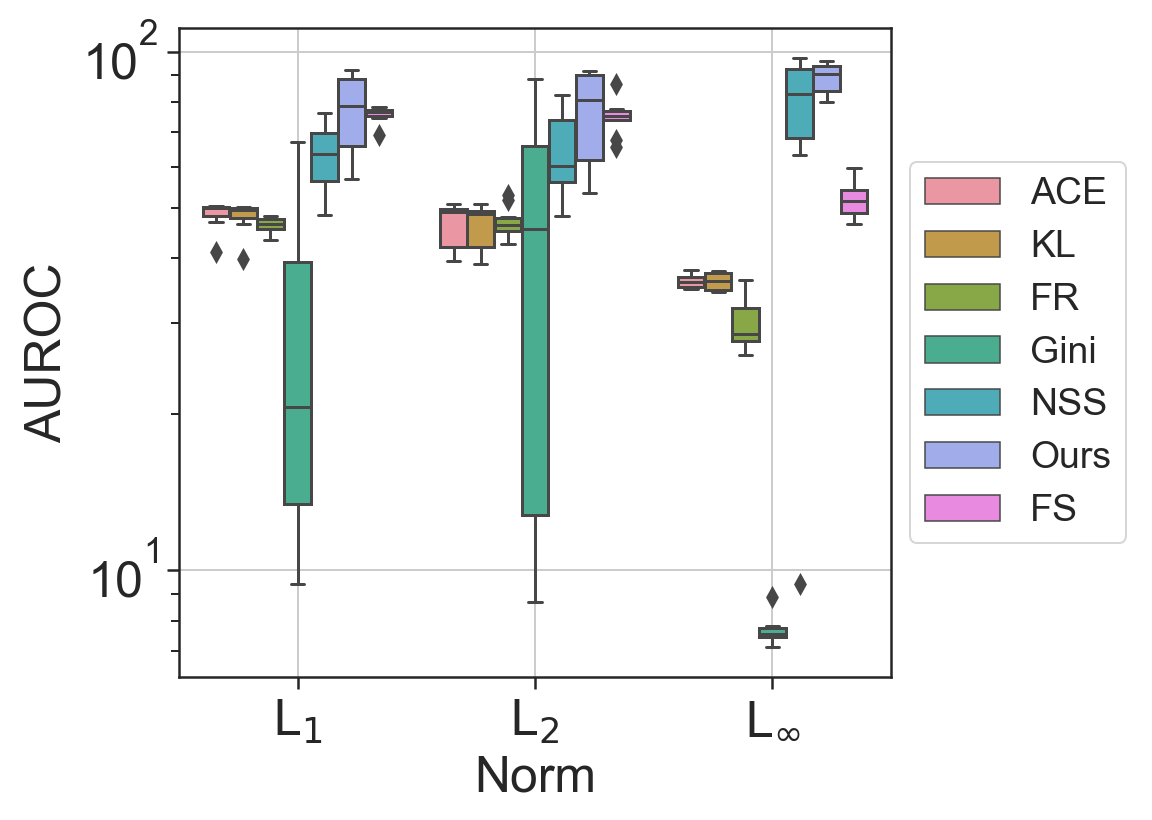}
    \caption{The \textit{shallow} detectors are named after the loss function used to craft the attacks they are trained to detect.}
    \label{fig:box_plot_fs}
\end{wrapfigure}
\cite{GranesePRMP2022ECMLPKDD} suggests NSS~\cite{NSS} and FS~\cite{FS} as the most robust methods in the simultaneous attacks detection scheme (i.e., {\mead}). We remind that NSS is a supervised method that extracts the \textit{natural scene statistics} of the natural and adversarial examples to train a SVM. On the contrary,  FS is an unsupervised method that uses \textit{feature squeezing} (i.e.,
reducing the color depth of images and using smoothing to
reduce the variation among the pixels) to compare the model’s predictions.

In particular, we choose NSS as a method to compare for multiple reasons:

1. NSS achieves the best overall score in terms of {\auc} and {\fpr} among the SOTA against simultaneous attacks (cf. Tab. 3~\cite{GranesePRMP2022ECMLPKDD}).

2. NSS achieves the best score in terms of {\auc} and {\fpr} under the L$_\infty$ norm where the biggest group of simultaneous attacks are evaluated (see~\cref{tab:attacks}). This is stressed in the plots in~\cref{fig:box_plot_fs}. Moreover, FS reaches better performance w.r.t. the proposed method only with PGD1 and PGD2 when the perturbation magnitude is small and in CW2. 

3. The case study for our aggregator in the experimental section is based on supervised detectors as a consequence the comparison with a supervised detector was a natural choice.

For the sake of completeness, the performances of NSS and FS under {\mead} are given in~\cref{fig:box_plot_fs}.

\subsection{Attacks}
\label{app:attacks}
\begin{wraptable}{r}{9cm}
\vspace{-.5cm}
\centering
\caption{Each cell corresponds to attacks simultaneously executed on the targeted classifier. 
Attacks created using all the losses in~\cite{GranesePRMP2022ECMLPKDD} are marked with $^\star$.
Attacks such as SA and DF are not dependent on the choice for the loss, but are equally considered as part of the multi-armed framework.
Empty cells correspond to combinations of perturbation magnitude and norm constraint that are usually not considered in the literature.}
\ra{1.3}
\resizebox{.5\columnwidth}{!}{%
\begin{tabular}{@{}rd|d|d|d@{}}
\multicolumn{1}{c}{} & \multicolumn{1}{c}{L$_1$} & \multicolumn{1}{c}{L$_2$} & \multicolumn{1}{c}{L$_\infty$} & \multicolumn{1}{c}{No norm} \\ \cmidrule{2-5}
$\varepsilon=0.01$ & 
- & CW2 & - & -\\ \cmidrule{2-5}
$\varepsilon=0.03125$ & 
- & - & PGDi$^{\star}$,FGSM$^{\star}$,BIM$^{\star}$ & -\\ \cmidrule{2-5}
$\varepsilon=0.0625$ & 
- & - & PGDi$^{\star}$,FGSM$^{\star}$,BIM$^{\star}$ & -\\\cmidrule{2-5}
$\varepsilon=0.1$ & 
- & HOP & - & -\\ \cmidrule{2-5}
$\varepsilon=0.125$ & 
- & PGD2$^{\star}$& PGDi$^{\star}$,FGSM$^{\star}$,BIM$^{\star}$,SA & -\\ \cmidrule{2-5}
$\varepsilon=0.25$ &
- & PGD2$^{\star}$ & PGDi$^{\star}$,FGSM$^{\star}$,BIM$^{\star}$ & -\\\cmidrule{2-5}
$\varepsilon=0.3125$ & 
- & PGD2$^{\star}$& PGDi$^{\star}$,FGSM$^{\star}$,BIM$^{\star}$,CWi & -\\ \cmidrule{2-5}
$\varepsilon=0.5$ & 
- & PGD2$^{\star}$& PGDi$^{\star}$,FGSM$^{\star}$,BIM$^{\star}$ & -\\ \cmidrule{2-5}
$\varepsilon=1$ & 
- & PGD2$^{\star}$& - & -\\ \cmidrule{2-5}
$\varepsilon=1.5$ & 
- & PGD2$^{\star}$& - & -\\ \cmidrule{2-5}
$\varepsilon=2$ & 
- & PGD2$^{\star}$& - & -\\ \cmidrule{2-5}
$\varepsilon=5$ & 
PGD1$^{\star}$& - & - & -\\ \cmidrule{2-5}
$\varepsilon=10$ &
PGD1$^{\star}$& - & - & -\\ \cmidrule{2-5}
$\varepsilon=15$ & 
PGD1$^{\star}$& - & - & -\\ \cmidrule{2-5}
$\varepsilon=20$ & 
PGD1$^{\star}$& - & - & -\\ \cmidrule{2-5}
$\varepsilon=25$ & 
PGD1$^{\star}$& - & - & -\\ \cmidrule{2-5}
$\varepsilon=30$ &
PGD1$^{\star}$& - & - & -\\ \cmidrule{2-5}
$\varepsilon=40$ & 
PGD1$^{\star}$& - & - & -\\ 
\midrule
No $\varepsilon$ & 
- & DF & - & -\\ \cmidrule{2-5}
\begin{tabular}{r}
     max. rotation $=30$\\
     max. translation $=8$\\
\end{tabular} &
- & - & - & STA\\
\end{tabular}
}
\label{tab:attacks}
\end{wraptable}

We want to emphasize that, differently from the literature, we are the first to consider a defense mechanism against the simultaneous attack setting in which we detect attacks based on four different losses. More specifically, for each 'clean dataset' (in our case CIFAR10 and SVHN):
\begin{itemize}
    \item No. of adversarial examples generated with:
    \begin{itemize}
        \item L$_1$ norm: 7 (no. of $\varepsilon$) * 1 (PGD algorithm) * 4 (no. of losses) = 28 ('adversarial datasets')
        \item L$_2$ norm: 7 (no. of $\varepsilon$) * 1 (PGD algorithm) * 4 (no. of losses) + 3 (CW2, HOP, DeepFool) = 31 ('adversarial datasets')
        \item L$_\infty$ norm: 6 (no. of $\varepsilon$) * 3 (PGD, FGSM, BIM algorithms) * 4 (no. of losses) + 2 = 74 ('adversarial datasets')
        \item No norm: 1 ('adversarial dataset')
    \end{itemize}
    \item[$=>$] For a total of \texttt{28 + 31 + 74 + 1 = \textbf{134}} 'adversarial datasets' for each 'clean dataset'.
\end{itemize}
Moreover, it is interesting to notice that the experiments on CIFAR10 and SVHN represent a satisfying choice to show that state-of-the-art detection mechanisms struggle to maintain good performance when they are faced with the framework of simultaneous attacks. That said, we leave the evaluation of larger datasets as future work.

We provide in~\cref{tab:attacks} below how the attacks are grouped in the evaluation of the multi-armed setting

\subsection{Simulations Adversarial Attack According to Different $\varepsilon$}
\label{app:varius_eps}
As discussed in~\cref{sec:eval_framework}, both NSS and the \textit{shallow} detectors aggregated via the proposed method are trained on natural and adversarial examples created with PGD algorithm and L$_\infty$ norm constraint. We show in~\cref{tab:cifar10_nss,tab:cifar10_salad,tab:svhn_nss,tab:svhn_salad} the results of the two methods according to $\varepsilon\in\left\{.03125, .0625,.125, .25, .3125, .5\right\}$. 
\begin{table*}[!htbp]
\centering
\caption{Simultaneous attacks detection: NSS on CIFAR10. We train NSS on natural and adversarial examples created with PGD algorithm and L$_\infty$ norm constraint. The perturbation magnitude $\varepsilon$ is shown in the columns. We indicate in \textbf{bold} the best result.}
\resizebox{\columnwidth}{!}{%
\begin{tabular}{r|cc|cc|cc|cc|cc|cc}
\toprule
 &
  \multicolumn{12}{c}{\textbf{NSS}} \Bstrut \\ \cline{2-13} &
   
  \multicolumn{2}{c|}{0.03125} & \multicolumn{2}{c|}{0.0625}& \multicolumn{2}{c|}{0.125}& \multicolumn{2}{c|}{0.25} &
  \multicolumn{2}{c|}{0.3125} &   \multicolumn{2}{c}{0.5}
  \Tstrut\Bstrut \\ \cmidrule{2-13} 
  &
  \multicolumn{1}{c}{\auc} &
  \fpr &
  \multicolumn{1}{c}{\auc} &
  \fpr &
  \multicolumn{1}{c}{\auc} &
  \fpr &
  \multicolumn{1}{c}{\auc} &
  \fpr &
  \multicolumn{1}{c}{\auc} &
  \fpr &
  \multicolumn{1}{c}{\auc} &
  \fpr \\
  \cmidrule{2-13}
  \textbf{Norm L$_1$}\\\textcolor{gray}{PGD1}\\
  $\varepsilon$ = 5 &
\multicolumn{1}{c}{\textbf{48.5}} & \textbf{94.2} &
\multicolumn{1}{c}{47.7} & 94.7 &
\multicolumn{1}{c}{46.6} & 95.6 &
\multicolumn{1}{c}{46.8} & 95.5 &
\multicolumn{1}{c}{47.0} & 95.4 &
\multicolumn{1}{c}{46.5} & 95.6
 \\
$\varepsilon$ = 10 &
\multicolumn{1}{c}{\textbf{54.0}} & \textbf{90.3} &
\multicolumn{1}{c}{53.4} & 90.8 &
\multicolumn{1}{c}{51.6} & 94.3 &
\multicolumn{1}{c}{50.4} & 94.9 &
\multicolumn{1}{c}{50.4} & 94.9 &
\multicolumn{1}{c}{50.9} & 94.7
 \\
$\varepsilon$ = 15 &
\multicolumn{1}{c}{\textbf{58.8}} & \textbf{86.8} &
\multicolumn{1}{c}{58.1} & 87.4 &
\multicolumn{1}{c}{55.8} & 92.8 &
\multicolumn{1}{c}{53.8} & 94.2 &
\multicolumn{1}{c}{53.2} & 94.4 &
\multicolumn{1}{c}{54.5} & 93.7
 \\
$\varepsilon$ = 20 &
\multicolumn{1}{c}{\textbf{63.5}} & \textbf{82.3} &
\multicolumn{1}{c}{62.7} & 82.7 &
\multicolumn{1}{c}{60.1} & 90.7 &
\multicolumn{1}{c}{57.4} & 93.2 &
\multicolumn{1}{c}{56.7} & 93.6 &
\multicolumn{1}{c}{58.2} & 92.3
 \\
$\varepsilon$ = 25 &
\multicolumn{1}{c}{\textbf{67.7}} & \textbf{77.2} &
\multicolumn{1}{c}{66.8} & 78.4 &
\multicolumn{1}{c}{64.0} & 87.8 &
\multicolumn{1}{c}{61.0} & 92.0 &
\multicolumn{1}{c}{60.1} & 92.6 &
\multicolumn{1}{c}{61.9} & 90.6
 \\
$\varepsilon$ = 30 &
\multicolumn{1}{c}{\textbf{71.4}} & \textbf{73.4} &
\multicolumn{1}{c}{70.5} & 73.5 &
\multicolumn{1}{c}{67.6} & 83.7 &
\multicolumn{1}{c}{64.4} & 90.4 &
\multicolumn{1}{c}{63.4} & 91.4 &
\multicolumn{1}{c}{65.4} & 88.2
 \\
$\varepsilon$ = 40 &
\multicolumn{1}{c}{\textbf{76.1}} & \textbf{67.3} &
\multicolumn{1}{c}{75.3} & 68.0 &
\multicolumn{1}{c}{72.6} & 75.4 &
\multicolumn{1}{c}{69.4} & 87.2 &
\multicolumn{1}{c}{68.5} & 88.9 &
\multicolumn{1}{c}{70.4} & 83.4
 \\
  
  \midrule
  
  
 \textbf{Norm L$_2$} \\ \textcolor{gray}{PGD2}\\

  $\varepsilon$ = 0.125 &
\multicolumn{1}{c}{\textbf{48.3}} & \textbf{94.3} &
\multicolumn{1}{c}{47.5} & 94.8 &
\multicolumn{1}{c}{46.6} & 95.6 &
\multicolumn{1}{c}{46.7} & 95.5 &
\multicolumn{1}{c}{47.1} & 95.4 &
\multicolumn{1}{c}{46.5} & 95.6
 \\
$\varepsilon$ = 0.25 &
\multicolumn{1}{c}{\textbf{53.2}} & \textbf{91.2} &
\multicolumn{1}{c}{52.6} & 91.6 &
\multicolumn{1}{c}{50.9} & 94.6 &
\multicolumn{1}{c}{50.0} & 95.0 &
\multicolumn{1}{c}{50.0} & 95.0 &
\multicolumn{1}{c}{50.3} & 94.8
 \\
$\varepsilon$ = 0.3125 &
\multicolumn{1}{c}{\textbf{55.8}} & \textbf{89.2} &
\multicolumn{1}{c}{55.2} & 89.9 &
\multicolumn{1}{c}{53.3} & 93.7 &
\multicolumn{1}{c}{51.7} & 94.6 &
\multicolumn{1}{c}{51.5} & 94.7 &
\multicolumn{1}{c}{52.3} & 94.3
 \\
$\varepsilon$ = 0.5 &
\multicolumn{1}{c}{\textbf{63.3}} & \textbf{82.6} &
\multicolumn{1}{c}{62.6} & 83.0 &
\multicolumn{1}{c}{60.0} & 90.7 &
\multicolumn{1}{c}{57.4} & 93.2 &
\multicolumn{1}{c}{56.7} & 93.5 &
\multicolumn{1}{c}{58.2} & 92.4
 \\
$\varepsilon$ = 1 &
\multicolumn{1}{c}{\textbf{76.4}} & \textbf{67.5} &
\multicolumn{1}{c}{75.7} & 67.8 &
\multicolumn{1}{c}{73.1} & 75.0 &
\multicolumn{1}{c}{70.1} & 86.7 &
\multicolumn{1}{c}{69.2} & 88.5 &
\multicolumn{1}{c}{71.0} & 83.0
 \\
$\varepsilon$ = 1.5 &
\multicolumn{1}{c}{\textbf{81.0}} & 63.0 &
\multicolumn{1}{c}{80.5} & \textbf{62.7} &
\multicolumn{1}{c}{78.5} & 63.5 &
\multicolumn{1}{c}{76.2} & 80.7 &
\multicolumn{1}{c}{75.6} & 83.2 &
\multicolumn{1}{c}{76.9} & 74.4
 \\
$\varepsilon$ = 2 &
\multicolumn{1}{c}{\textbf{82.6}} & 62.3 &
\multicolumn{1}{c}{82.1} & \textbf{61.6} &
\multicolumn{1}{c}{80.6} & 62.5 &
\multicolumn{1}{c}{78.6} & 78.5 &
\multicolumn{1}{c}{78.1} & 81.2 &
\multicolumn{1}{c}{79.1} & 72.1
 \\

  \textcolor{gray}{DeepFool} \\ 
No $\varepsilon$ & 
\multicolumn{1}{c}{\textbf{57.0}} & \textbf{91.7} &
\multicolumn{1}{c}{56.7} & \textbf{91.7} &
\multicolumn{1}{c}{55.6} & 93.6 &
\multicolumn{1}{c}{54.6} & 94.1 &
\multicolumn{1}{c}{54.2} & 94.3 &
\multicolumn{1}{c}{54.7} & 94.0
\Bstrut \\ 

  \textcolor{gray}{CW2} \\ 
$\varepsilon$ = 0.01 &
\multicolumn{1}{c}{\textbf{56.4}} & \textbf{90.8} &
\multicolumn{1}{c}{55.9} & 90.9 &
\multicolumn{1}{c}{54.5} & 93.7 &
\multicolumn{1}{c}{53.4} & 94.3 &
\multicolumn{1}{c}{53.0} & 94.5 &
\multicolumn{1}{c}{53.6} & 94.1
\Bstrut \\ 

  \textcolor{gray}{HOP}   \\ 
$\varepsilon$ = 0.1 &
\multicolumn{1}{c}{\textbf{66.1}} & \textbf{87.0} &
\multicolumn{1}{c}{65.1} & 88.2 &
\multicolumn{1}{c}{63.0} & 91.3 &
\multicolumn{1}{c}{61.2} & 92.6 &
\multicolumn{1}{c}{60.8} & 92.9 &
\multicolumn{1}{c}{61.6} & 92.1
\Bstrut \\ \midrule
  

 \textbf{Norm L$_\infty$} \\ \textcolor{gray}{PGDi, FGSM, BIM}\\

  $\varepsilon$ = 0.03125 &
\multicolumn{1}{c}{\textbf{83.0}} & 55.3 &
\multicolumn{1}{c}{82.1} & \textbf{55.2} &
\multicolumn{1}{c}{80.3} & 57.8 &
\multicolumn{1}{c}{77.4} & 77.0 &
\multicolumn{1}{c}{76.8} & 81.3 &
\multicolumn{1}{c}{78.3} & 65.4
 \\
$\varepsilon$ = 0.0625 &
\multicolumn{1}{c}{\textbf{96.0}} & \textbf{17.2} &
\multicolumn{1}{c}{94.6} & 17.4 &
\multicolumn{1}{c}{94.9} & 19.2 &
\multicolumn{1}{c}{94.3} & 21.6 &
\multicolumn{1}{c}{94.4} & 21.1 &
\multicolumn{1}{c}{94.4} & 21.1
 \\
$\varepsilon$ = 0.25 &
\multicolumn{1}{c}{\textbf{97.3}} & \textbf{0.6} &
\multicolumn{1}{c}{94.7} & 5.9 &
\multicolumn{1}{c}{96.5} & 2.5 &
\multicolumn{1}{c}{96.9} & 1.7 &
\multicolumn{1}{c}{97.2} & 1.1 &
\multicolumn{1}{c}{96.7} & 2.1
 \\
$\varepsilon$ = 0.5 &
\multicolumn{1}{c}{\textbf{82.5}} & \textbf{100.0} &
\multicolumn{1}{c}{80.4} & \textbf{100.0} &
\multicolumn{1}{c}{81.9} & \textbf{100.0} &
\multicolumn{1}{c}{82.2} & \textbf{100.0} &
\multicolumn{1}{c}{82.4} & \textbf{100.0} &
\multicolumn{1}{c}{82.0} & \textbf{100.0}
 \\


  \textcolor{gray}{PGDi, FGSM, BIM, SA} \\ 
$\varepsilon$ = 0.125 &
\multicolumn{1}{c}{9.4} & \textbf{99.9} &
\multicolumn{1}{c}{10.4} & 100.0 &
\multicolumn{1}{c}{26.2} & \textbf{99.9} &
\multicolumn{1}{c}{30.9} & 100.0 &
\multicolumn{1}{c}{\textbf{33.8}} & 100.0 &
\multicolumn{1}{c}{27.3} & 100.0
 \\
  
 \textcolor{gray}{PGDi, FGSM, BIM, CWi} \\ 
$\varepsilon$ = 0.3125 &
\multicolumn{1}{c}{\textbf{63.2}} & 99.1 &
\multicolumn{1}{c}{62.7} & \textbf{99.0} &
\multicolumn{1}{c}{61.9} & 99.3 &
\multicolumn{1}{c}{60.9} & 99.5 &
\multicolumn{1}{c}{60.5} & 99.5 &
\multicolumn{1}{c}{61.2} & 99.4
\Bstrut \\ \midrule
  
  \textbf{No norm} \\ \textcolor{gray}{STA}\\

No $\varepsilon$ & 
\multicolumn{1}{c}{88.5} & 38.8 &
\multicolumn{1}{c}{92.0} & 25.1 &
\multicolumn{1}{c}{92.1} & 22.4 &
\multicolumn{1}{c}{\textbf{93.3}} & \textbf{18.3} &
\multicolumn{1}{c}{92.7} & 19.6 &
\multicolumn{1}{c}{92.7} & 19.7
\Bstrut \\ 
  \bottomrule
\end{tabular}
}
\label{tab:cifar10_nss}
\end{table*}

\begin{table*}[!htbp]
\centering
\caption{Simultaneous attacks detection: the proposed method on CIFAR10. We train the detectors on natural and adversarial examples created with PGD algorithm and L$_\infty$ norm constraint. The perturbation magnitude $\varepsilon$ is shown in the columns. We indicate in \textbf{bold} the best result.}
\resizebox{\columnwidth}{!}{%
\begin{tabular}{r|cc|cc|cc|cc|cc|cc}
\toprule
 &
  \multicolumn{12}{c}{\textbf{Ours }} \Bstrut \\ \cline{2-13}  &
   
  \multicolumn{2}{c|}{0.03125} & \multicolumn{2}{c|}{0.0625}& \multicolumn{2}{c|}{0.125}& \multicolumn{2}{c|}{0.25} &
  \multicolumn{2}{c|}{0.3125} &   \multicolumn{2}{c}{0.5}
  \Tstrut\Bstrut \\ \cmidrule{2-13} 
  &
  \multicolumn{1}{c}{\auc} &
  \fpr &
  \multicolumn{1}{c}{\auc} &
  \fpr &
  \multicolumn{1}{c}{\auc} &
  \fpr &
  \multicolumn{1}{c}{\auc} &
  \fpr &
  \multicolumn{1}{c}{\auc} &
  \fpr &
  \multicolumn{1}{c}{\auc} &
  \fpr \\
  \cmidrule{2-13}
\textbf{Norm L$_1$}\\\textcolor{gray}{PGD1}\\
$\varepsilon$ = 5 &
\multicolumn{1}{c}{\textbf{69.7}} & 82.5 &
\multicolumn{1}{c}{65.5} & \textbf{81.5} &
\multicolumn{1}{c}{62.1} & 87.1 &
\multicolumn{1}{c}{56.3} & 93.8 &
\multicolumn{1}{c}{53.2} & 94.7 &
\multicolumn{1}{c}{48.6} & 95.5
 \\
$\varepsilon$ = 10 &
\multicolumn{1}{c}{62.2} & \textbf{83.2} &
\multicolumn{1}{c}{\textbf{62.7}} & 86.3 &
\multicolumn{1}{c}{56.8} & 90.4 &
\multicolumn{1}{c}{52.2} & 94.6 &
\multicolumn{1}{c}{52.9} & 94.6 &
\multicolumn{1}{c}{51.0} & 95.0
 \\
$\varepsilon$ = 15 &
\multicolumn{1}{c}{66.6} & \textbf{72.7} &
\multicolumn{1}{c}{\textbf{73.9}} & 77.9 &
\multicolumn{1}{c}{69.3} & 84.3 &
\multicolumn{1}{c}{65.5} & 89.1 &
\multicolumn{1}{c}{64.4} & 90.9 &
\multicolumn{1}{c}{60.5} & 93.0
 \\
$\varepsilon$ = 20 &
\multicolumn{1}{c}{72.8} & \textbf{58.0} &
\multicolumn{1}{c}{\textbf{83.7}} & 59.3 &
\multicolumn{1}{c}{78.7} & 73.1 &
\multicolumn{1}{c}{73.9} & 82.3 &
\multicolumn{1}{c}{73.6} & 85.3 &
\multicolumn{1}{c}{69.2} & 90.2
 \\
$\varepsilon$ = 25 &
\multicolumn{1}{c}{76.8} & 42.4 &
\multicolumn{1}{c}{\textbf{89.5}} & \textbf{35.9} &
\multicolumn{1}{c}{87.1} & 50.9 &
\multicolumn{1}{c}{81.3} & 68.5 &
\multicolumn{1}{c}{79.3} & 77.7 &
\multicolumn{1}{c}{74.8} & 87.1
 \\
$\varepsilon$ = 30 &
\multicolumn{1}{c}{79.1} & 31.1 &
\multicolumn{1}{c}{\textbf{91.7}} & \textbf{21.5} &
\multicolumn{1}{c}{90.3} & 35.4 &
\multicolumn{1}{c}{84.3} & 61.0 &
\multicolumn{1}{c}{81.9} & 73.1 &
\multicolumn{1}{c}{77.5} & 85.2
 \\
$\varepsilon$ = 40 &
\multicolumn{1}{c}{80.8} & 22.2 &
\multicolumn{1}{c}{\textbf{93.0}} & \textbf{15.0} &
\multicolumn{1}{c}{92.1} & 26.4 &
\multicolumn{1}{c}{85.9} & 56.7 &
\multicolumn{1}{c}{83.2} & 71.1 &
\multicolumn{1}{c}{78.9} & 84.4
 \\
  \midrule
  
  
 \textbf{Norm L$_2$} \\ \textcolor{gray}{PGD2}\\
$\varepsilon$ = 0.125 &
\multicolumn{1}{c}{\textbf{71.3}} & 80.8 &
\multicolumn{1}{c}{67.0} & \textbf{80.0} &
\multicolumn{1}{c}{63.9} & 85.2 &
\multicolumn{1}{c}{56.3} & 93.7 &
\multicolumn{1}{c}{53.8} & 94.6 &
\multicolumn{1}{c}{48.7} & 95.5
 \\
$\varepsilon$ = 0.25 &
\multicolumn{1}{c}{\textbf{63.0}} & \textbf{83.4} &
\multicolumn{1}{c}{62.9} & 86.5 &
\multicolumn{1}{c}{57.2} & 90.3 &
\multicolumn{1}{c}{52.3} & 94.5 &
\multicolumn{1}{c}{52.6} & 94.7 &
\multicolumn{1}{c}{50.0} & 95.2
 \\
$\varepsilon$ = 0.3125 &
\multicolumn{1}{c}{64.1} & \textbf{79.2} &
\multicolumn{1}{c}{\textbf{67.3}} & 83.1 &
\multicolumn{1}{c}{61.0} & 88.8 &
\multicolumn{1}{c}{58.0} & 92.8 &
\multicolumn{1}{c}{57.8} & 93.3 &
\multicolumn{1}{c}{54.6} & 94.3
 \\
$\varepsilon$ = 0.5 &
\multicolumn{1}{c}{72.9} & \textbf{58.9} &
\multicolumn{1}{c}{\textbf{83.7}} & 60.7 &
\multicolumn{1}{c}{79.4} & 73.0 &
\multicolumn{1}{c}{74.6} & 81.1 &
\multicolumn{1}{c}{73.4} & 85.3 &
\multicolumn{1}{c}{68.9} & 90.4
 \\
$\varepsilon$ = 1 &
\multicolumn{1}{c}{81.0} & 21.7 &
\multicolumn{1}{c}{\textbf{92.9}} & \textbf{15.5} &
\multicolumn{1}{c}{91.4} & 26.4 &
\multicolumn{1}{c}{85.5} & 57.2 &
\multicolumn{1}{c}{83.0} & 71.9 &
\multicolumn{1}{c}{78.8} & 84.6
 \\
$\varepsilon$ = 1.5 &
\multicolumn{1}{c}{81.5} & 19.2 &
\multicolumn{1}{c}{\textbf{93.1}} & \textbf{14.2} &
\multicolumn{1}{c}{91.9} & 24.3 &
\multicolumn{1}{c}{85.9} & 56.3 &
\multicolumn{1}{c}{83.3} & 71.6 &
\multicolumn{1}{c}{79.2} & 84.3
 \\
$\varepsilon$ = 2 &
\multicolumn{1}{c}{81.6} & 19.0 &
\multicolumn{1}{c}{\textbf{93.1}} & \textbf{14.2} &
\multicolumn{1}{c}{91.9} & 24.2 &
\multicolumn{1}{c}{85.9} & 56.2 &
\multicolumn{1}{c}{83.3} & 71.5 &
\multicolumn{1}{c}{79.2} & 84.2
 \\
  \textcolor{gray}{DeepFool} \\ 
No $\varepsilon$ & 
\multicolumn{1}{c}{\textbf{91.0}} & \textbf{22.0} &
\multicolumn{1}{c}{87.2} & 33.7 &
\multicolumn{1}{c}{81.6} & 55.5 &
\multicolumn{1}{c}{69.7} & 84.6 &
\multicolumn{1}{c}{63.9} & 91.5 &
\multicolumn{1}{c}{56.2} & 94.4
\Bstrut \\ 

  \textcolor{gray}{CW2} \\ 
$\varepsilon$ = 0.01 &
\multicolumn{1}{c}{52.9} & \textbf{90.6} &
\multicolumn{1}{c}{50.7} & 90.7 &
\multicolumn{1}{c}{\textbf{53.4}} & 92.3 &
\multicolumn{1}{c}{53.2} & 94.4 &
\multicolumn{1}{c}{52.0} & 94.8 &
\multicolumn{1}{c}{50.8} & 95.0
\Bstrut \\ 

  \textcolor{gray}{HOP}   \\ 
  $\varepsilon$ = 0.1 &
\multicolumn{1}{c}{\textbf{91.3}} & \textbf{20.9} &
\multicolumn{1}{c}{89.0} & 31.0 &
\multicolumn{1}{c}{86.1} & 49.1 &
\multicolumn{1}{c}{77.0} & 80.6 &
\multicolumn{1}{c}{72.4} & 88.1 &
\multicolumn{1}{c}{64.3} & 92.8
\Bstrut \\ \midrule
  

 \textbf{Norm L$_\infty$} \\ \textcolor{gray}{PGDi, FGSM, BIM}\\
  
$\varepsilon$ = 0.03125 &
\multicolumn{1}{c}{67.2} & 77.3 &
\multicolumn{1}{c}{77.7} & 65.2 &
\multicolumn{1}{c}{\textbf{82.3}} & \textbf{59.7} &
\multicolumn{1}{c}{78.0} & 72.1 &
\multicolumn{1}{c}{73.7} & 83.6 &
\multicolumn{1}{c}{64.1} & 92.2
 \\
$\varepsilon$ = 0.0625 &
\multicolumn{1}{c}{69.0} & 83.6 &
\multicolumn{1}{c}{85.3} & 47.4 &
\multicolumn{1}{c}{\textbf{92.0}} & \textbf{29.6} &
\multicolumn{1}{c}{90.7} & 35.7 &
\multicolumn{1}{c}{88.0} & 45.6 &
\multicolumn{1}{c}{81.3} & 78.2
 \\
$\varepsilon$ = 0.25 &
\multicolumn{1}{c}{72.0} & 67.4 &
\multicolumn{1}{c}{91.8} & 23.2 &
\multicolumn{1}{c}{\textbf{95.9}} & \textbf{8.8} &
\multicolumn{1}{c}{94.1} & 15.4 &
\multicolumn{1}{c}{92.6} & 19.5 &
\multicolumn{1}{c}{91.6} & 26.5
 \\
$\varepsilon$ = 0.5 &
\multicolumn{1}{c}{58.3} & 84.8 &
\multicolumn{1}{c}{84.2} & 44.1 &
\multicolumn{1}{c}{\textbf{94.6}} & \textbf{9.7} &
\multicolumn{1}{c}{91.2} & 16.5 &
\multicolumn{1}{c}{90.5} & 18.8 &
\multicolumn{1}{c}{91.3} & 22.3
 \\


  \textcolor{gray}{PGDi, FGSM, BIM, SA} \\ 
$\varepsilon$ = 0.125 &
\multicolumn{1}{c}{69.0} & 79.1 &
\multicolumn{1}{c}{84.1} & 41.9 &
\multicolumn{1}{c}{\textbf{88.9}} & \textbf{40.9} &
\multicolumn{1}{c}{86.6} & 52.2 &
\multicolumn{1}{c}{85.4} & 60.3 &
\multicolumn{1}{c}{80.8} & 79.0
 \\
  
 \textcolor{gray}{PGDi, FGSM, BIM, CWi} \\ 
$\varepsilon$ = 0.3125 &
\multicolumn{1}{c}{66.6} & 75.0 &
\multicolumn{1}{c}{\textbf{80.5}} & \textbf{51.5} &
\multicolumn{1}{c}{80.0} & 61.1 &
\multicolumn{1}{c}{72.0} & 83.8 &
\multicolumn{1}{c}{67.2} & 89.9 &
\multicolumn{1}{c}{60.0} & 93.6
\Bstrut \\ \midrule
  
  \textbf{No norm} \\ \textcolor{gray}{STA}\\

No $\varepsilon$ & 
\multicolumn{1}{c}{84.8} & \textbf{33.9} &
\multicolumn{1}{c}{\textbf{85.0}} & 41.5 &
\multicolumn{1}{c}{82.7} & 52.4 &
\multicolumn{1}{c}{72.9} & 77.6 &
\multicolumn{1}{c}{70.3} & 81.4 &
\multicolumn{1}{c}{63.1} & 92.1
\Bstrut \\ 
  \bottomrule
\end{tabular}
}
\label{tab:cifar10_salad}
\end{table*}

\begin{table*}[!htbp]
\centering
\caption{Simultaneous attacks detection: NSS on SVHN. We train NSS on natural and adversarial examples created with PGD algorithm and L$_\infty$ norm constraint. The perturbation magnitude $\varepsilon$ is shown in the columns. We indicate in \textbf{bold} the best result.}
\resizebox{\columnwidth}{!}{%
\begin{tabular}{r|cc|cc|cc|cc|cc|cc}
\toprule
 &
  \multicolumn{12}{c}{\textbf{NSS}} \Bstrut \\ \cline{2-13} &
   
  \multicolumn{2}{c|}{0.03125} & \multicolumn{2}{c|}{0.0625}& \multicolumn{2}{c|}{0.125}& \multicolumn{2}{c|}{0.25} &
  \multicolumn{2}{c|}{0.3125} &   \multicolumn{2}{c}{0.5}
  \Tstrut\Bstrut \\ \cmidrule{2-13} 
  &
  \multicolumn{1}{c}{\auc} &
\fpr &
  \multicolumn{1}{c}{\auc} &
\fpr &
  \multicolumn{1}{c}{\auc} &
\fpr &
  \multicolumn{1}{c}{\auc} &
\fpr &
  \multicolumn{1}{c}{\auc} &
\fpr &
  \multicolumn{1}{c}{\auc} &
\fpr \\\cmidrule{2-13}
\textbf{Norm L$_1$} \\ \textcolor{gray}{PGD1} \\
 $\varepsilon$ = 5 &
\multicolumn{1}{c}{37.9} & 89.3 &
\multicolumn{1}{c}{\textbf{40.2}} & 91.3 &
\multicolumn{1}{c}{37.2} & 89.2 &
\multicolumn{1}{c}{4.9} & 35.5 &
\multicolumn{1}{c}{0.3} & 8.5 &
\multicolumn{1}{c}{0.0} & \textbf{3.1}\\
$\varepsilon$ = 10 &
\multicolumn{1}{c}{33.7} & 89.3 &
\multicolumn{1}{c}{\textbf{36.9}} & 91.3 &
\multicolumn{1}{c}{34.6} & 89.2 &
\multicolumn{1}{c}{6.0} & 35.5 &
\multicolumn{1}{c}{0.4} & 8.5 &
\multicolumn{1}{c}{0.0} & \textbf{3.1}\\
$\varepsilon$ = 15 &
\multicolumn{1}{c}{31.9} & 89.3 &
\multicolumn{1}{c}{\textbf{35.6}} & 91.3 &
\multicolumn{1}{c}{34.4} & 89.2 &
\multicolumn{1}{c}{7.6} & 35.5 &
\multicolumn{1}{c}{0.5} & 8.5 &
\multicolumn{1}{c}{0.1} & \textbf{3.1}\\
$\varepsilon$ = 20 &
\multicolumn{1}{c}{31.5} & 89.3 &
\multicolumn{1}{c}{\textbf{36.1}} & 91.3 &
\multicolumn{1}{c}{35.7} & 89.2 &
\multicolumn{1}{c}{9.5} & 35.5 &
\multicolumn{1}{c}{0.6} & 8.5 &
\multicolumn{1}{c}{0.1} & \textbf{3.1}\\
$\varepsilon$ = 25 &
\multicolumn{1}{c}{32.8} & 89.3 &
\multicolumn{1}{c}{37.8} & 91.3 &
\multicolumn{1}{c}{\textbf{38.2}} & 89.2 &
\multicolumn{1}{c}{11.7} & 35.5 &
\multicolumn{1}{c}{0.9} & 8.5 &
\multicolumn{1}{c}{0.1} & \textbf{3.1}\\
$\varepsilon$ = 30 &
\multicolumn{1}{c}{34.5} & 89.3 &
\multicolumn{1}{c}{39.8} & 91.3 &
\multicolumn{1}{c}{\textbf{40.6}} & 89.2 &
\multicolumn{1}{c}{14.1} & 35.5 &
\multicolumn{1}{c}{1.2} & 8.5 &
\multicolumn{1}{c}{0.1} & \textbf{3.1}\\
$\varepsilon$ = 40 &
\multicolumn{1}{c}{37.9} & 89.3 &
\multicolumn{1}{c}{43.1} & 91.3 &
\multicolumn{1}{c}{\textbf{43.4}} & 89.0 &
\multicolumn{1}{c}{16.4} & 35.5 &
\multicolumn{1}{c}{2.2} & 8.5 &
\multicolumn{1}{c}{0.3} & \textbf{3.1}\\
  \midrule

  
\textbf{Norm L$_2$} \\ \textcolor{gray}{PGD2} \\
  $\varepsilon$ = 0.125 &
\multicolumn{1}{c}{38.7} & 89.3 &
\multicolumn{1}{c}{\textbf{40.8}} & 91.3 &
\multicolumn{1}{c}{37.6} & 89.2 &
\multicolumn{1}{c}{4.7} & 35.5 &
\multicolumn{1}{c}{0.3} & 8.5 &
\multicolumn{1}{c}{0.0} & \textbf{3.1}\\
$\varepsilon$ = 0.25 &
\multicolumn{1}{c}{34.0} & 89.3 &
\multicolumn{1}{c}{\textbf{37.2}} & 91.3 &
\multicolumn{1}{c}{34.6} & 89.2 &
\multicolumn{1}{c}{5.4} & 35.5 &
\multicolumn{1}{c}{0.3} & 8.5 &
\multicolumn{1}{c}{0.0} & \textbf{3.1}\\
$\varepsilon$ = 0.3125 &
\multicolumn{1}{c}{32.6} & 89.3 &
\multicolumn{1}{c}{\textbf{36.1}} & 91.3 &
\multicolumn{1}{c}{34.1} & 89.2 &
\multicolumn{1}{c}{6.1} & 35.5 &
\multicolumn{1}{c}{0.4} & 8.5 &
\multicolumn{1}{c}{0.0} & \textbf{3.1}\\
$\varepsilon$ = 0.5 &
\multicolumn{1}{c}{31.4} & 89.3 &
\multicolumn{1}{c}{\textbf{35.9}} & 91.3 &
\multicolumn{1}{c}{35.4} & 89.2 &
\multicolumn{1}{c}{8.9} & 35.5 &
\multicolumn{1}{c}{0.5} & 8.5 &
\multicolumn{1}{c}{0.1} & \textbf{3.1}\\
$\varepsilon$ = 1 &
\multicolumn{1}{c}{37.4} & 89.3 &
\multicolumn{1}{c}{42.5} & 91.3 &
\multicolumn{1}{c}{\textbf{42.9}} & 89.2 &
\multicolumn{1}{c}{16.0} & 35.5 &
\multicolumn{1}{c}{2.1} & 8.5 &
\multicolumn{1}{c}{0.3} & \textbf{3.1}\\
$\varepsilon$ = 1.5 &
\multicolumn{1}{c}{40.0} & 89.3 &
\multicolumn{1}{c}{46.3} & 91.3 &
\multicolumn{1}{c}{\textbf{46.5}} & 88.4 &
\multicolumn{1}{c}{17.2} & 35.5 &
\multicolumn{1}{c}{2.8} & 8.5 &
\multicolumn{1}{c}{0.6} & \textbf{3.1}\\
$\varepsilon$ = 2 &
\multicolumn{1}{c}{42.1} & 89.3 &
\multicolumn{1}{c}{49.8} & 91.3 &
\multicolumn{1}{c}{\textbf{50.5}} & 88.0 &
\multicolumn{1}{c}{18.7} & 35.5 &
\multicolumn{1}{c}{3.2} & 8.5 &
\multicolumn{1}{c}{0.8} & \textbf{3.1}
 \\

  \textcolor{gray}{DeepFool} \\
No $\varepsilon$ &
\multicolumn{1}{c}{38.1} & 89.3 &
\multicolumn{1}{c}{\textbf{41.3}} & 91.3 &
\multicolumn{1}{c}{39.7} & 89.2 &
\multicolumn{1}{c}{9.2} & 35.5 &
\multicolumn{1}{c}{0.8} & 8.5 &
\multicolumn{1}{c}{0.1} & \textbf{3.1}
\Bstrut \\ 
  
  \textcolor{gray}{CW2} \\ 
  $\varepsilon$ = 0.01 &
\multicolumn{1}{c}{37.9} & 89.3 &
\multicolumn{1}{c}{\textbf{41.0}} & 91.3 &
\multicolumn{1}{c}{39.5} & 89.2 &
\multicolumn{1}{c}{9.3} & 35.5 &
\multicolumn{1}{c}{0.8} & 8.5 &
\multicolumn{1}{c}{0.1} & \textbf{3.1}
\Bstrut \\ 
  
 \textcolor{gray}{HOP}   \\ 
 $\varepsilon$ = 0.1 &
\multicolumn{1}{c}{66.8} & 82.3 &
\multicolumn{1}{c}{\textbf{67.6}} & 84.2 &
\multicolumn{1}{c}{60.3} & 84.6 &
\multicolumn{1}{c}{16.4} & 35.5 &
\multicolumn{1}{c}{2.7} & 8.5 &
\multicolumn{1}{c}{0.7} & \textbf{3.1}
\Bstrut \\ \midrule
  

\textbf{Norm L$_\infty$} \\ \textcolor{gray}{PGDi, FGSM, BIM}\\
 $\varepsilon$ = 0.03125 &
\multicolumn{1}{c}{84.1} & 49.7 &
\multicolumn{1}{c}{\textbf{86.3}} & 46.9 &
\multicolumn{1}{c}{77.5} & 72.1 &
\multicolumn{1}{c}{22.2} & 33.2 &
\multicolumn{1}{c}{4.3} & 8.5 &
\multicolumn{1}{c}{1.2} & \textbf{3.1}\\
$\varepsilon$ = 0.0625 &
\multicolumn{1}{c}{87.4} & \textbf{0.2} &
\multicolumn{1}{c}{\textbf{88.9}} & 0.7 &
\multicolumn{1}{c}{87.5} & 0.6 &
\multicolumn{1}{c}{33.7} & 16.8 &
\multicolumn{1}{c}{7.4} & 6.8 &
\multicolumn{1}{c}{2.5} & 2.7\\
$\varepsilon$ = 0.25 &
\multicolumn{1}{c}{16.7} & 89.3 &
\multicolumn{1}{c}{51.6} & 88.9 &
\multicolumn{1}{c}{\textbf{52.0}} & 85.1 &
\multicolumn{1}{c}{35.4} & \textbf{0.1} &
\multicolumn{1}{c}{8.4} &\textbf{ 0.1} &
\multicolumn{1}{c}{3.0} & \textbf{0.1}\\
$\varepsilon$ = 0.5 &
\multicolumn{1}{c}{4.1} & 89.3 &
\multicolumn{1}{c}{\textbf{46.7}} & 86.7 &
\multicolumn{1}{c}{46.0} & 84.6 &
\multicolumn{1}{c}{35.4} & \textbf{0.1} &
\multicolumn{1}{c}{8.4} & \textbf{0.1} &
\multicolumn{1}{c}{3.0} & \textbf{0.1}
\Bstrut \\ 


\textcolor{gray}{PGDi, FGSM, BIM, SA} 
 \\ 
$\varepsilon$ = 0.125 &
\multicolumn{1}{c}{22.8} & 89.3 &
\multicolumn{1}{c}{32.9} & 91.3 &
\multicolumn{1}{c}{\textbf{43.6}} & 89.2 &
\multicolumn{1}{c}{30.3} & 32.7 &
\multicolumn{1}{c}{7.1} & 8.5 &
\multicolumn{1}{c}{2.5} & \textbf{3.1}
\Bstrut \\

  
 \textcolor{gray}{PGDi, FGSM, BIM, CWi} 
 \\ 
$\varepsilon$ = 0.3125 &
\multicolumn{1}{c}{4.7} & 89.3 &
\multicolumn{1}{c}{\textbf{41.3}} & 91.3 &
\multicolumn{1}{c}{40.8} & 89.2 &
\multicolumn{1}{c}{12.7} & 35.5 &
\multicolumn{1}{c}{1.7} & 8.5 &
\multicolumn{1}{c}{0.4} & \textbf{3.1}
\Bstrut \\ \midrule
  
\textbf{No norm} \\ \textcolor{gray}{STA} \\ 
No $\varepsilon$ & 
\multicolumn{1}{c}{89.3} & \textbf{0.0} &
\multicolumn{1}{c}{\textbf{91.2}} & 0.2 &
\multicolumn{1}{c}{85.9} & 23.4 &
\multicolumn{1}{c}{19.9} & 33.5 &
\multicolumn{1}{c}{4.2} & 8.3 &
\multicolumn{1}{c}{1.4} & 3.1
\Bstrut \\ 
  \bottomrule
\end{tabular}
}
\label{tab:svhn_nss}
\end{table*}


\begin{table*}[!htbp]
\centering
\caption{Simultaneous attacks detection: the proposed method on SVHN. We train NSS on natural and adversarial examples created with PGD algorithm and L$_\infty$ norm constraint. The perturbation magnitude $\varepsilon$ is shown in the columns. We indicate in \textbf{bold} the best result.}
\resizebox{\columnwidth}{!}{%
\begin{tabular}{r|cc|cc|cc|cc|cc|cc}
\toprule
 &
  \multicolumn{12}{c}{\textbf{Ours}} \Bstrut \\ \cline{2-13} &
   
  \multicolumn{2}{c|}{0.03125} & \multicolumn{2}{c|}{0.0625}& \multicolumn{2}{c|}{0.125}& \multicolumn{2}{c|}{0.25} &
  \multicolumn{2}{c|}{0.3125} &   \multicolumn{2}{c}{0.5}
  \Tstrut\Bstrut \\ \cmidrule{2-13} 
  &
  \multicolumn{1}{c}{\auc} &
\fpr &
  \multicolumn{1}{c}{\auc} &
\fpr &
  \multicolumn{1}{c}{\auc} &
\fpr &
  \multicolumn{1}{c}{\auc} &
\fpr &
  \multicolumn{1}{c}{\auc} &
\fpr &
  \multicolumn{1}{c}{\auc} &
\fpr \\\cmidrule{2-13}
\textbf{Norm L$_1$} \\ \textcolor{gray}{PGD1}\\
$\varepsilon$ = 5 &
\multicolumn{1}{c}{\textbf{79.3}} & \textbf{65.2} &
\multicolumn{1}{c}{\textbf{77.4}} & \textbf{73.3} &
\multicolumn{1}{c}{76.9} & 78.9 &
\multicolumn{1}{c}{76.9} & 78.9 &
\multicolumn{1}{c}{76.6} & 79.6 &
\multicolumn{1}{c}{74.1} & 83.8
 \\
$\varepsilon$ = 10 &
\multicolumn{1}{c}{\textbf{74.4}} & \textbf{65.2} &
\multicolumn{1}{c}{72.8} & 73.1 &
\multicolumn{1}{c}{71.9} & 81.6 &
\multicolumn{1}{c}{73.0} & 82.7 &
\multicolumn{1}{c}{71.9} & 84.0 &
\multicolumn{1}{c}{66.9} & 89.5
 \\
$\varepsilon$ = 15 &
\multicolumn{1}{c}{76.0} & \textbf{57.0} &
\multicolumn{1}{c}{75.7} & 64.6 &
\multicolumn{1}{c}{75.8} & 73.0 &
\multicolumn{1}{c}{\textbf{78.9}} & 72.5 &
\multicolumn{1}{c}{77.3} & 75.0 &
\multicolumn{1}{c}{71.8} & 85.1
 \\
$\varepsilon$ = 20 &
\multicolumn{1}{c}{77.3} & \textbf{48.1} &
\multicolumn{1}{c}{77.9} & 55.0 &
\multicolumn{1}{c}{79.2} & 61.9 &
\multicolumn{1}{c}{\textbf{83.6}} & 60.6 &
\multicolumn{1}{c}{82.2} & 64.3 &
\multicolumn{1}{c}{77.4} & 77.0
 \\
$\varepsilon$ = 25 &
\multicolumn{1}{c}{78.2} & \textbf{40.9} &
\multicolumn{1}{c}{79.4} & 44.4 &
\multicolumn{1}{c}{81.5} & 49.4 &
\multicolumn{1}{c}{\textbf{87.0}} & 48.7 &
\multicolumn{1}{c}{85.7} & 52.4 &
\multicolumn{1}{c}{81.4} & 66.6
 \\
$\varepsilon$ = 30 &
\multicolumn{1}{c}{78.8} & \textbf{34.4} &
\multicolumn{1}{c}{80.4} & 35.3 &
\multicolumn{1}{c}{83.0} & 36.6 &
\multicolumn{1}{c}{\textbf{89.3}} & 37.2 &
\multicolumn{1}{c}{88.1} & 41.6 &
\multicolumn{1}{c}{84.4} & 53.9
 \\
$\varepsilon$ = 40 &
\multicolumn{1}{c}{79.7} & 23.5 &
\multicolumn{1}{c}{81.6} & 22.4 &
\multicolumn{1}{c}{84.7} & 20.2 &
\multicolumn{1}{c}{\textbf{92.7}} & \textbf{20.0} &
\multicolumn{1}{c}{91.1} & 23.0 &
\multicolumn{1}{c}{87.8} & 30.5
 \\
  \midrule
  
  
\textbf{Norm L$_2$} \\ \textcolor{gray}{PGD2}\\
 $\varepsilon$ = 0.125 &
\multicolumn{1}{c}{\textbf{82.2}} & \textbf{61.7} &
\multicolumn{1}{c}{80.6} & 68.4 &
\multicolumn{1}{c}{80.3} & 72.5 &
\multicolumn{1}{c}{80.2} & 74.6 &
\multicolumn{1}{c}{80.1} & 73.4 &
\multicolumn{1}{c}{79.6} & 75.7
 \\
$\varepsilon$ = 0.25 &
\multicolumn{1}{c}{\textbf{75.7}} & \textbf{63.6} &
\multicolumn{1}{c}{74.0} & 71.7 &
\multicolumn{1}{c}{73.3} & 80.3 &
\multicolumn{1}{c}{74.1} & 81.6 &
\multicolumn{1}{c}{72.6} & 82.9 &
\multicolumn{1}{c}{67.8} & 89.1
 \\
$\varepsilon$ = 0.3125 &
\multicolumn{1}{c}{\textbf{75.5}} & \textbf{61.6} &
\multicolumn{1}{c}{74.3} & 70.1 &
\multicolumn{1}{c}{73.9} & 78.4 &
\multicolumn{1}{c}{75.1} & 79.7 &
\multicolumn{1}{c}{73.9} & 81.7 &
\multicolumn{1}{c}{70.6} & 86.9
 \\
$\varepsilon$ = 0.5 &
\multicolumn{1}{c}{77.2} & \textbf{50.6} &
\multicolumn{1}{c}{77.6} & 57.4 &
\multicolumn{1}{c}{78.6} & 64.2 &
\multicolumn{1}{c}{\textbf{82.5}} & 64.4 &
\multicolumn{1}{c}{81.1} & 67.3 &
\multicolumn{1}{c}{76.3} & 79.5
 \\
$\varepsilon$ = 1 &
\multicolumn{1}{c}{79.6} & 25.8 &
\multicolumn{1}{c}{81.3} & 24.8 &
\multicolumn{1}{c}{84.3} & \textbf{24.1} &
\multicolumn{1}{c}{\textbf{92.3}} & 24.6 &
\multicolumn{1}{c}{90.7} & 27.7 &
\multicolumn{1}{c}{87.1} & 36.4
 \\
$\varepsilon$ = 1.5 &
\multicolumn{1}{c}{80.2} & 19.5 &
\multicolumn{1}{c}{82.2} & 17.6 &
\multicolumn{1}{c}{85.6} & 14.3 &
\multicolumn{1}{c}{\textbf{94.1}} & \textbf{7.5} &
\multicolumn{1}{c}{92.9} & 8.6 &
\multicolumn{1}{c}{89.9} & 11.8
 \\
$\varepsilon$ = 2 &
\multicolumn{1}{c}{80.5} & 19.4 &
\multicolumn{1}{c}{82.5} & 17.5 &
\multicolumn{1}{c}{85.9} & 14.1 &
\multicolumn{1}{c}{\textbf{94.9}} & \textbf{5.3} &
\multicolumn{1}{c}{94.5} & 6.8 &
\multicolumn{1}{c}{90.7} & 9.5
 \\ 
  \textcolor{gray}{DeepFool} \\ 
No $\varepsilon$ & 
\multicolumn{1}{c}{\textbf{96.3}} & \textbf{8.5} &
\multicolumn{1}{c}{95.9} & 10.4 &
\multicolumn{1}{c}{95.1} & 12.9 &
\multicolumn{1}{c}{94.9} & 12.0 &
\multicolumn{1}{c}{95.3} & 12.0 &
\multicolumn{1}{c}{95.5} & 12.6
\Bstrut \\

  \textcolor{gray}{CW2} \\ 
$\varepsilon$ = 0.01 &
\multicolumn{1}{c}{\textbf{59.7}} & \textbf{76.3} &
\multicolumn{1}{c}{57.2} & 80.2 &
\multicolumn{1}{c}{53.4} & 89.9 &
\multicolumn{1}{c}{54.2} & 92.0 &
\multicolumn{1}{c}{51.1} & 93.5 &
\multicolumn{1}{c}{44.3} & 96.2
\Bstrut \\

  \textcolor{gray}{HOP}  \\
  $\varepsilon$ = 0.1 &
\multicolumn{1}{c}{\textbf{96.1}} & \textbf{7.9} &
\multicolumn{1}{c}{95.6} & 9.8 &
\multicolumn{1}{c}{95.9} & 11.7 &
\multicolumn{1}{c}{96.0} & 10.2 &
\multicolumn{1}{c}{95.9} & 9.9 &
\multicolumn{1}{c}{\textbf{96.1}} & 10.0
\Bstrut \\ \midrule
  

 \textbf{Norm L$_\infty$}\\\textcolor{gray}{PGDi, FGSM, BIM} \\
$\varepsilon$ = 0.03125 &
\multicolumn{1}{c}{74.3} & \textbf{60.0} &
\multicolumn{1}{c}{75.8} & 60.3 &
\multicolumn{1}{c}{77.8} & 62.6 &
\multicolumn{1}{c}{\textbf{81.4}} & 64.8 &
\multicolumn{1}{c}{80.1} & 67.0 &
\multicolumn{1}{c}{76.7} & 75.4
 \\
$\varepsilon$ = 0.0625 &
\multicolumn{1}{c}{78.4} & 36.0 &
\multicolumn{1}{c}{80.3} & 34.1 &
\multicolumn{1}{c}{83.2} & 33.8 &
\multicolumn{1}{c}{\textbf{89.1}} & \textbf{33.3} &
\multicolumn{1}{c}{87.9} & 34.4 &
\multicolumn{1}{c}{85.7} & 37.4
 \\
$\varepsilon$ = 0.25 &
\multicolumn{1}{c}{80.1} & 19.4 &
\multicolumn{1}{c}{82.1} & 17.5 &
\multicolumn{1}{c}{85.3} & \textbf{15.8} &
\multicolumn{1}{c}{\textbf{92.3}} & 16.4 &
\multicolumn{1}{c}{92.1} & 16.8 &
\multicolumn{1}{c}{89.6} & 17.0
 \\
$\varepsilon$ = 0.5 &
\multicolumn{1}{c}{80.3} & 19.4 &
\multicolumn{1}{c}{82.3} & 17.5 &
\multicolumn{1}{c}{85.5} & \textbf{14.1} &
\multicolumn{1}{c}{\textbf{92.9}} & 14.4 &
\multicolumn{1}{c}{91.7} & 15.2 &
\multicolumn{1}{c}{90.1} & 14.8\\

  
 \textcolor{gray}{PGDi, FGSM, BIM, SA} \\ 
$\varepsilon$ = 0.125 &
\multicolumn{1}{c}{78.9} & 29.0 &
\multicolumn{1}{c}{80.8} & \textbf{28.2} &
\multicolumn{1}{c}{83.8} & 28.7 &
\multicolumn{1}{c}{\textbf{89.2}} & 29.0 &
\multicolumn{1}{c}{88.4} & 28.9 &
\multicolumn{1}{c}{86.8} & 28.4
\Bstrut \\ 

 \textcolor{gray}{PGDi, FGSM, BIM, CWi} \\
$\varepsilon$ = 0.3125 &
\multicolumn{1}{c}{78.7} & 33.4 &
\multicolumn{1}{c}{80.5} & 31.9 &
\multicolumn{1}{c}{83.1} & 34.0 &
\multicolumn{1}{c}{\textbf{88.2}} & 33.1 &
\multicolumn{1}{c}{88.1} & 31.7 &
\multicolumn{1}{c}{86.7} & \textbf{31.3}
\Bstrut \\\midrule

\textbf{No norm} \\ \textcolor{gray}{STA}\\
No $\varepsilon$ &
\multicolumn{1}{c}{\textbf{94.7}} & \textbf{14.5} &
\multicolumn{1}{c}{93.3} & 16.8 &
\multicolumn{1}{c}{89.9} & 23.1 &
\multicolumn{1}{c}{90.2} & 23.2 &
\multicolumn{1}{c}{91.0} & 22.4 &
\multicolumn{1}{c}{91.1} & 22.5
\Bstrut \\ 
  \bottomrule
\end{tabular}
}
\label{tab:svhn_salad}
\end{table*}

\subsection{Additional Plots}
\label{app:additional_plot}
\begin{figure}[htbp]
\centering
\begin{subfigure}[b]{.24\columnwidth}
\centering
\includegraphics[width=\columnwidth]{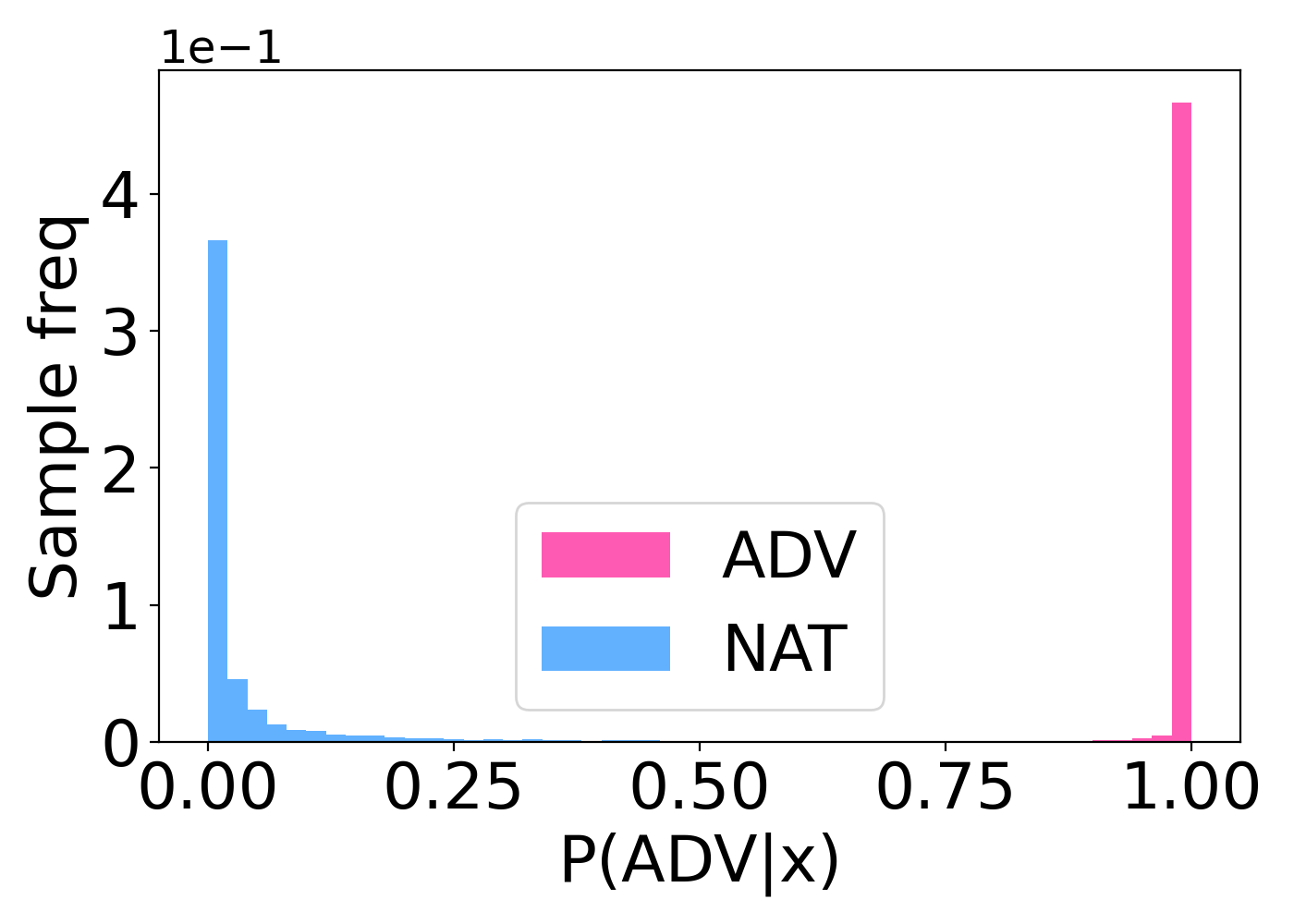}
\caption{PGD-L$_1$-40-ACE}
\label{fig:h_CE}
\end{subfigure}
\hfill
\begin{subfigure}[b]{.24\columnwidth}
\includegraphics[width=\columnwidth]{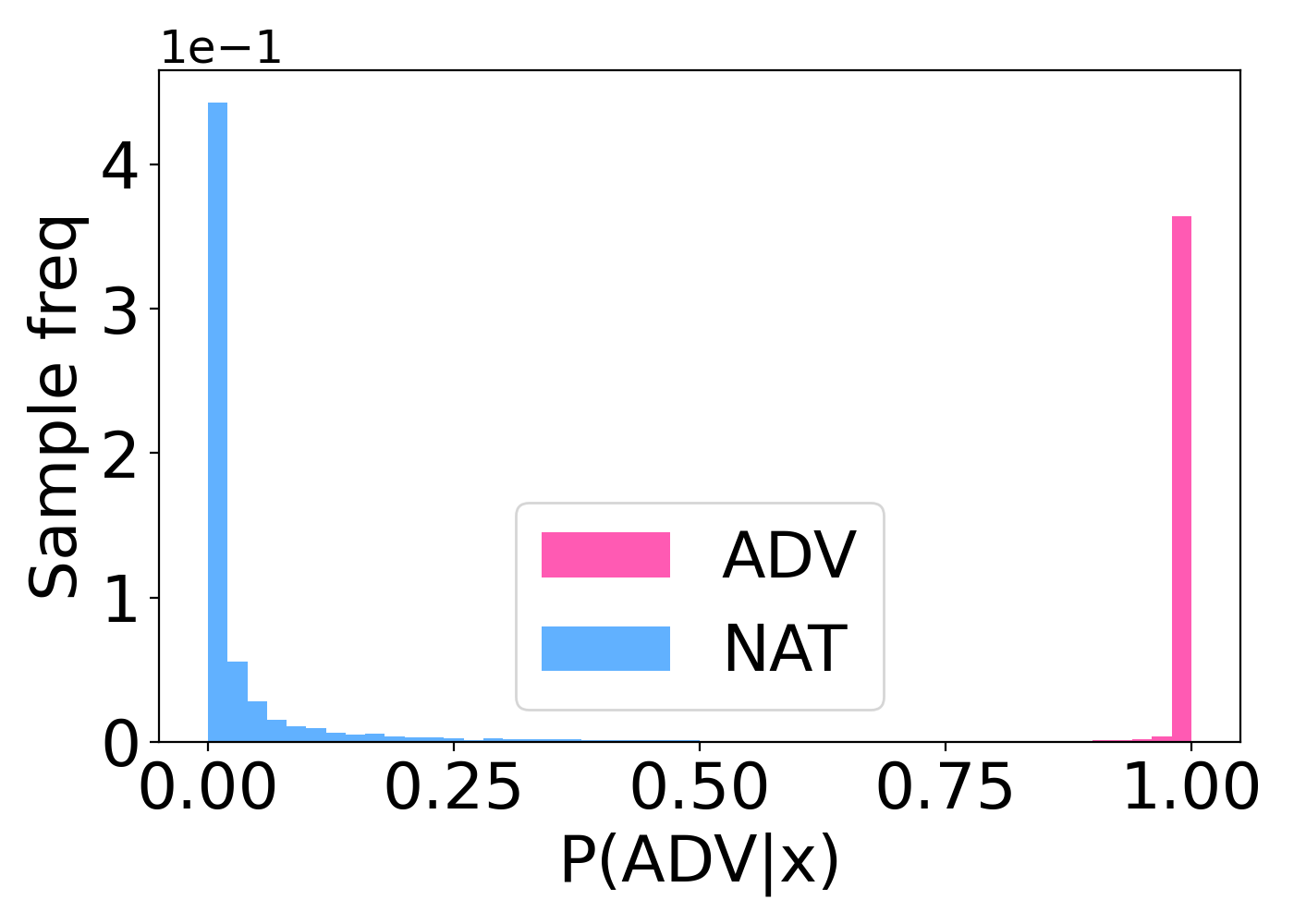}
\caption{PGD-L$_1$-40-KL}
\label{fig:h_KL}
\end{subfigure}
\hfill
\begin{subfigure}[b]{.24\columnwidth}
\includegraphics[width=\columnwidth]{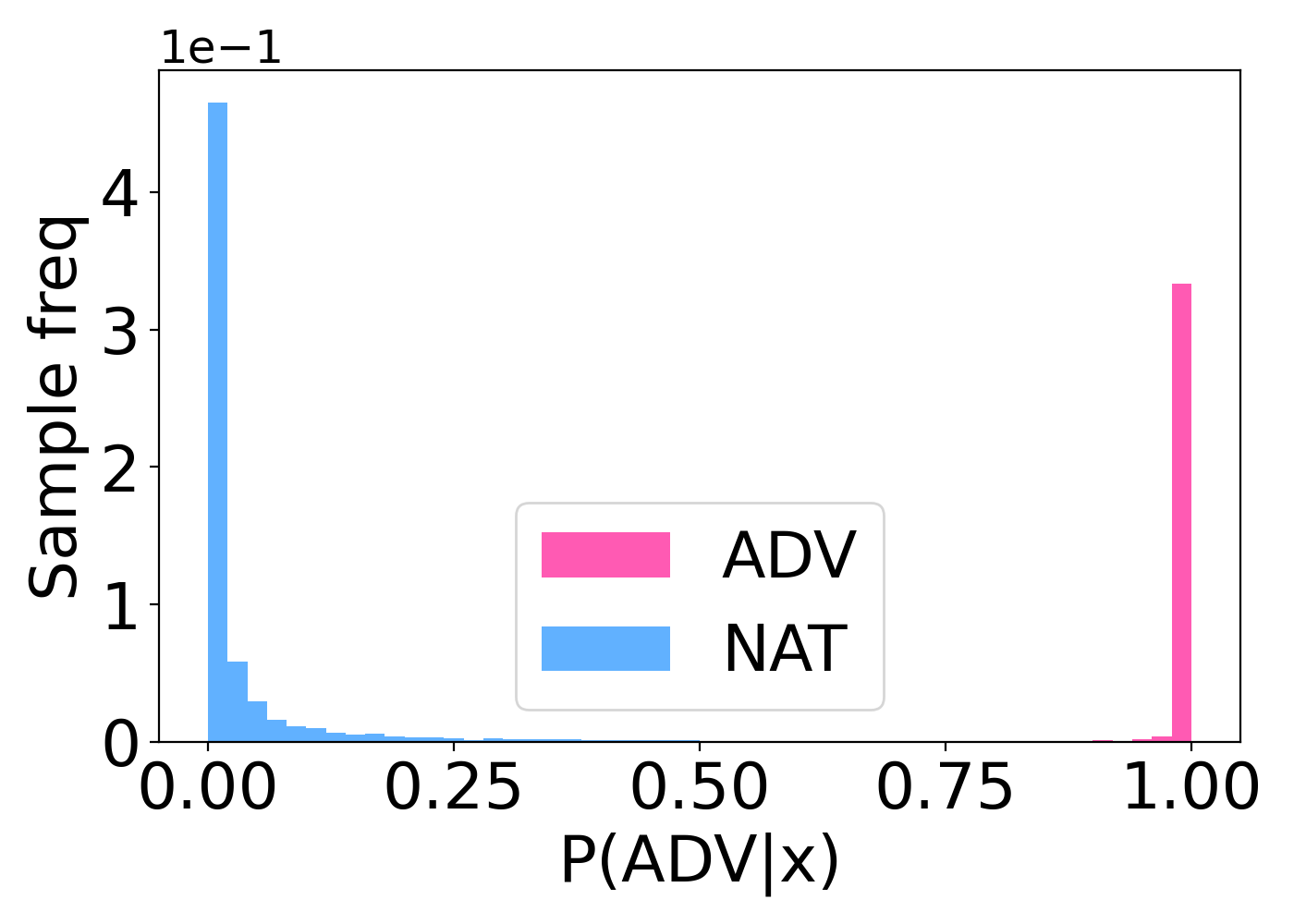}
\caption{PGD-L$_1$-40-FR}
\label{fig:h_rao}
\end{subfigure}
\hfill
\begin{subfigure}[b]{.24\columnwidth}
\includegraphics[width=\columnwidth]{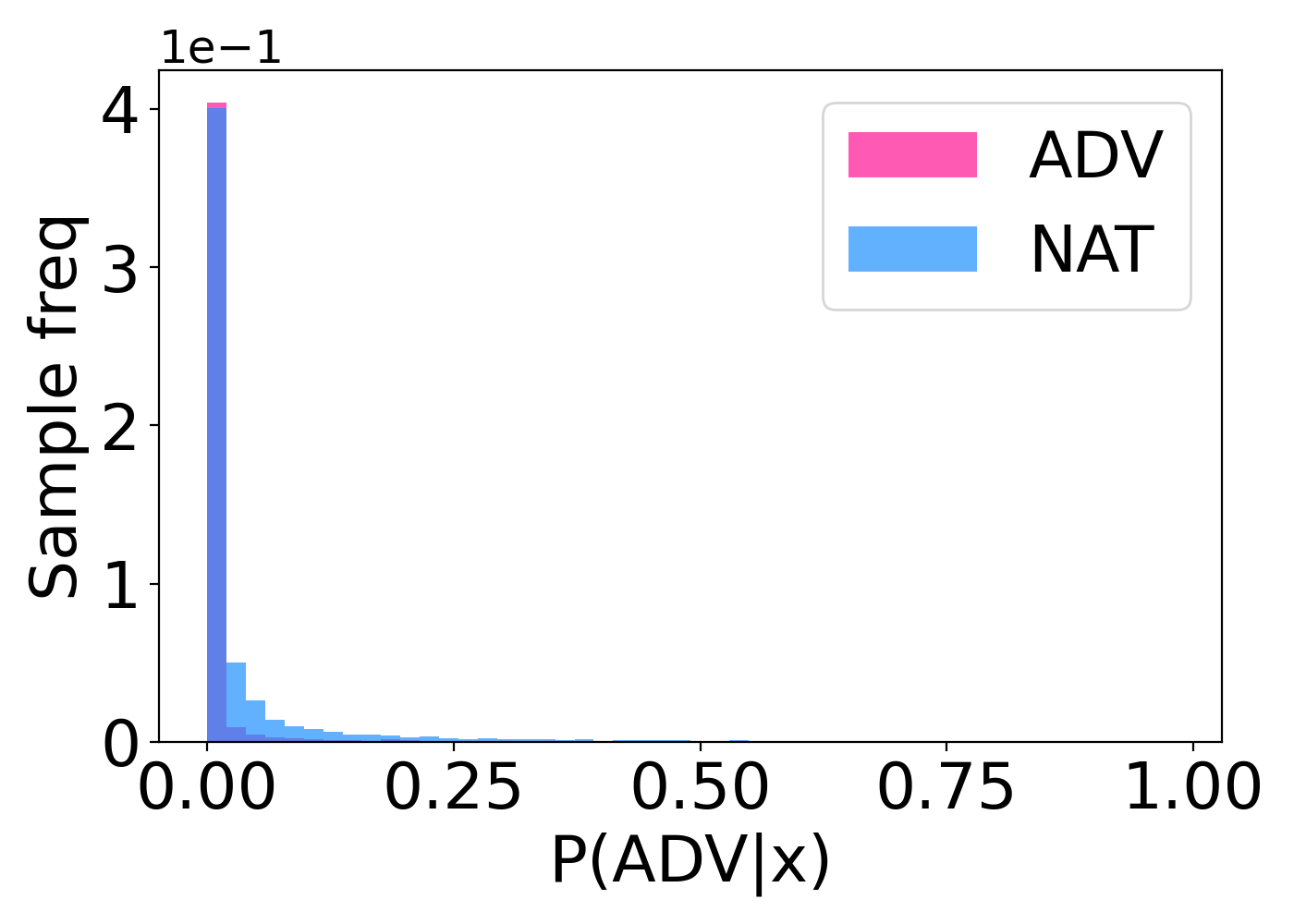}
\caption{PGD-L$_1$-40-Gini}
\label{fig:h_g}
\end{subfigure}
\caption{In pink the results for the adversarial examples and in blue the ones for the naturals. In this simulation, we consider a subset of the available detectors (ACE, KL, FR). Under each plot, we indicate the tested attack configuration parameters: algorithm-L$_p$-$\varepsilon$-loss.}
\label{fig:evaluation_2}
\end{figure}

The specific shape in the histograms depends on the set of considered detectors. To shed light on this fact, we include the plots in~\cref{fig:evaluation_2} in which we consider a subset of the available detectors (ACE, KL, FR). These plots should be compared with the ones in~\cref{fig:evaluation}.

\begin{figure*}[!htbp]
	\centering
	\begin{subfigure}[b]{.3\columnwidth}
	    \centering
	    \includegraphics[width=\columnwidth]{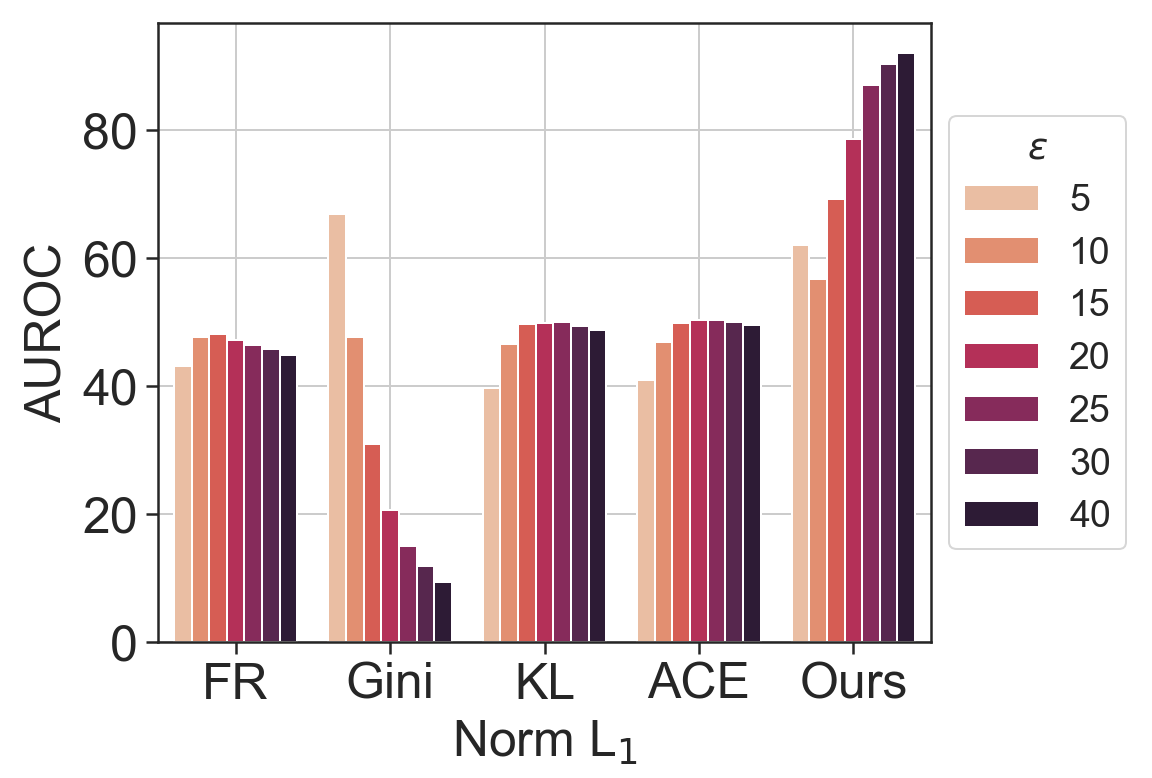}
	    \caption{}
	    \label{fig:eps_l1}
	\end{subfigure}
        \begin{subfigure}[b]{.3\columnwidth}
            \centering
            \includegraphics[width=\columnwidth]{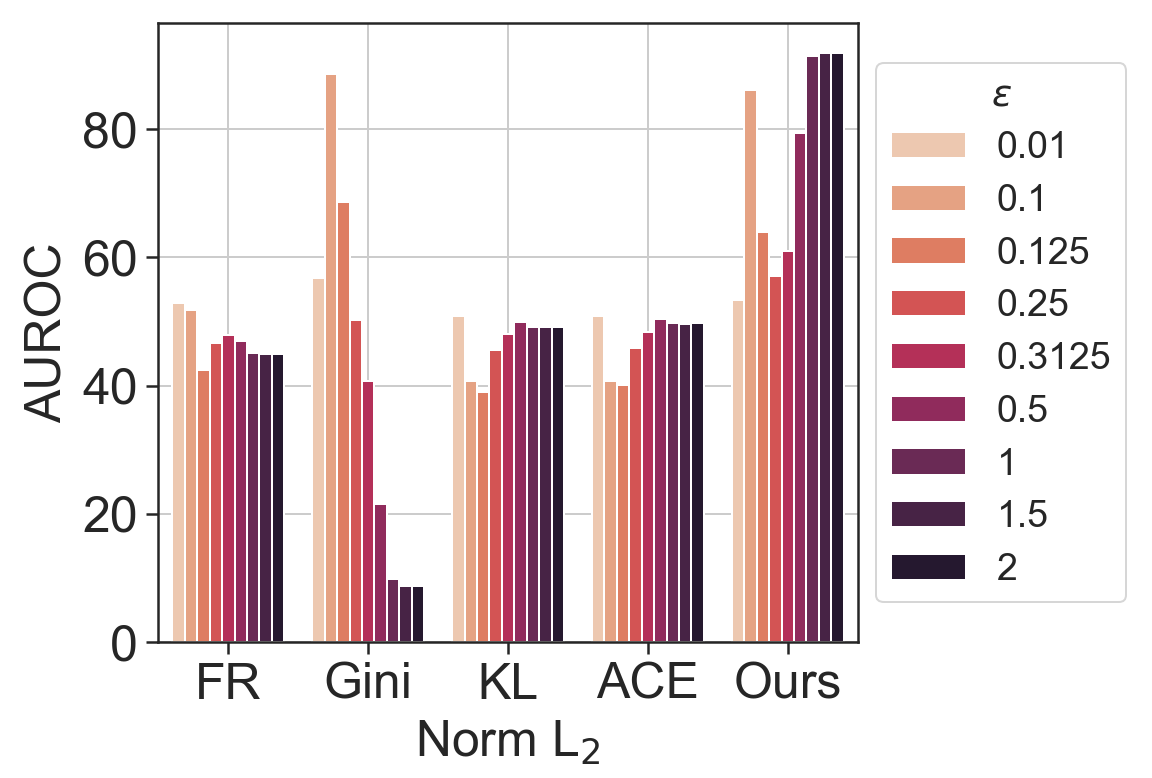}
            \caption{}
        \label{fig:eps_l2}
	\end{subfigure}
        \begin{subfigure}[b]{.3\columnwidth}
            \centering
            \includegraphics[width=\columnwidth]{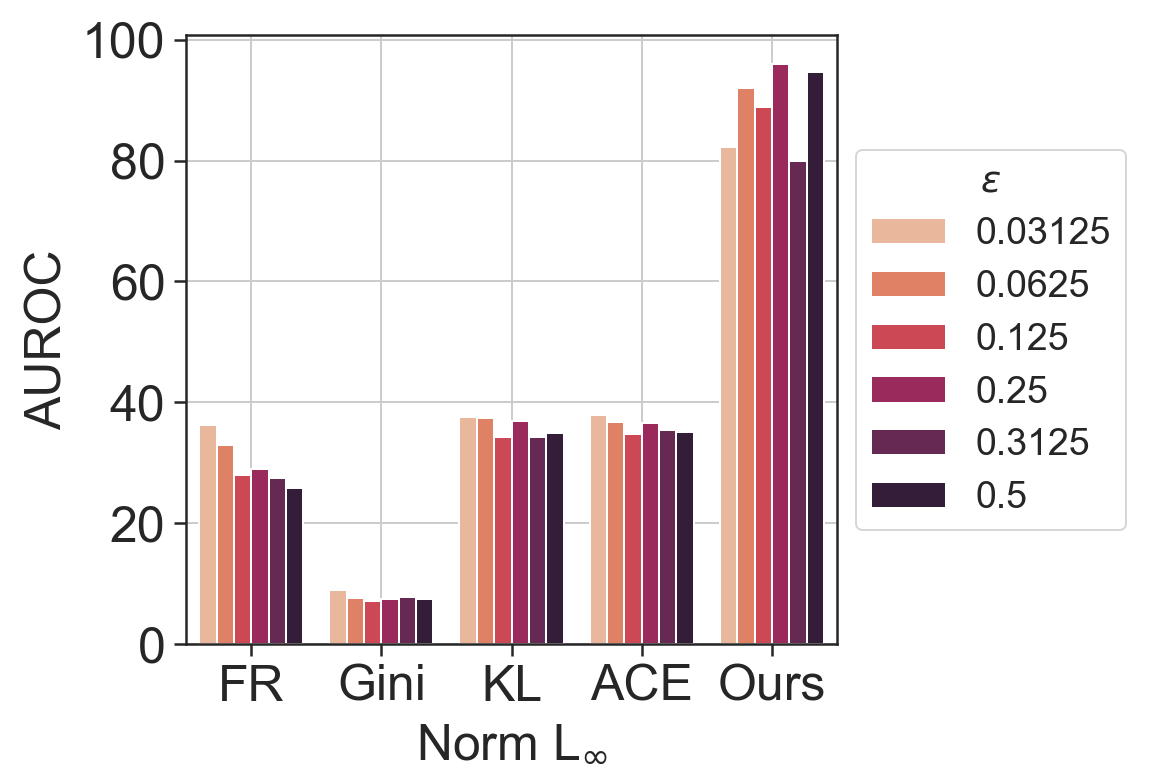}
            \caption{}
        \label{fig:eps_linf}
        \end{subfigure}
	\caption{Performance of the various detectors grouped by L$_p$-norm and perturbation magnitude $\varepsilon$ on CIFAR10. Each \textit{shallow} detector is named after the loss function used to craft the attacks they it is trained to detect. The plot shows how our method consistently attains better performance than the single one on all the different adversarial attacks, supporting the claim of optimality in~\cref{sec:math_framework}.}
	\label{fig:eps_norm}
\end{figure*}

Moreover, we include the detectors' performances analyzed by perturbation magnitude in~\cref{fig:eps_norm}

\subsection{Evaluation of the Proposed Solution in the Optimal Setting}
\subsubsection{Ablation Study}
\label{app:res_th_framework}
In~\cref{tab:003125,tab:00625,tab:0125,tab:025,tab:03125,tab:05} we provide the complete set of results for the theoretical setting applied to CIFAR10. Additionally, the~\cref{tab:svhn003125,tab:svhn00625,tab:svhn0125,tab:svhn025,tab:svhn03125,tab:svhn05} include the results for SVHN. Notice that, we include the results for all the possible combinations of the detectors in the pool.
\begin{table}[!htb]
\begin{minipage}{0.47\columnwidth}
\caption{Theoretical framework setting: CIFAR10, PGD algorithm, $\varepsilon=0.03125$, L$_\infty$ norm.}
\centering
\resizebox{\columnwidth}{!}{%
\begin{tabular}{llcc}
\toprule
Pool of detectors & Attacks & \auc & \fpr \\
\midrule
NSS & ACE+KL & 91.0 & 35.0 \\
ACE+KL & ACE+KL & \textbf{99.9} & \textbf{0.0} \\
ACE+KL+NSS & ACE+KL & \textbf{99.9} & 0.2 \\
\midrule
NSS & ACE+FR & 91.1 & 34.8 \\
ACE+FR & ACE+FR & \textbf{99.9} & \textbf{0.0} \\
ACE+FR+NSS & ACE+FR & \textbf{99.9} & 0.3 \\
\midrule
NSS & ACE+Gini & 88.7 & 43.6 \\
ACE+Gini & ACE+Gini & \textbf{96.0} & 17.6 \\
ACE+Gini+NSS & ACE+Gini & 95.7 & \textbf{15.7} \\
\midrule
NSS & KL+FR & 91.0 & 35.5 \\
KL+FR & KL+FR & \textbf{99.9} & \textbf{0.0} \\
KL+FR+NSS & KL+FR & \textbf{99.9} & 0.4 \\
\midrule
NSS & KL+Gini & 88.7 & 44.3 \\
KL+Gini & KL+Gini & \textbf{95.4} & 18.7 \\
KL+Gini+NSS & KL+Gini & 94.9 & \textbf{16.9} \\
\midrule
NSS & FR+Gini & 88.7 & 44.0 \\
FR+Gini & FR+Gini & \textbf{93.4} & 21.9 \\
FR+Gini+NSS & FR+Gini & 93.0 & \textbf{19.2} \\
\midrule
NSS & ACE+KL+FR & 90.1 & 37.6 \\
ACE+KL+FR & ACE+KL+FR & \textbf{99.9} & \textbf{0.0} \\
ACE+KL+FR+NSS & ACE+KL+FR & \textbf{99.9} & 0.3 \\
\midrule
NSS & ACE+KL+Gini & 87.8 & 45.5 \\
ACE+KL+Gini & ACE+KL+Gini & \textbf{95.9} & 17.8 \\
ACE+KL+Gini+NSS & ACE+KL+Gini & 95.6 & \textbf{16.0} \\
\midrule
NSS & KL+FR+Gini & 88.0 & 45.4 \\
KL+FR+Gini & KL+FR+Gini & \textbf{95.1} & 19.7 \\
KL+FR+Gini+NSS & KL+FR+Gini & 94.7 & \textbf{17.4} \\
\midrule
NSS & ACE+KL+FR+Gini & 87.3 & 46.7 \\
ACE+KL+FR+Gini & ACE+KL+FR+Gini & \textbf{95.6} & 18.8 \\
ACE+KL+FR+Gini+NSS & ACE+KL+FR+Gini & 95.3 & \textbf{16.5} \\
\bottomrule
\end{tabular}
}
\label{tab:003125}
\end{minipage}
\hfill
\begin{minipage}{0.47\columnwidth}
\caption{Theoretical framework setting: CIFAR10, PGD algorithm, $\varepsilon=0.0625$, L$_\infty$ norm.}
\centering
\resizebox{\columnwidth}{!}{%
\begin{tabular}{llcc}
\toprule
Pool of detectors & Attacks & \auc & \fpr \\
\midrule
NSS & ACE+KL & 98.9 & 4.3 \\
ACE+KL & ACE+KL & \textbf{100.0} & \textbf{0.0} \\
ACE+KL+NSS & ACE+KL & \textbf{100.0} & \textbf{0.0} \\
\midrule
NSS & ACE+FR & 98.9 & 4.2 \\
ACE+FR & ACE+FR & \textbf{100.0} & \textbf{0.0} \\
ACE+FR+NSS & ACE+FR & \textbf{100.0} & \textbf{0.0} \\
\midrule
NSS & ACE+Gini & \textbf{98.0} & 8.1 \\
ACE+Gini & ACE+Gini & 97.7 & 7.2 \\
ACE+Gini+NSS & ACE+Gini & 97.6 & \textbf{5.8} \\
\midrule
NSS & KL+FR & 98.9 & 4.3 \\
KL+FR & KL+FR & \textbf{100.0} & \textbf{0.0} \\
KL+FR+NSS & KL+FR & \textbf{100.0} & \textbf{0.0} \\
\midrule
NSS & KL+Gini & \textbf{98.0} & 8.3 \\
KL+Gini & KL+Gini & 97.2 & 7.7 \\
KL+Gini+NSS & KL+Gini & 96.8 & \textbf{6.6} \\
\midrule
NSS & FR+Gini & \textbf{98.0} & \textbf{8.1} \\
FR+Gini & FR+Gini & 95.4 & 10.1 \\
FR+Gini+NSS & FR+Gini & 95.5 & \textbf{8.1} \\
\midrule
NSS & ACE+KL+FR & 98.7 & 4.9 \\
ACE+KL+FR & ACE+KL+FR & \textbf{100.0} & \textbf{0.0} \\
ACE+KL+FR+NSS & ACE+KL+FR & \textbf{100.0} & \textbf{0.0} \\
\midrule
NSS & ACE+KL+Gini & \textbf{97.9} & 8.6 \\
ACE+KL+Gini & ACE+KL+Gini & 97.5 & 7.4 \\
ACE+KL+Gini+NSS & ACE+KL+Gini & 97.3 & \textbf{6.1} \\
\midrule
NSS & KL+FR+Gini & \textbf{97.9} & 8.6 \\
KL+FR+Gini & KL+FR+Gini & 96.9 & 8.3 \\
KL+FR+Gini+NSS & KL+FR+Gini & 96.6 & \textbf{6.9} \\
\midrule
NSS & ACE+KL+FR+Gini & \textbf{97.8} & 8.9 \\
ACE+KL+FR+Gini & ACE+KL+FR+Gini & 97.3 & 8.0 \\
ACE+KL+FR+Gini+NSS & ACE+KL+FR+Gini & 97.1 & \textbf{6.3} \\
\bottomrule
\end{tabular}
}
\label{tab:00625}
\end{minipage}
\end{table}

\begin{table}[!htb]
\begin{minipage}{0.47\columnwidth}
\caption{Theoretical framework setting: CIFAR10, PGD algorithm, $\varepsilon=0.125$, L$_\infty$ norm.}
\centering
\resizebox{\columnwidth}{!}{%
\begin{tabular}{llcc}
\toprule
Pool of detectors & Attacks & \auc & \fpr \\
\midrule
NSS & ACE+KL & 99.7 & 0.6 \\
ACE+KL & ACE+KL & \textbf{100.0} & \textbf{0.0} \\
ACE+KL+NSS & ACE+KL & \textbf{100.0} & \textbf{0.0} \\
\midrule
NSS & ACE+FR & 99.6 & 0.6 \\
ACE+FR & ACE+FR & \textbf{100.0} & \textbf{0.0} \\
ACE+FR+NSS & ACE+FR & \textbf{100.0} & \textbf{0.0} \\
\midrule
NSS & ACE+Gini & \textbf{99.6} & \textbf{0.6} \\
ACE+Gini & ACE+Gini & 97.1 & 7.3 \\
ACE+Gini+NSS & ACE+Gini & 97.3 & 3.2 \\
\midrule
NSS & KL+FR & 99.6 & 0.6 \\
KL+FR & KL+FR & \textbf{100.0} & \textbf{0.0} \\
KL+FR+NSS & KL+FR & \textbf{100.0} & \textbf{0.0} \\
\midrule
NSS & KL+Gini & \textbf{99.6} & \textbf{0.6} \\
KL+Gini & KL+Gini & 96.9 & 7.3 \\
KL+Gini+NSS & KL+Gini & 97.2 & 3.1 \\
\midrule
NSS & FR+Gini & \textbf{99.6} & \textbf{0.6} \\
FR+Gini & FR+Gini & 95.3 & 9.4 \\
FR+Gini+NSS & FR+Gini & 96.1 & 4.2 \\
\midrule
NSS & ACE+KL+FR & 99.6 & 0.6 \\
ACE+KL+FR & ACE+KL+FR & \textbf{100.0} & \textbf{0.0} \\
ACE+KL+FR+NSS & ACE+KL+FR & \textbf{100.0} & \textbf{0.0} \\
\midrule
NSS & ACE+KL+Gini & \textbf{99.6} & \textbf{0.6} \\
ACE+KL+Gini & ACE+KL+Gini & 97.2 & 7.2 \\
ACE+KL+Gini+NSS & ACE+KL+Gini & 97.3 & 3.1 \\
\midrule
NSS & KL+FR+Gini & \textbf{99.6} & \textbf{0.6} \\
KL+FR+Gini & KL+FR+Gini & 96.6 & 8.0 \\
KL+FR+Gini+NSS & KL+FR+Gini & 97.0 & 3.3 \\
\midrule
NSS & ACE+KL+FR+Gini & \textbf{99.6} & \textbf{0.6} \\
ACE+KL+FR+Gini & ACE+KL+FR+Gini & 96.8 & 7.8 \\
ACE+KL+FR+Gini+NSS & ACE+KL+FR+Gini & 97.1 & 3.2 \\
\bottomrule
\end{tabular}
}
\label{tab:0125}
\end{minipage}
\hfill
\begin{minipage}{0.47\columnwidth}
\caption{Theoretical framework setting: CIFAR10, PGD algorithm, $\varepsilon=0.25$, L$_\infty$ norm.}
\centering
\resizebox{\columnwidth}{!}{%
\begin{tabular}{llcc}
\toprule
Pool of detectors & Attacks & \auc & \fpr \\
\midrule
NSS & ACE+KL & 99.7 & 0.6 \\
ACE+KL & ACE+KL & \textbf{100.0} & \textbf{0.0} \\
ACE+KL+NSS & ACE+KL & \textbf{100.0} & \textbf{0.0} \\
\midrule
NSS & ACE+FR & 99.7 & 0.6 \\
ACE+FR & ACE+FR & \textbf{100.0} & \textbf{0.0} \\
ACE+FR+NSS & ACE+FR & \textbf{100.0} & \textbf{0.0} \\
\midrule
NSS & ACE+Gini & \textbf{99.6} & \textbf{0.6} \\
ACE+Gini & ACE+Gini & 97.2 & 7.2 \\
ACE+Gini+NSS & ACE+Gini & 97.6 & 2.8 \\
\midrule
NSS & KL+FR & 99.7 & 0.6 \\
KL+FR & KL+FR & \textbf{100.0} & \textbf{0.0} \\
KL+FR+NSS & KL+FR & \textbf{100.0} & \textbf{0.0} \\
\midrule
NSS & KL+Gini & \textbf{99.6} & \textbf{0.6} \\
KL+Gini & KL+Gini & 96.4 & 7.7 \\
KL+Gini+NSS & KL+Gini & 96.6 & 3.8 \\
\midrule
NSS & FR+Gini & \textbf{99.6} & \textbf{0.6} \\
FR+Gini & FR+Gini & 95.1 & 9.5 \\
FR+Gini+NSS & FR+Gini & 96.1 & 4.2 \\
\midrule
NSS & ACE+KL+FR & 99.7 & 0.6 \\
ACE+KL+FR & ACE+KL+FR & \textbf{100.0} & \textbf{0.0} \\
ACE+KL+FR+NSS & ACE+KL+FR & \textbf{100.0} & \textbf{0.0} \\
\midrule
NSS & ACE+KL+Gini & \textbf{99.6} & \textbf{0.6} \\
ACE+KL+Gini & ACE+KL+Gini & 96.9 & 7.4 \\
ACE+KL+Gini+NSS & ACE+KL+Gini & 97.1 & 3.3 \\
\midrule
NSS & KL+FR+Gini & \textbf{99.6} & \textbf{0.6} \\
KL+FR+Gini & KL+FR+Gini & 96.1 & 8.4 \\
KL+FR+Gini+NSS & KL+FR+Gini & 96.4 & 4.0 \\
\midrule
NSS & ACE+KL+FR+Gini & \textbf{99.6} & \textbf{0.6} \\
ACE+KL+FR+Gini & ACE+KL+FR+Gini & 96.5 & 8.1 \\
ACE+KL+FR+Gini+NSS & ACE+KL+FR+Gini & 96.8 & 3.5 \\
\bottomrule
\end{tabular}
}
\label{tab:025}
\end{minipage}
\end{table}

\begin{table}[!htb]
\begin{minipage}{0.47\columnwidth}
\caption{Theoretical framework setting: CIFAR10, PGD algorithm, $\varepsilon=0.3125$, L$_\infty$ norm.}
\centering
\resizebox{\columnwidth}{!}{%
\begin{tabular}{llcc}
\toprule
Pool of detectors & Attacks & \auc & \fpr \\
\midrule
NSS & ACE+KL & 99.7 & 0.6 \\
ACE+KL & ACE+KL & \textbf{100.0} & \textbf{0.0} \\
ACE+KL+NSS & ACE+KL & \textbf{100.0} & \textbf{0.0} \\
\midrule
NSS & ACE+FR & 99.7 & 0.6 \\
ACE+FR & ACE+FR & \textbf{100.0} & \textbf{0.0} \\
ACE+FR+NSS & ACE+FR & \textbf{100.0} & \textbf{0.0} \\
\midrule
NSS & ACE+Gini & \textbf{99.7} & \textbf{0.6} \\
ACE+Gini & ACE+Gini & 96.9 & 7.4 \\
ACE+Gini+NSS & ACE+Gini & 97.4 & 3.1 \\
\midrule
NSS & KL+FR & 99.7 & 0.6 \\
KL+FR & KL+FR & \textbf{100.0} & \textbf{0.0} \\
KL+FR+NSS & KL+FR & \textbf{100.0} & \textbf{0.0} \\
\midrule
NSS & KL+Gini & \textbf{99.7} & \textbf{0.6} \\
KL+Gini & KL+Gini & 96.3 & 7.9 \\
KL+Gini+NSS & KL+Gini & 96.4 & 4.1 \\
\midrule
NSS & FR+Gini & \textbf{99.6} & \textbf{0.6} \\
FR+Gini & FR+Gini & 95.1 & 9.6 \\
FR+Gini+NSS & FR+Gini & 96.1 & 4.2 \\
\midrule
NSS & ACE+KL+FR & 99.7 & 0.6 \\
ACE+KL+FR & ACE+KL+FR & \textbf{100.0} & \textbf{0.0} \\
ACE+KL+FR+NSS & ACE+KL+FR & \textbf{100.0} & \textbf{0.0} \\
\midrule
NSS & ACE+KL+Gini & \textbf{99.7} & \textbf{0.6} \\
ACE+KL+Gini & ACE+KL+Gini & 96.7 & 7.6 \\
ACE+KL+Gini+NSS & ACE+KL+Gini & 96.8 & 3.7 \\
\midrule
NSS & KL+FR+Gini & \textbf{99.6} & \textbf{0.6} \\
KL+FR+Gini & KL+FR+Gini & 95.9 & 8.6 \\
KL+FR+Gini+NSS & KL+FR+Gini & 96.1 & 4.3 \\
\midrule
NSS & ACE+KL+FR+Gini & \textbf{99.6} & \textbf{0.6} \\
ACE+KL+FR+Gini & ACE+KL+FR+Gini & 96.3 & 8.2 \\
ACE+KL+FR+Gini+NSS & ACE+KL+FR+Gini & 96.5 & 3.8 \\
\bottomrule
\end{tabular}
}
\label{tab:03125}
\end{minipage}
\hfill
\begin{minipage}{0.47\columnwidth}
\caption{Theoretical framework setting: CIFAR10, PGD algorithm, $\varepsilon=0.5$, L$_\infty$ norm.}
\centering
\resizebox{\columnwidth}{!}{%
\begin{tabular}{llcc}
\toprule
Pool of detectors & Attacks & \auc & \fpr \\
\midrule
NSS & ACE+KL & 99.7 & 0.6 \\
ACE+KL & ACE+KL & \textbf{100.0} & \textbf{0.0} \\
ACE+KL+NSS & ACE+KL & \textbf{100.0} & \textbf{0.0} \\
\midrule
NSS & ACE+FR & 99.7 & 0.6 \\
ACE+FR & ACE+FR & \textbf{100.0} & \textbf{0.0} \\
ACE+FR+NSS & ACE+FR & \textbf{100.0} & \textbf{0.0} \\
\midrule
NSS & ACE+Gini & \textbf{99.7} & \textbf{0.6} \\
ACE+Gini & ACE+Gini & 97.0 & 7.4 \\
ACE+Gini+NSS & ACE+Gini & 97.5 & 3.0 \\
\midrule
NSS & KL+FR & 99.7 & 0.6 \\
KL+FR & KL+FR & \textbf{100.0} & \textbf{0.0} \\
KL+FR+NSS & KL+FR & \textbf{100.0} & \textbf{0.0} \\
\midrule
NSS & KL+Gini & \textbf{99.6} & \textbf{0.6} \\
KL+Gini & KL+Gini & 96.5 & 7.7 \\
KL+Gini+NSS & KL+Gini & 96.7 & 3.7 \\
\midrule
NSS & FR+Gini & \textbf{99.6} & \textbf{0.6} \\
FR+Gini & FR+Gini & 95.1 & 9.6 \\
FR+Gini+NSS & FR+Gini & 96.1 & 4.2 \\
\midrule
NSS & ACE+KL+FR & 99.7 & 0.6 \\
ACE+KL+FR & ACE+KL+FR & \textbf{100.0} & \textbf{0.0} \\
ACE+KL+FR+NSS & ACE+KL+FR & \textbf{100.0} & \textbf{0.0} \\
\midrule
NSS & ACE+KL+Gini & \textbf{99.6} & \textbf{0.6} \\
ACE+KL+Gini & ACE+KL+Gini & 96.9 & 7.4 \\
ACE+KL+Gini+NSS & ACE+KL+Gini & 97.2 & 3.3 \\
\midrule
NSS & KL+FR+Gini & \textbf{99.6} & \textbf{0.6} \\
KL+FR+Gini & KL+FR+Gini & 96.1 & 8.4 \\
KL+FR+Gini+NSS & KL+FR+Gini & 96.5 & 3.9 \\
\midrule
NSS & ACE+KL+FR+Gini & \textbf{99.6} & \textbf{0.6} \\
ACE+KL+FR+Gini & ACE+KL+FR+Gini & 96.5 & 8.1 \\
ACE+KL+FR+Gini+NSS & ACE+KL+FR+Gini & 96.9 & 3.4 \\
\bottomrule
\end{tabular}
}
\label{tab:05}
\end{minipage}
\end{table}
\begin{table}[!htb]
\begin{minipage}{0.47\columnwidth}
\caption{Theoretical framework setting: SVHN, PGD algorithm, $\varepsilon=0.03125$, L$_\infty$ norm.}
\centering
\resizebox{\columnwidth}{!}{%
\begin{tabular}{llcc}
\toprule
Pool of detectors & Attacks & \auc & \fpr \\
\midrule
NSS & ACE+KL & 90.0 & 0.9 \\
ACE+KL & ACE+KL & 97.4 & 13.5 \\
ACE+KL+NSS & ACE+KL & \textbf{99.3} & \textbf{0.5} \\
\midrule
NSS & ACE+FR & 90.0 & 0.9 \\
ACE+FR & ACE+FR & 97.4 & 12.9 \\
ACE+FR+NSS & ACE+FR & \textbf{99.2} & \textbf{0.5} \\
\midrule
NSS & ACE+Gini & 89.0 & \textbf{1.9} \\
ACE+Gini & ACE+Gini & \textbf{91.6} & 24.9 \\
ACE+Gini+NSS & ACE+Gini & 90.2 & 14.0 \\
\midrule
NSS & KL+FR & 90.0 & 0.9 \\
KL+FR & KL+FR & 97.6 & 11.2 \\
KL+FR+NSS & KL+FR & \textbf{99.3} & \textbf{0.5} \\
\midrule
NSS & KL+Gini & 89.0 & \textbf{1.9} \\
KL+Gini & KL+Gini & 92.8 & 15.0 \\
KL+Gini+NSS & KL+Gini & \textbf{95.4} & 8.9 \\
\midrule
NSS & FR+Gini & 89.1 & \textbf{1.9} \\
FR+Gini & FR+Gini & 92.5 & 15.3 \\
FR+Gini+NSS & FR+Gini & \textbf{95.1} & 9.0 \\
\midrule
NSS & ACE+KL+FR & 89.9 & 1.0 \\
ACE+KL+FR & ACE+KL+FR & 97.3 & 14.7 \\
ACE+KL+FR+NSS & ACE+KL+FR & \textbf{99.2} & \textbf{0.6} \\
\midrule
NSS & ACE+KL+Gini & 88.9 & \textbf{2.0} \\
ACE+KL+Gini & ACE+KL+Gini & \textbf{91.8} & 25.7 \\
ACE+KL+Gini+NSS & ACE+KL+Gini & 89.3 & 14.8 \\
\midrule
NSS & KL+FR+Gini & 89.0 & \textbf{2.0} \\
KL+FR+Gini & KL+FR+Gini & \textbf{92.5} & 18.7 \\
KL+FR+Gini+NSS & KL+FR+Gini & 91.9 & 12.3 \\
\midrule
NSS & ACE+KL+FR+Gini & 88.8 & \textbf{2.1} \\
ACE+KL+FR+Gini & ACE+KL+FR+Gini & \textbf{91.9} & 25.9 \\
ACE+KL+FR+Gini+NSS & ACE+KL+FR+Gini & 88.9 & 15.2 \\
\bottomrule
\end{tabular}
}
\label{tab:svhn003125}
\end{minipage}
\hfill
\begin{minipage}{0.47\columnwidth}
\caption{Theoretical framework setting: SVHN, PGD algorithm, $\varepsilon=0.0625$, L$_\infty$ norm.}
\centering
\resizebox{\columnwidth}{!}{%
\begin{tabular}{llcc}
\toprule
Pool of detectors & Attacks & \auc & \fpr \\
\midrule
NSS & ACE+KL & 91.3 & \textbf{0.0} \\
ACE+KL & ACE+KL & 99.9 & \textbf{0.0} \\
ACE+KL+NSS & ACE+KL & \textbf{100.0} & \textbf{0.0} \\
\midrule
NSS & ACE+FR & 91.3 & \textbf{0.0} \\
ACE+FR & ACE+FR & 99.9 & \textbf{0.0} \\
ACE+FR+NSS & ACE+FR & \textbf{100.0} & \textbf{0.0} \\
\midrule
NSS & ACE+Gini & 91.2 & \textbf{0.0} \\
ACE+Gini & ACE+Gini & \textbf{94.7} & 6.0 \\
ACE+Gini+NSS & ACE+Gini & 89.5 & 10.9 \\
\midrule
NSS & KL+FR & 91.3 & \textbf{0.0} \\
KL+FR & KL+FR & 99.9 & \textbf{0.0} \\
KL+FR+NSS & KL+FR & \textbf{100.0} & \textbf{0.0} \\
\midrule
NSS & KL+Gini & 91.2 & \textbf{0.0} \\
KL+Gini & KL+Gini & 94.4 & 6.1 \\
KL+Gini+NSS & KL+Gini & \textbf{99.6} & 0.9 \\
\midrule
NSS & FR+Gini & 91.2 & \textbf{0.0} \\
FR+Gini & FR+Gini & 93.6 & 7.0 \\
FR+Gini+NSS & FR+Gini & \textbf{94.5} & 10.0 \\
\midrule
NSS & ACE+KL+FR & 91.3 & \textbf{0.0} \\
ACE+KL+FR & ACE+KL+FR & 99.9 & \textbf{0.0} \\
ACE+KL+FR+NSS & ACE+KL+FR & \textbf{100.0} & \textbf{0.0} \\
\midrule
NSS & ACE+KL+Gini & 91.2 & \textbf{0.0} \\
ACE+KL+Gini & ACE+KL+Gini & \textbf{95.0} & 5.7 \\
ACE+KL+Gini+NSS & ACE+KL+Gini & 87.9 & 12.4 \\
\midrule
NSS & KL+FR+Gini & 91.2 & \textbf{0.0} \\
KL+FR+Gini & KL+FR+Gini & \textbf{94.4} & 6.0 \\
KL+FR+Gini+NSS & KL+FR+Gini & 92.4 & 12.4 \\
\midrule
NSS & ACE+KL+FR+Gini & 91.2 & \textbf{0.0} \\
ACE+KL+FR+Gini & ACE+KL+FR+Gini & \textbf{94.6} & 6.0 \\
ACE+KL+FR+Gini+NSS & ACE+KL+FR+Gini & 87.4 & 13.0 \\
\bottomrule
\end{tabular}
}
\label{tab:svhn00625}
\end{minipage}
\end{table}

\begin{table}[!htb]
\begin{minipage}{0.47\columnwidth}
\caption{Theoretical framework setting: SVHN, PGD algorithm, $\varepsilon=0.125$, L$_\infty$ norm.}
\centering
\resizebox{\columnwidth}{!}{%
\begin{tabular}{llcc}
\toprule
Pool of detectors & Attacks & \auc & \fpr \\
\midrule
NSS & ACE+KL & 91.3 & \textbf{0.0} \\
ACE+KL & ACE+KL & \textbf{100.0} & \textbf{0.0} \\
ACE+KL+NSS & ACE+KL & \textbf{100.0} & \textbf{0.0} \\
\midrule
NSS & ACE+FR & 91.3 & \textbf{0.0} \\
ACE+FR & ACE+FR & \textbf{100.0} & \textbf{0.0} \\
ACE+FR+NSS & ACE+FR & \textbf{100.0} & \textbf{0.0} \\
\midrule
NSS & ACE+Gini & 91.3 & \textbf{0.0} \\
ACE+Gini & ACE+Gini & 94.8 & 5.6 \\
ACE+Gini+NSS & ACE+Gini & \textbf{99.7} & 0.4 \\
\midrule
NSS & KL+FR & 91.3 & \textbf{0.0} \\
KL+FR & KL+FR & \textbf{100.0} & \textbf{0.0} \\
KL+FR+NSS & KL+FR & \textbf{100.0} & \textbf{0.0} \\
\midrule
NSS & KL+Gini & 91.3 & \textbf{0.0} \\
KL+Gini & KL+Gini & 94.5 & 6.1 \\
KL+Gini+NSS & KL+Gini & \textbf{99.6} & 0.9 \\
\midrule
NSS & FR+Gini & 91.3 & \textbf{0.0} \\
FR+Gini & FR+Gini & 93.7 & 7.0 \\
FR+Gini+NSS & FR+Gini & \textbf{94.6} & 10.0 \\
\midrule
NSS & ACE+KL+FR & 91.3 & \textbf{0.0} \\
ACE+KL+FR & ACE+KL+FR & \textbf{100.0} & \textbf{0.0} \\
ACE+KL+FR+NSS & ACE+KL+FR & \textbf{100.0} & \textbf{0.0} \\
\midrule
NSS & ACE+KL+Gini & 91.3 & \textbf{0.0} \\
ACE+KL+Gini & ACE+KL+Gini & 95.2 & 5.4 \\
ACE+KL+Gini+NSS & ACE+KL+Gini & \textbf{97.8} & 2.5 \\
\midrule
NSS & KL+FR+Gini & 91.3 & \textbf{0.0} \\
KL+FR+Gini & KL+FR+Gini & \textbf{94.6} & 6.0 \\
KL+FR+Gini+NSS & KL+FR+Gini & 93.4 & 12.4 \\
\midrule
NSS & ACE+KL+FR+Gini & 91.3 & \textbf{0.0} \\
ACE+KL+FR+Gini & ACE+KL+FR+Gini & \textbf{94.8} & 6.0 \\
ACE+KL+FR+Gini+NSS & ACE+KL+FR+Gini & 92.5 & 13.0 \\
\bottomrule
\end{tabular}
}
\label{tab:svhn0125}
\end{minipage}
\hfill
\begin{minipage}{0.47\columnwidth}
\caption{Theoretical framework setting: SVHN, PGD algorithm, $\varepsilon=0.25$, L$_\infty$ norm.}
\centering
\resizebox{\columnwidth}{!}{%
\begin{tabular}{llcc}
\toprule
Pool of detectors & Attacks & \auc & \fpr \\
\midrule
NSS & ACE+KL & 91.3 & \textbf{0.0} \\
ACE+KL & ACE+KL & \textbf{100.0} & \textbf{0.0} \\
ACE+KL+NSS & ACE+KL & \textbf{100.0} & \textbf{0.0} \\
\midrule
NSS & ACE+FR & 91.3 & \textbf{0.0} \\
ACE+FR & ACE+FR & \textbf{100.0} & \textbf{0.0} \\
ACE+FR+NSS & ACE+FR & \textbf{100.0} & \textbf{0.0} \\
\midrule
NSS & ACE+Gini & 91.3 & \textbf{0.0} \\
ACE+Gini & ACE+Gini & 94.8 & 5.6 \\
ACE+Gini+NSS & ACE+Gini & \textbf{99.7} & 0.4 \\
\midrule
NSS & KL+FR & 91.3 & \textbf{0.0} \\
KL+FR & KL+FR & \textbf{100.0} & \textbf{0.0} \\
KL+FR+NSS & KL+FR & \textbf{100.0} & \textbf{0.0} \\
\midrule
NSS & KL+Gini & 91.3 & \textbf{0.0} \\
KL+Gini & KL+Gini & 94.5 & 6.1 \\
KL+Gini+NSS & KL+Gini & \textbf{99.6} & 0.9 \\
\midrule
NSS & FR+Gini & 91.3 & \textbf{0.0} \\
FR+Gini & FR+Gini & 93.7 & 7.0 \\
FR+Gini+NSS & FR+Gini & \textbf{94.7} & 10.0 \\
\midrule
NSS & ACE+KL+FR & 91.3 & \textbf{0.0} \\
ACE+KL+FR & ACE+KL+FR & \textbf{100.0} & \textbf{0.0} \\
ACE+KL+FR+NSS & ACE+KL+FR & \textbf{100.0} & \textbf{0.0} \\
\midrule
NSS & ACE+KL+Gini & 91.3 & \textbf{0.0} \\
ACE+KL+Gini & ACE+KL+Gini & 95.2 & 5.3 \\
ACE+KL+Gini+NSS & ACE+KL+Gini & \textbf{99.7} & 0.4 \\
\midrule
NSS & KL+FR+Gini & 91.3 & \textbf{0.0} \\
KL+FR+Gini & KL+FR+Gini & \textbf{94.6} & 6.0 \\
KL+FR+Gini+NSS & KL+FR+Gini & 93.5 & 12.4 \\
\midrule
NSS & ACE+KL+FR+Gini & 91.3 & \textbf{0.0} \\
ACE+KL+FR+Gini & ACE+KL+FR+Gini & \textbf{94.8} & 6.0 \\
ACE+KL+FR+Gini+NSS & ACE+KL+FR+Gini & 93.5 & 13.0 \\
\bottomrule
\end{tabular}
}
\label{tab:svhn025}
\end{minipage}
\end{table}

\begin{table}[!htb]
\begin{minipage}{0.47\columnwidth}
\caption{Theoretical framework setting: SVHN, PGD algorithm, $\varepsilon=0.3125$, L$_\infty$ norm.}
\centering
\resizebox{\columnwidth}{!}{%
\begin{tabular}{llcc}
\toprule
Pool of detectors & Attacks & \auc & \fpr \\
\midrule
NSS & ACE+KL & 91.3 & \textbf{0.0} \\
ACE+KL & ACE+KL & \textbf{100.0} & \textbf{0.0} \\
ACE+KL+NSS & ACE+KL & \textbf{100.0} & \textbf{0.0} \\
\midrule
NSS & ACE+FR & 91.3 & \textbf{0.0} \\
ACE+FR & ACE+FR & \textbf{100.0} & \textbf{0.0} \\
ACE+FR+NSS & ACE+FR & \textbf{100.0} & \textbf{0.0} \\
\midrule
NSS & ACE+Gini & 91.3 & \textbf{0.0} \\
ACE+Gini & ACE+Gini & 94.8 & 5.6 \\
ACE+Gini+NSS & ACE+Gini & \textbf{99.7} & 0.4 \\
\midrule
NSS & KL+FR & 91.3 & \textbf{0.0} \\
KL+FR & KL+FR & \textbf{100.0} & \textbf{0.0} \\
KL+FR+NSS & KL+FR & \textbf{100.0} & \textbf{0.0} \\
\midrule
NSS & KL+Gini & 91.3 & \textbf{0.0} \\
KL+Gini & KL+Gini & 94.5 & 6.1 \\
KL+Gini+NSS & KL+Gini & \textbf{99.6} & 0.9 \\
\midrule
NSS & FR+Gini & 91.3 & \textbf{0.0} \\
FR+Gini & FR+Gini & 93.7 & 7.0 \\
FR+Gini+NSS & FR+Gini & \textbf{94.7} & 10.0 \\
\midrule
NSS & ACE+KL+FR & 91.3 & \textbf{0.0} \\
ACE+KL+FR & ACE+KL+FR & \textbf{100.0} & \textbf{0.0} \\
ACE+KL+FR+NSS & ACE+KL+FR & \textbf{100.0} & \textbf{0.0} \\
\midrule
NSS & ACE+KL+Gini & 91.3 & \textbf{0.0} \\
ACE+KL+Gini & ACE+KL+Gini & 95.2 & 5.4 \\
ACE+KL+Gini+NSS & ACE+KL+Gini & \textbf{99.7} & 0.4 \\
\midrule
NSS & KL+FR+Gini & 91.3 & \textbf{0.0} \\
KL+FR+Gini & KL+FR+Gini & \textbf{94.6} & 6.0 \\
KL+FR+Gini+NSS & KL+FR+Gini & 93.5 & 12.4 \\
\midrule
NSS & ACE+KL+FR+Gini & 91.3 & \textbf{0.0} \\
ACE+KL+FR+Gini & ACE+KL+FR+Gini & \textbf{94.8} & 6.0 \\
ACE+KL+FR+Gini+NSS & ACE+KL+FR+Gini & 93.5 & 13.0 \\
\bottomrule
\end{tabular}
}
\label{tab:svhn03125}
\end{minipage}
\hfill
\begin{minipage}{0.47\columnwidth}
\caption{Theoretical framework setting: SVHN, PGD algorithm, $\varepsilon=0.5$, L$_\infty$ norm.}
\centering
\resizebox{\columnwidth}{!}{%
\begin{tabular}{llcc}
\toprule
Pool of detectors & Attacks & \auc & \fpr \\
\midrule
NSS & ACE+KL & 91.3 & \textbf{0.0} \\
ACE+KL & ACE+KL & \textbf{100.0} & \textbf{0.0} \\
ACE+KL+NSS & ACE+KL & \textbf{100.0} & \textbf{0.0} \\
\midrule
NSS & ACE+FR & 91.3 & \textbf{0.0} \\
ACE+FR & ACE+FR & \textbf{100.0} & \textbf{0.0} \\
ACE+FR+NSS & ACE+FR & \textbf{100.0} & \textbf{0.0} \\
\midrule
NSS & ACE+Gini & 91.3 & \textbf{0.0} \\
ACE+Gini & ACE+Gini & 94.8 & 5.6 \\
ACE+Gini+NSS & ACE+Gini & 99.7 & 0.4 \\
\midrule
NSS & KL+FR & 91.3 & \textbf{0.0} \\
KL+FR & KL+FR & \textbf{100.0} & \textbf{0.0} \\
KL+FR+NSS & KL+FR & \textbf{100.0} & \textbf{0.0} \\
\midrule
NSS & KL+Gini & 91.3 & \textbf{0.0} \\
KL+Gini & KL+Gini & 94.5 & 6.1 \\
KL+Gini+NSS & KL+Gini & \textbf{99.6} & 0.9 \\
\midrule
NSS & FR+Gini & 91.3 & \textbf{0.0} \\
FR+Gini & FR+Gini & 93.7 & 7.0 \\
FR+Gini+NSS & FR+Gini & \textbf{94.7} & 10.0 \\
\midrule
NSS & ACE+KL+FR & 91.3 & \textbf{0.0} \\
ACE+KL+FR & ACE+KL+FR & \textbf{100.0} & \textbf{0.0} \\
ACE+KL+FR+NSS & ACE+KL+FR & \textbf{100.0} & \textbf{0.0} \\
\midrule
NSS & ACE+KL+Gini & 91.3 & \textbf{0.0} \\
ACE+KL+Gini & ACE+KL+Gini & 95.2 & 5.3 \\
ACE+KL+Gini+NSS & ACE+KL+Gini & \textbf{99.7} & 0.4 \\
\midrule
NSS & KL+FR+Gini & 91.3 & \textbf{0.0} \\
KL+FR+Gini & KL+FR+Gini & \textbf{94.6} & 6.0 \\
KL+FR+Gini+NSS & KL+FR+Gini & 93.5 & 12.4 \\
\midrule
NSS & ACE+KL+FR+Gini & 91.3 & \textbf{0.0} \\
ACE+KL+FR+Gini & ACE+KL+FR+Gini & \textbf{94.8} & 6.0 \\
ACE+KL+FR+Gini+NSS & ACE+KL+FR+Gini & 93.5 & 13.0 \\
\bottomrule
\end{tabular}
}
\label{tab:svhn05}
\end{minipage}
\end{table}

\subsection{Evaluation of the Proposed Solution in the Setting of~\cite{GranesePRMP2022ECMLPKDD}}
\begin{table*}[!htbp]
\centering
\caption{Comparison between the proposed method and NSS on CIFAR10 and SVHN. The $^\star$ symbol means the perturbation mechanism is executed in parallel four times starting from the same original clean sample, each time using one of the objective losses between ACE Eq.~(3), KL Eq.~(4), FR Eq.~(5), or Eq.~(6) in~\cite{GranesePRMP2022ECMLPKDD}.}
\ra{1.3}
\resizebox{1\columnwidth}{!}{%
\begin{tabular}{@{}r|bbcbbcyycyy@{}}\toprule
& \multicolumn{5}{c}{CIFAR10} & & \multicolumn{5}{c}{SVHN} \\
\cmidrule{2-6} \cmidrule{8-12}
& \multicolumn{2}{c}{NSS} & \phantom{abc}& \multicolumn{2}{c}{Ours} &
  \phantom{abc} & \multicolumn{2}{c}{NSS} & \phantom{abc}& \multicolumn{2}{c}{Ours} \\ \cmidrule{2-3}
\cmidrule{5-6} \cmidrule{8-9} \cmidrule{11-12}
  & \auc & \fpr  && \auc & \fpr && \auc & \fpr && \auc & \fpr\\ 
 \midrule

\textbf{Norm L$_1$}\\

\underline{PGD1$^\star$}\\
$\varepsilon=5$ & 
48.5 & 94.2 && 
\textbf{62.1} & \textbf{87.1} &&
40.2 & 91.3 && 
\textbf{76.9} & \textbf{78.9}\\
$\varepsilon=10$ &
54.0 & \textbf{90.3} &&
\textbf{56.8} & 90.4 &&
36.9 & 91.3 &&
\textbf{73.0} & \textbf{82.7} \\
$\varepsilon=15$ & 
58.8 & 86.8 &&
\textbf{69.3} & \textbf{84.4} &&
35.6 & 91.3&&
\textbf{78.9} & \textbf{72.5} \\
$\varepsilon=20$ & 
63.5 & 82.3 && 
\textbf{78.7} & \textbf{73.1} &&
36.1 & 91.3 &&
\textbf{83.6} & \textbf{60.6}\\
$\varepsilon=25$ & 
67.7 & 77.2 && 
\textbf{87.1} & \textbf{50.8} &&
37.8 & 91.3 &&
\textbf{87.0} & \textbf{48.7}\\
$\varepsilon=30$ &
71.4 & 73.4 &&
\textbf{90.3} & \textbf{35.4} &&
39.8 & 91.3 &&
\textbf{89.3} & \textbf{37.2}\\
$\varepsilon=40$ & 
76.1 & 67.3 && 
\textbf{92.1} & \textbf{26.4} &&
43.1 & 91.3 &&
\textbf{92.7} & \textbf{20.0}\\
\midrule

\textbf{Norm L$_2$}\\

\underline{PGD2$^\star$}\\
$\varepsilon=0.125$ & 
48.3 & 94.3 &&
\textbf{63.9} & \textbf{85.3} &&
40.8 & 91.3 &&
\textbf{80.2} & \textbf{74.6}\\
$\varepsilon=0.25$ & 
53.2 & 91.2 && 
\textbf{57.2} & \textbf{90.3} &&
37.2 & 91.3 &&
\textbf{74.1} & \textbf{81.6}\\
$\varepsilon=0.3125$ & 
55.8 & 89.2 && 
\textbf{61.0} & \textbf{88.8} &&
36.1 & 91.3 &&
\textbf{75.1} & \textbf{79.7}\\
$\varepsilon=0.5$ & 
63.3 & 82.6 && 
\textbf{79.4} & \textbf{73.0}&&
35.9 & 91.3 &&
\textbf{82.5} &\textbf{64.4}\\
$\varepsilon=1$ & 
76.4 & 67.5 && 
\textbf{91.4} & \textbf{26.3}&&
42.5 & 91.3 &&
\textbf{92.3} & \textbf{24.6}\\
$\varepsilon=1.5$ & 
81.0 & 63.0 && 
\textbf{91.9} & \textbf{24.3} &&
46.3 & 91.3 &&
\textbf{94.1} & \textbf{7.5}\\
$\varepsilon=2$ & 
82.6 & 62.3 &&
\textbf{91.9} & \textbf{24.2} &&
49.8 & 91.3 &&
\textbf{94.9} & \textbf{5.3}\\

\underline{DeepFool}\\
No $\varepsilon$ & 
57.0 & 91.7 &&
\textbf{82.3} & \textbf{53.7}&&
41.3 & 91.3 &&
\textbf{94.9} & \textbf{12.0}\\

\underline{CW2}\\
$\varepsilon=0.01$ & 
\textbf{56.4} & \textbf{90.8} &&
53.4 & 92.3 &&
41.0 & \textbf{91.3} &&
\textbf{54.2} & 92.0\\

\underline{HOP}\\
$\varepsilon=0.1$ & 
66.1 & 87.0 &&
\textbf{86.1} & \textbf{49.0}&&
67.6 & 84.2 &&
\textbf{96.0} & \textbf{10.2}\\
\midrule

\textbf{Norm L$_\infty$}\\
\underline{PGDi$^\star$, FGSM$^\star$, BIM$^\star$}\\\
$\varepsilon=0.03125$ & 
\textbf{83.0} & \textbf{55.3} &&
82.3 & 59.8&&
\textbf{86.3} & \textbf{46.9} &&
81.4 & 64.8\\
$\varepsilon=0.0625$ & 
\textbf{96.0} & \textbf{17.2} &&
92.0 & 29.6 &&
88.9 & \textbf{0.7} &&
\textbf{89.1} & 33.3\\
$\varepsilon=0.25$ &
\textbf{97.3} & \textbf{0.6} &&
95.9 & 8.8 &&
51.6 & 88.9 &&
\textbf{92.3} & \textbf{16.4}\\
$\varepsilon=0.5$ & 
82.5 & 100.0 &&
\textbf{94.6} & \textbf{9.7} &&
46.7 & 86.7 &&
\textbf{92.9} & \textbf{14.4}\\

\underline{PGDi$^\star$, FGSM$^\star$, BIM$^\star$, SA}\\
$\varepsilon=0.125$ & 
9.4 & 99.9 &&
\textbf{88.9} & \textbf{40.8} &&
32.9 & 91.3 && 
\textbf{89.2} & \textbf{29.0} \\

\underline{PGDi$^\star$, FGSM$^\star$, BIM$^\star$, CWi}\\
$\varepsilon=0.3125$ &
63.2 & 99.1 &&
\textbf{80.0} & \textbf{61.2} &&
41.3 & 91.3 &&
\textbf{88.2} & \textbf{33.1}\\
\midrule
\textbf{No norm}\\
\underline{STA}\\
No $\varepsilon$ & 
\textbf{88.5} & \textbf{38.8} && 
82.7 & 52.4 &&
\textbf{91.2} & \textbf{0.2} &&
90.2 & 23.2\\
\bottomrule
\end{tabular}
}
\label{tab:final_table}
\end{table*}

\subsubsection{Ablation Study}
\label{app:res_mead_ablation}
\begin{figure*}
	\centering
		\begin{subfigure}[b]{.25\columnwidth}
		\centering
		\includegraphics[width=\columnwidth]{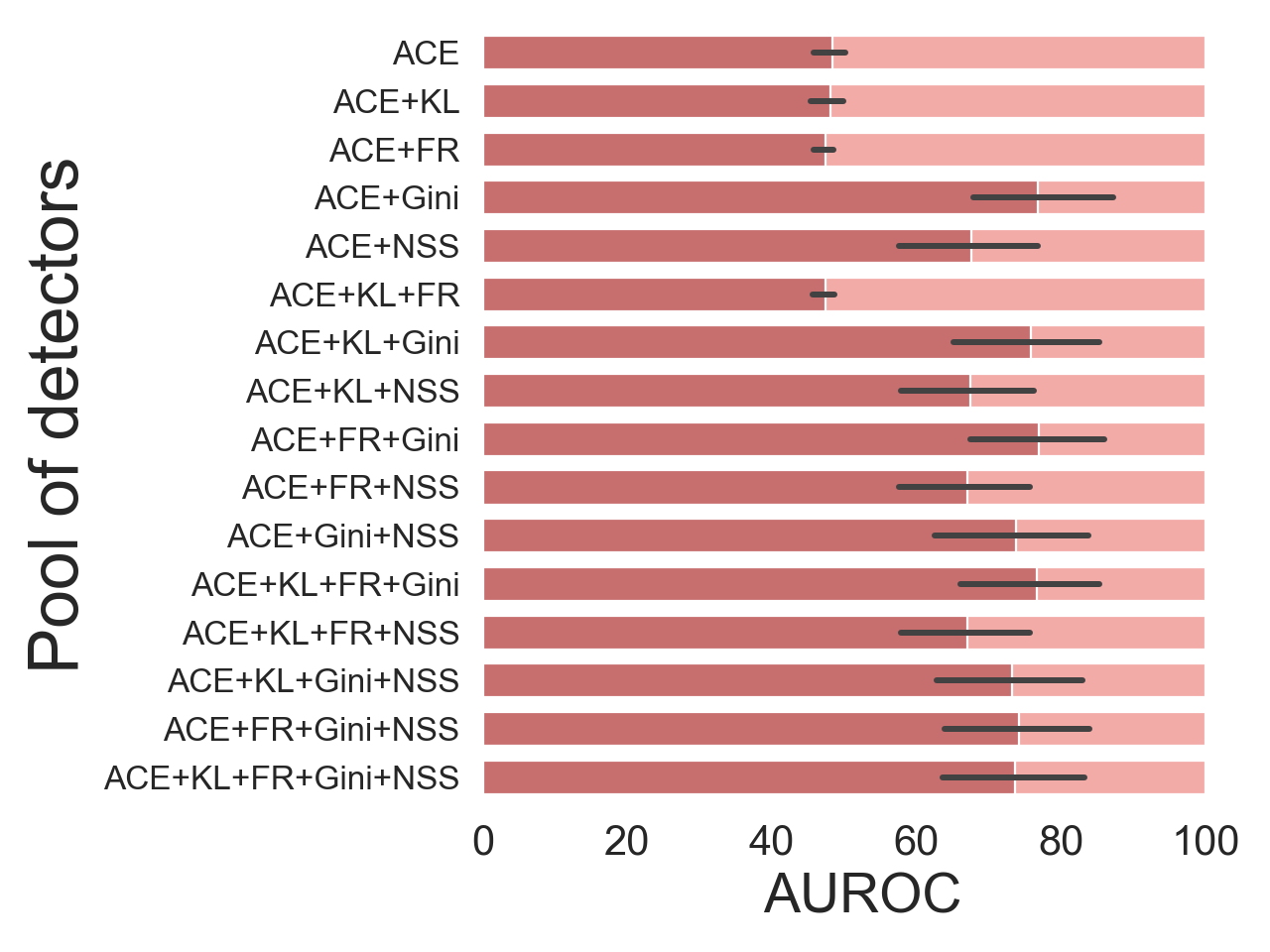}
		\vspace{-1.5\baselineskip}
		\caption{Attacks created with L$_1$}
	\end{subfigure}
	    \hfill
		\begin{subfigure}[b]{.25\columnwidth}
		\centering
		\includegraphics[width=\columnwidth]{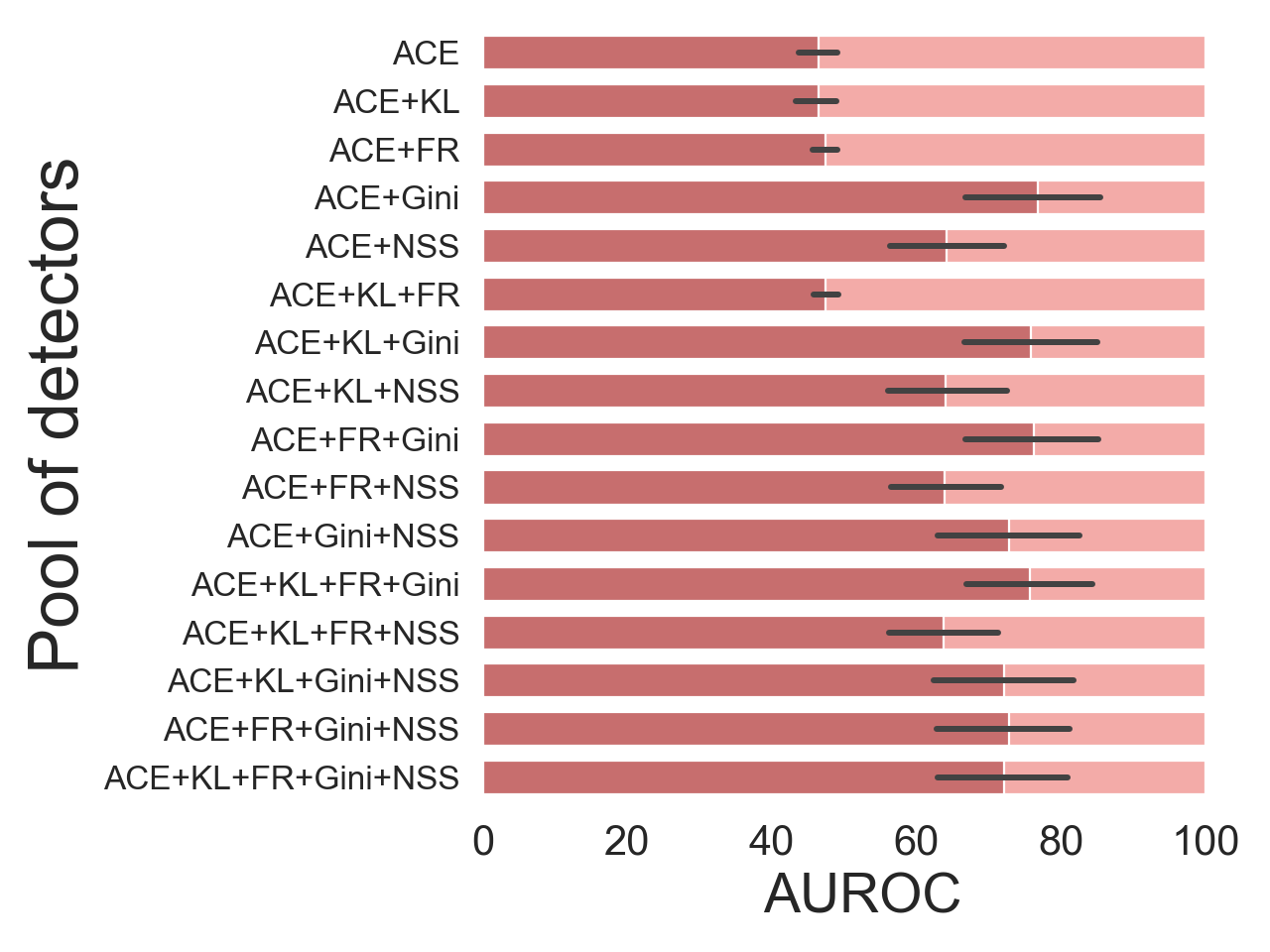}
		\vspace{-1.5\baselineskip}
		\caption{Attacks created with L$_2$}
	\end{subfigure}
	 \hfill
	\begin{subfigure}[b]{.25\columnwidth}
	    \centering
	    \includegraphics[width=\columnwidth]{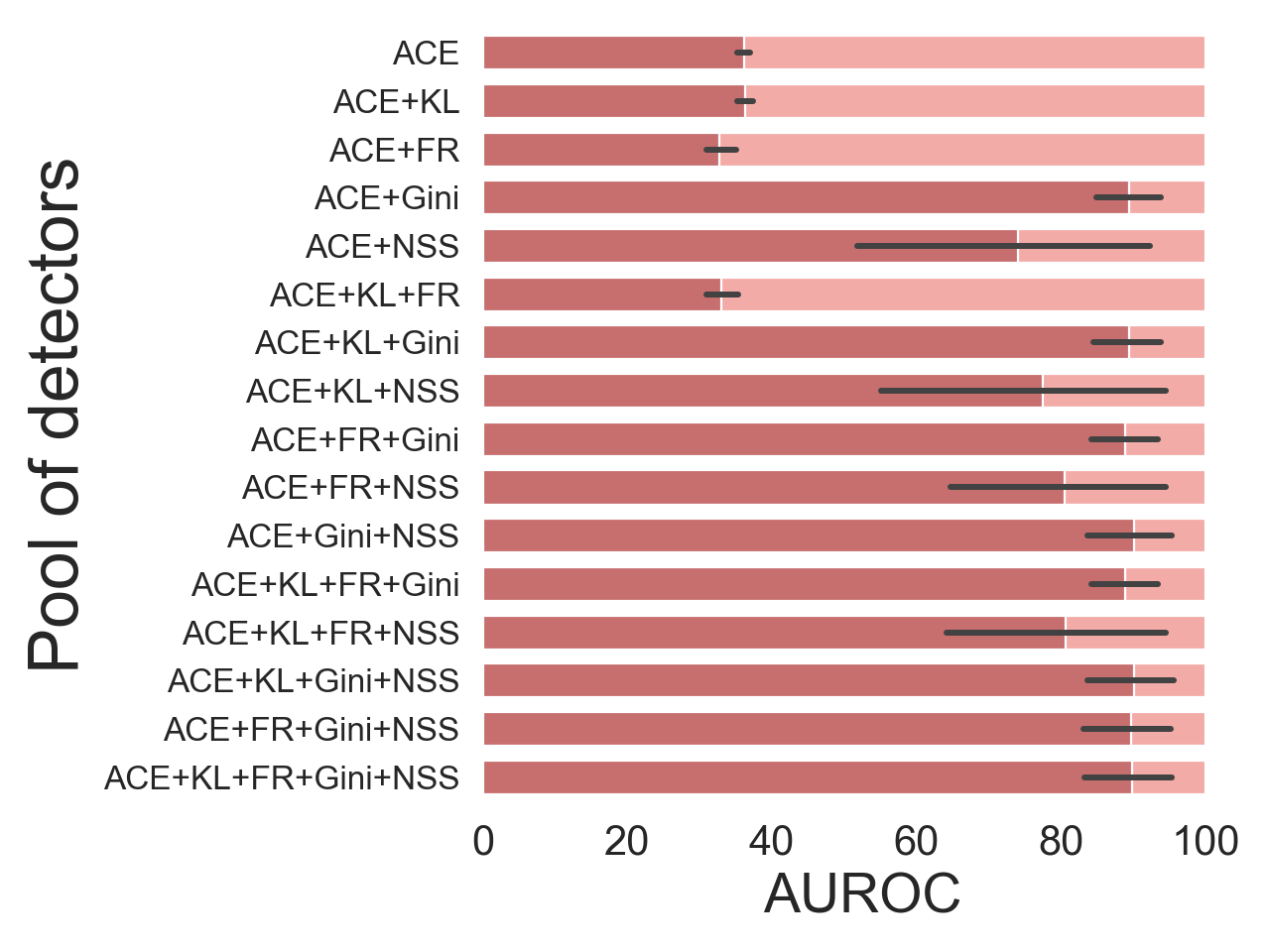}
		\vspace{-1.5\baselineskip}
		\caption{Attacks created with L$_\infty$}
	\end{subfigure}
    \hfill
    	\centering
		\begin{subfigure}[b]{.25\columnwidth}
		\centering
		\includegraphics[width=\columnwidth]{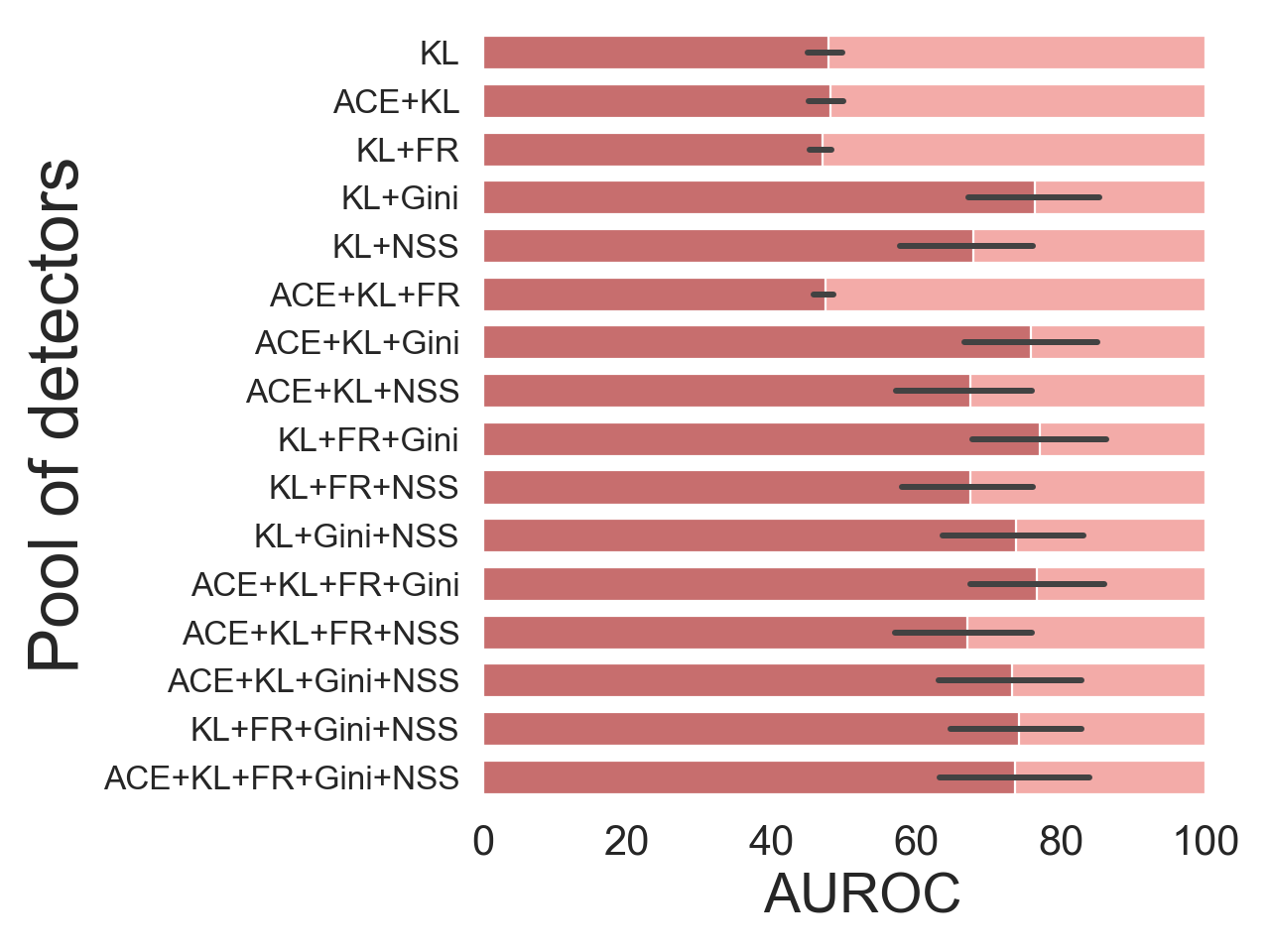}
		\vspace{-1.5\baselineskip}
		\caption{Attacks created with L$_1$}
	\end{subfigure}
	    \hfill
		\begin{subfigure}[b]{.25\columnwidth}
		\centering
		\includegraphics[width=\columnwidth]{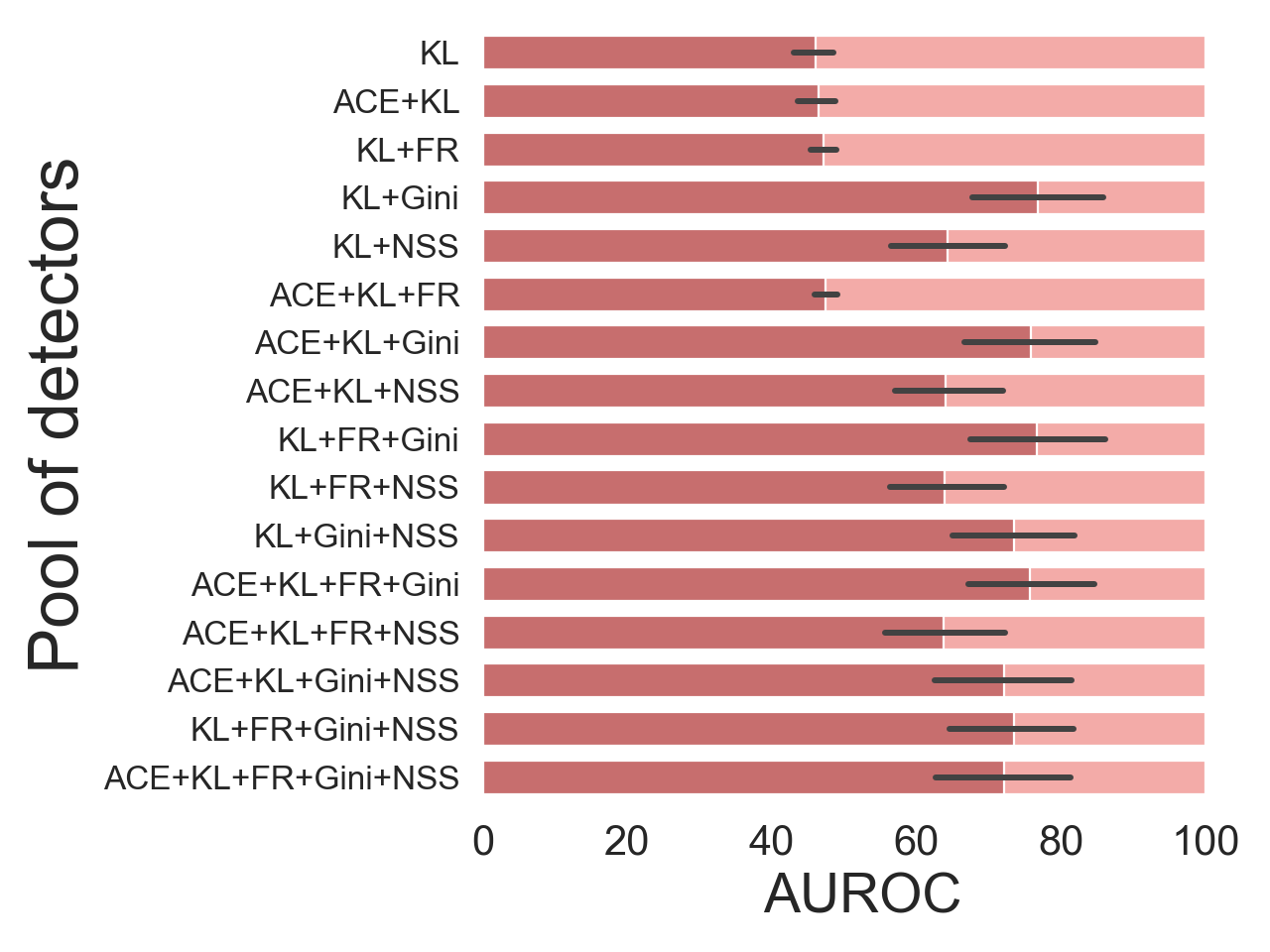}
		\vspace{-1.5\baselineskip}
		\caption{Attacks created with L$_2$}
	\end{subfigure}
	 \hfill
	\begin{subfigure}[b]{.25\columnwidth}
	    \centering
	    \includegraphics[width=\columnwidth]{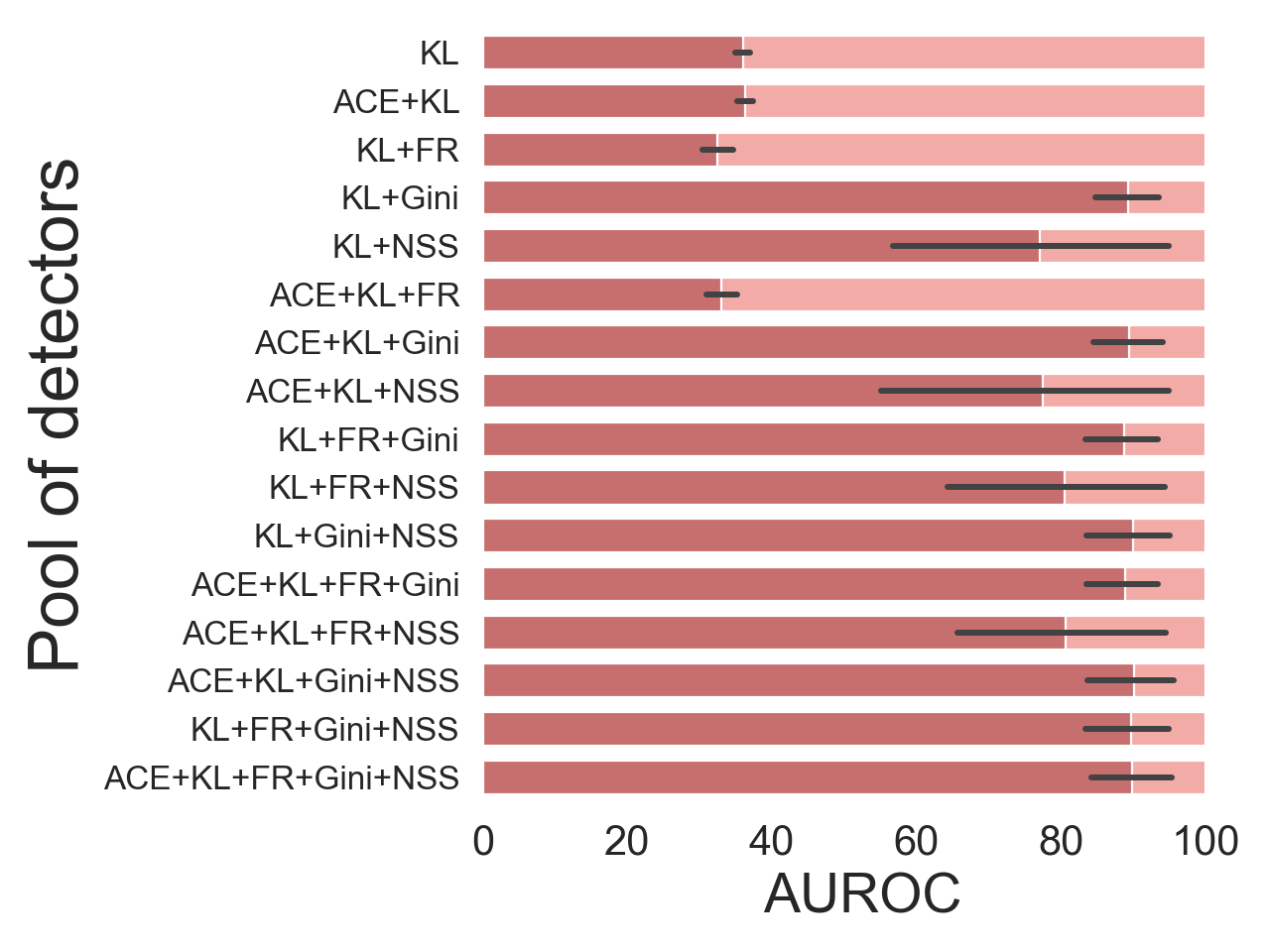}
		\vspace{-1.5\baselineskip}
		\caption{Attacks created with L$_\infty$}
	\end{subfigure}
    \hfill
 	\centering
		\begin{subfigure}[b]{.25\columnwidth}
		\centering
		\includegraphics[width=\columnwidth]{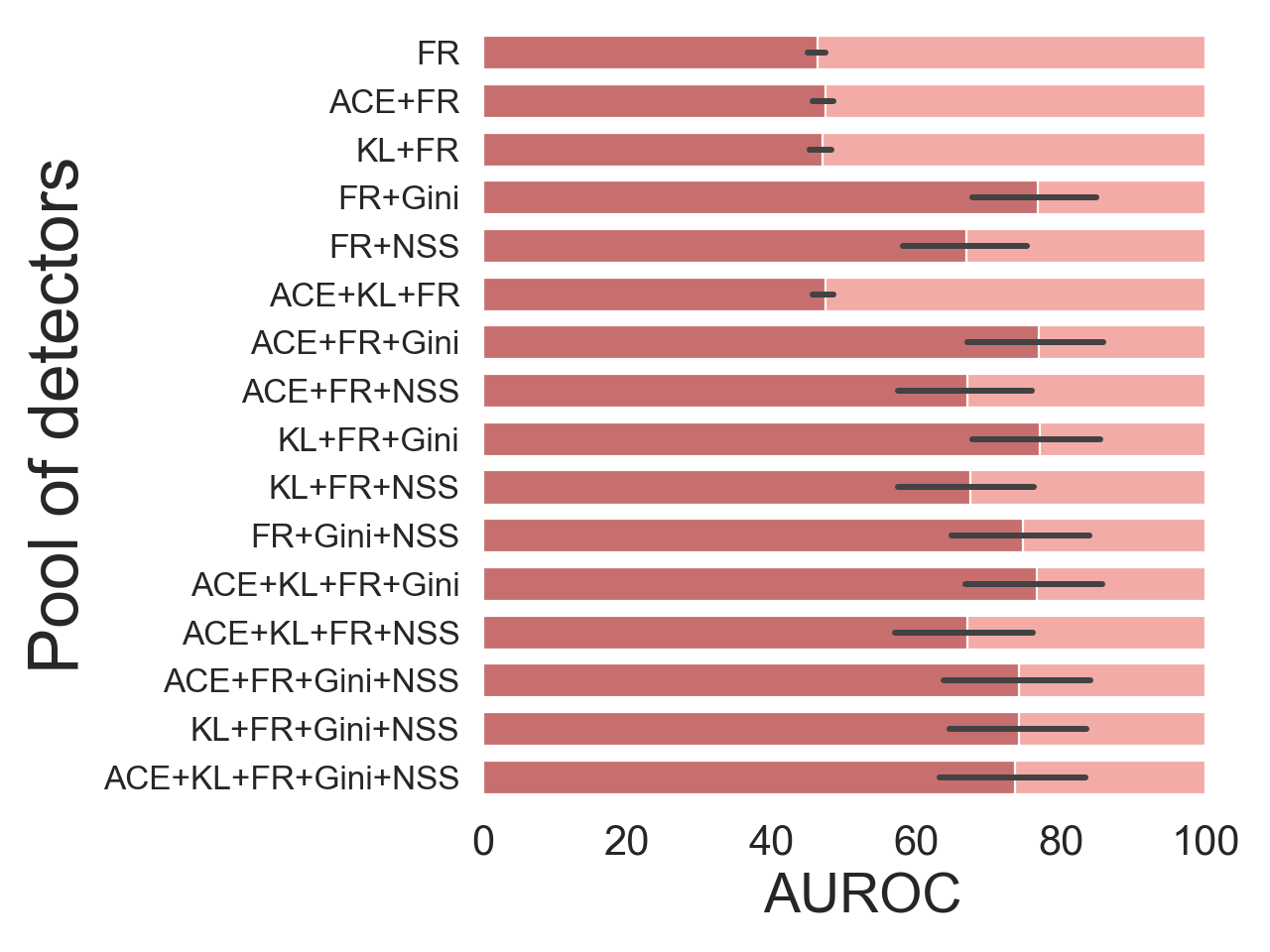}
		\vspace{-1.5\baselineskip}
		\caption{Attacks created with L$_1$}
	\end{subfigure}
	    \hfill
		\begin{subfigure}[b]{.25\columnwidth}
		\centering
		\includegraphics[width=\columnwidth]{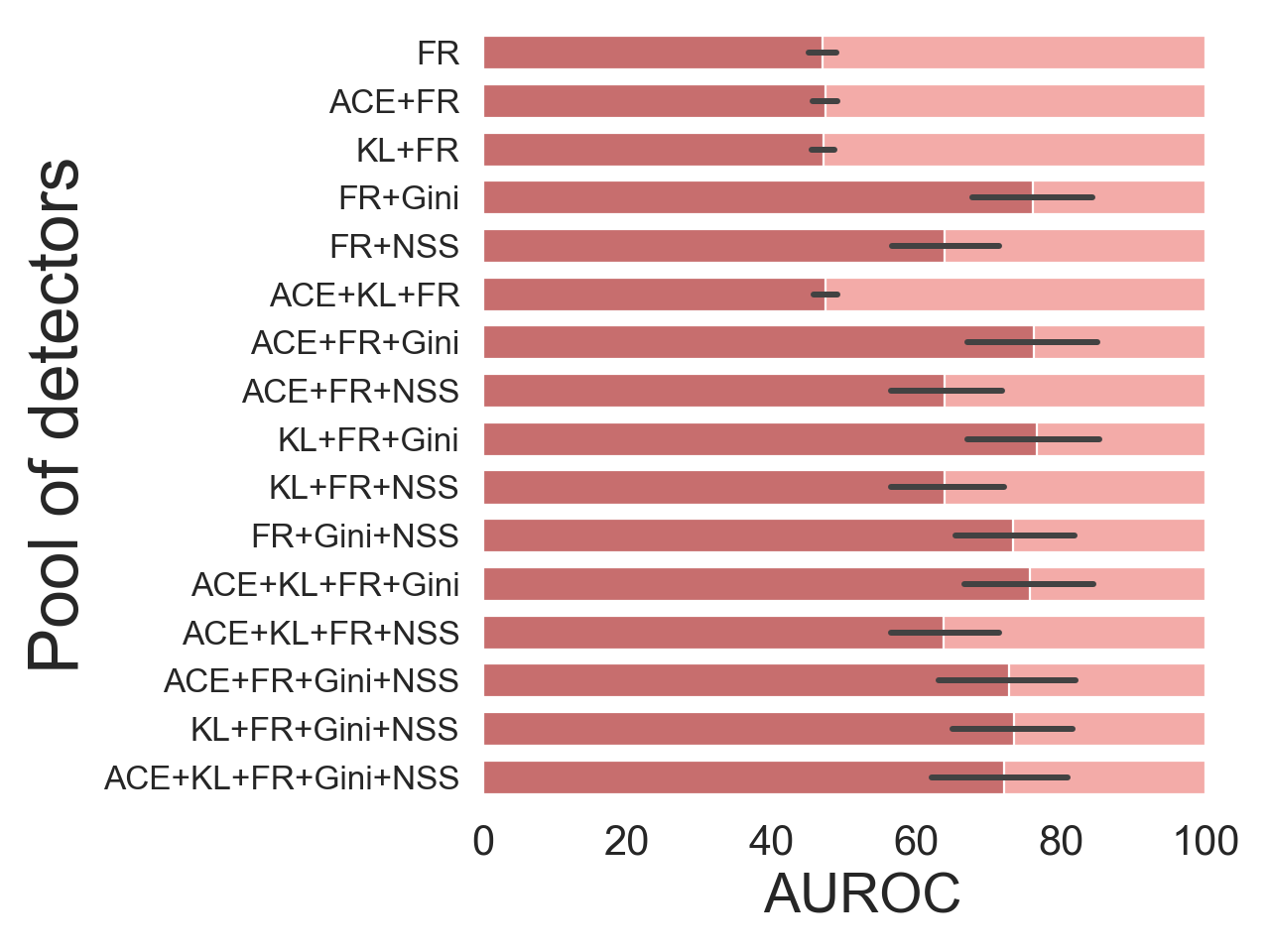}
		\vspace{-1.5\baselineskip}
		\caption{Attacks created with L$_2$}
	\end{subfigure}
	 \hfill
	\begin{subfigure}[b]{.25\columnwidth}
	    \centering
	    \includegraphics[width=\columnwidth]{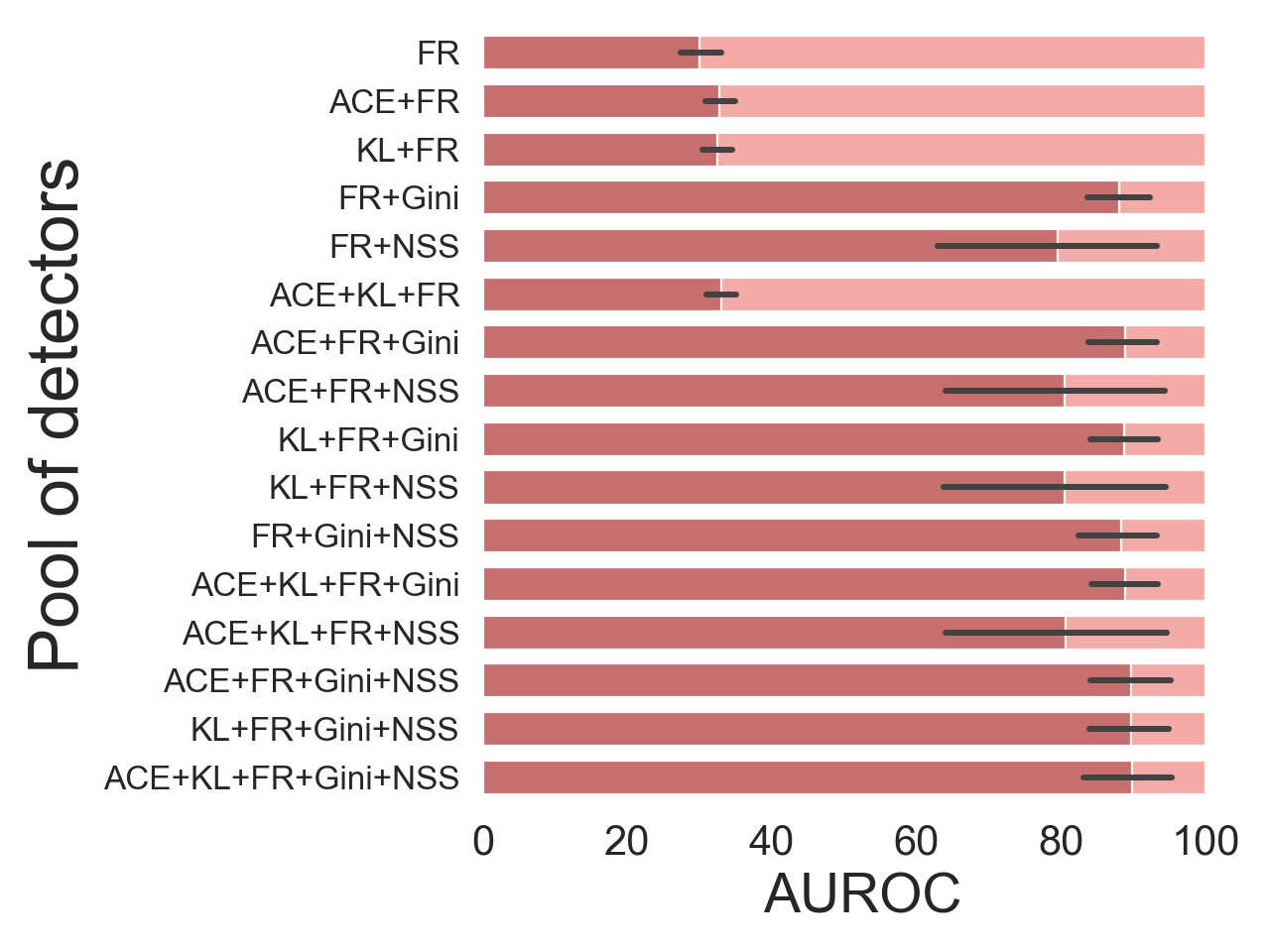}
		\vspace{-1.5\baselineskip}
		\caption{Attacks created with L$_\infty$}
	\end{subfigure}
    \hfill
    	\centering
		\begin{subfigure}[b]{.25\columnwidth}
		\centering
		\includegraphics[width=\columnwidth]{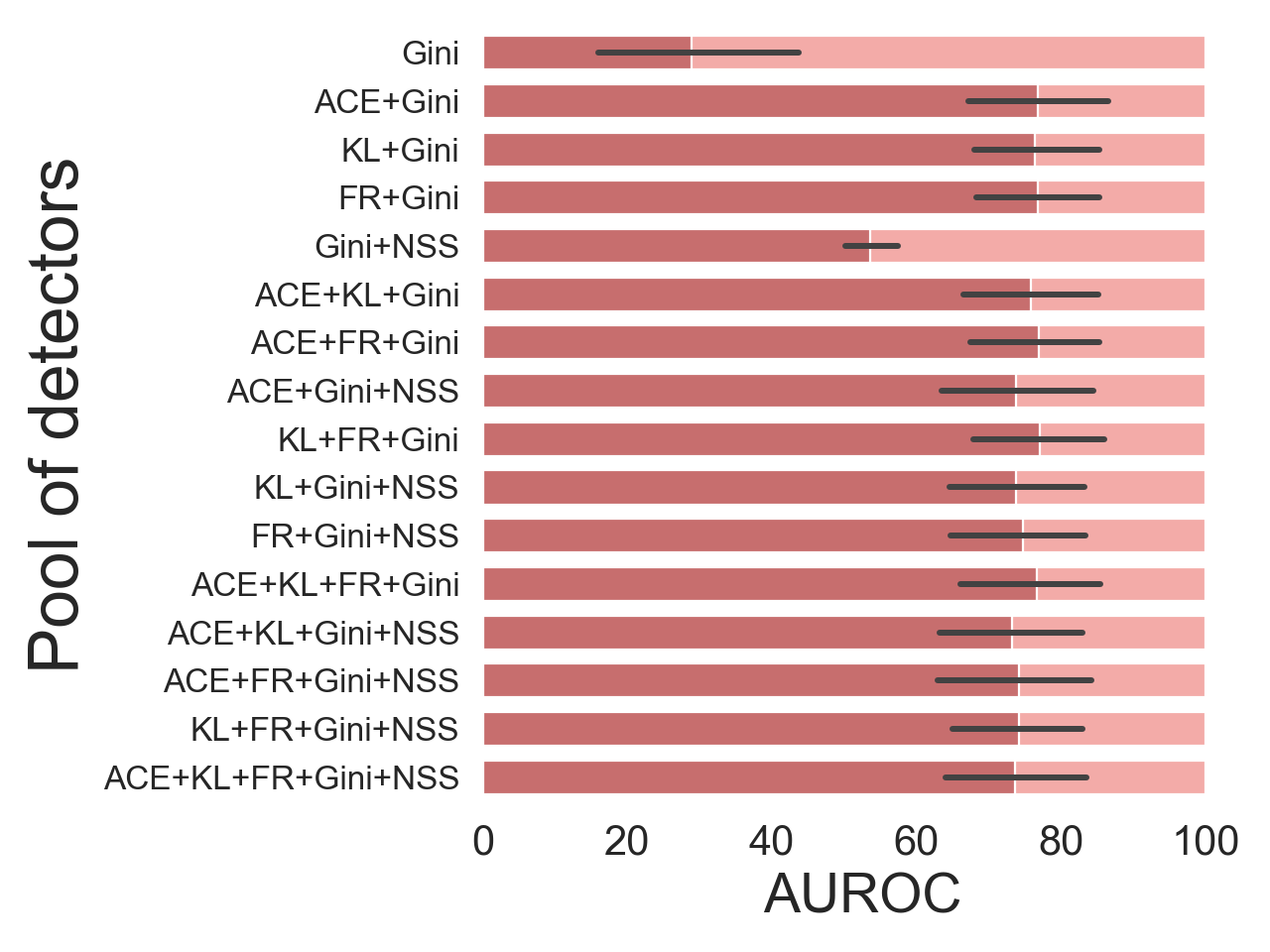}
		\vspace{-1.5\baselineskip}
		\caption{Attacks created with L$_1$}
  \label{fig:a_Gini1}
	\end{subfigure}
	    \hfill
		\begin{subfigure}[b]{.25\columnwidth}
		\centering
		\includegraphics[width=\columnwidth]{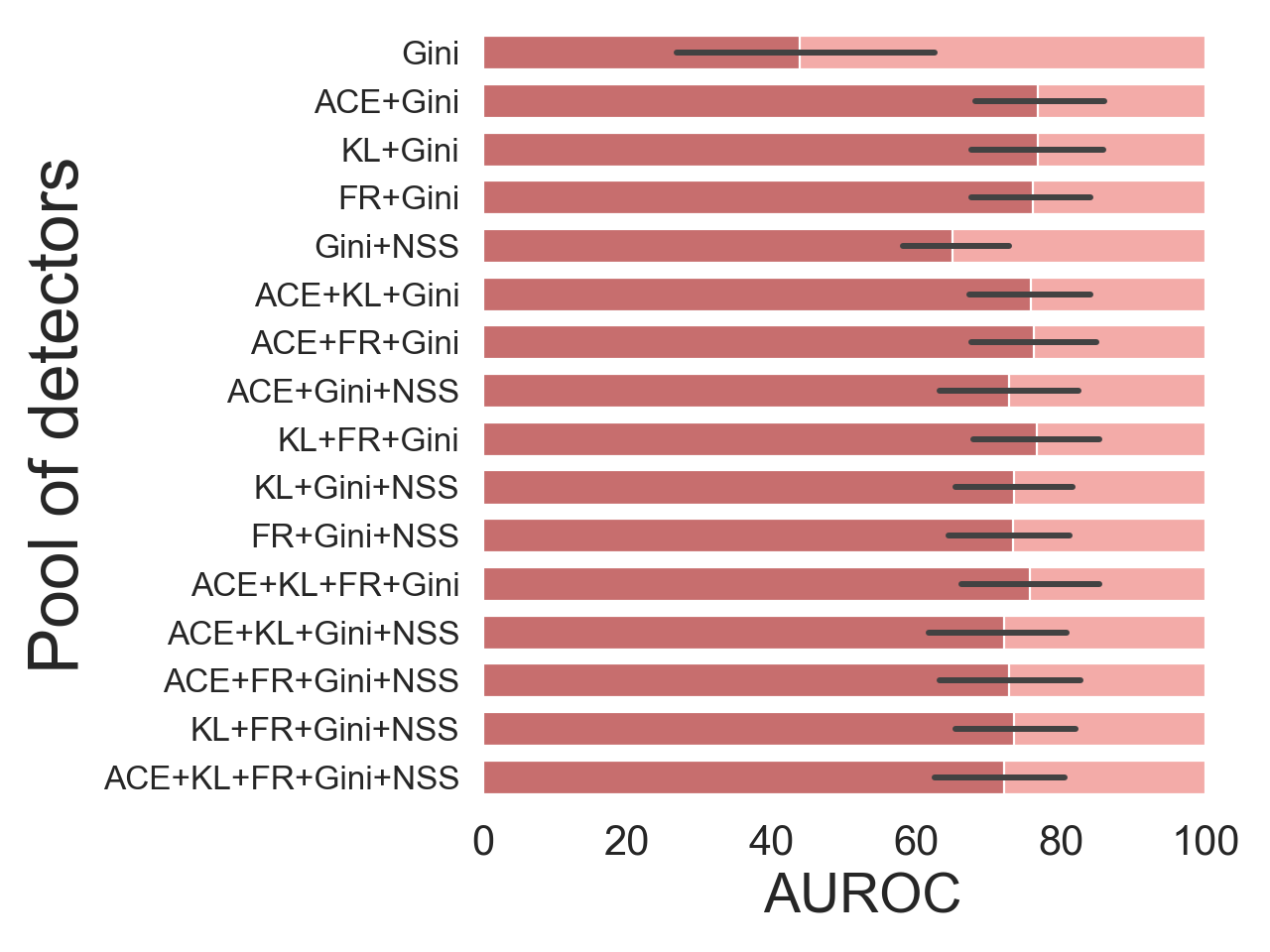}
		\vspace{-1.5\baselineskip}
		\caption{Attacks created with L$_2$}
    \label{fig:a_Gini2}

	\end{subfigure}
	 \hfill
	\begin{subfigure}[b]{.25\columnwidth}
	    \centering
	    \includegraphics[width=\columnwidth]{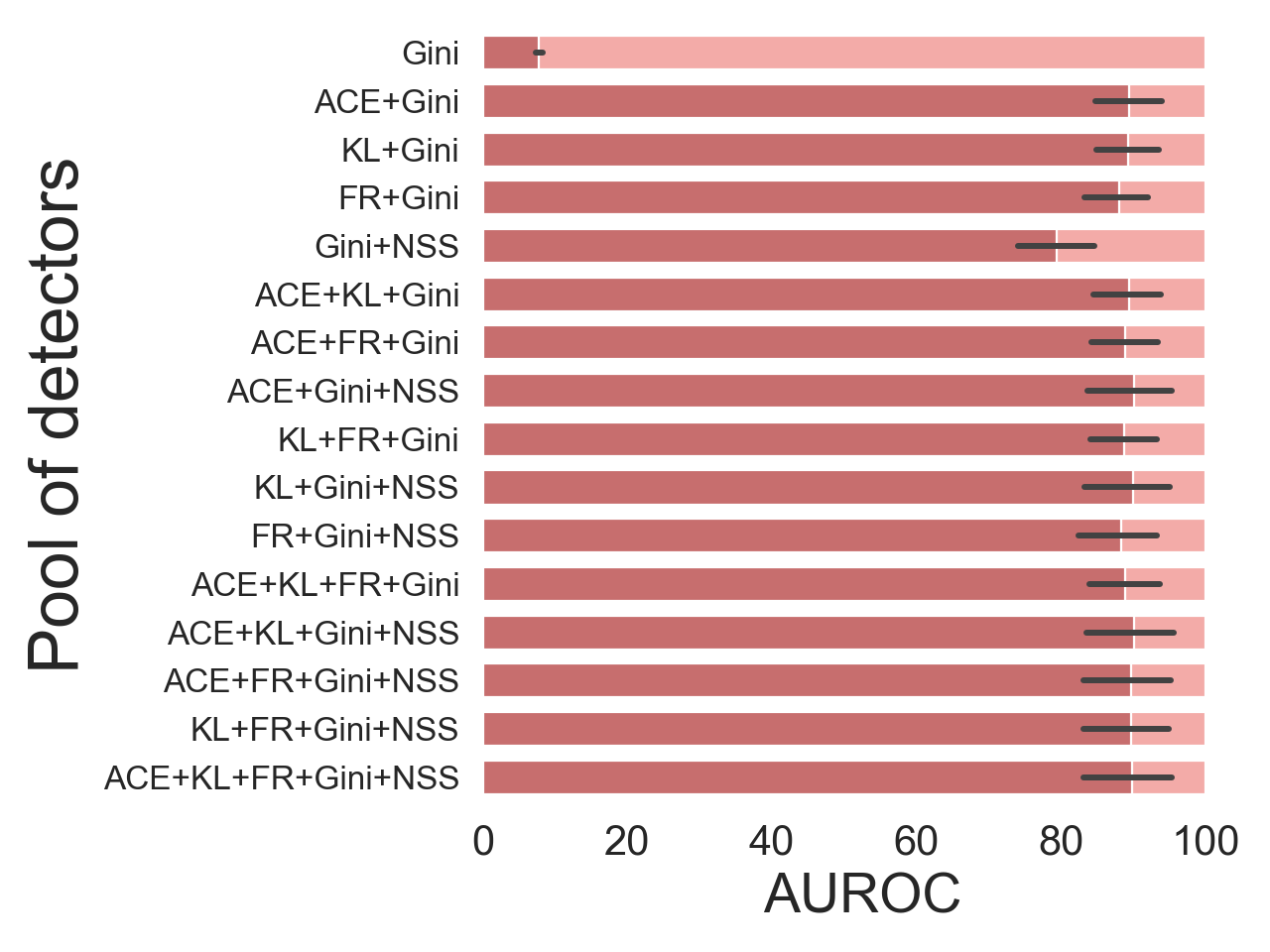}
		\vspace{-1.5\baselineskip}
		\caption{Attacks created with L$_\infty$}
  \label{fig:a_Giniinf}	
 \end{subfigure}
    \hfill
        	\centering
		\begin{subfigure}[b]{.25\columnwidth}
		\centering
		\includegraphics[width=\columnwidth]{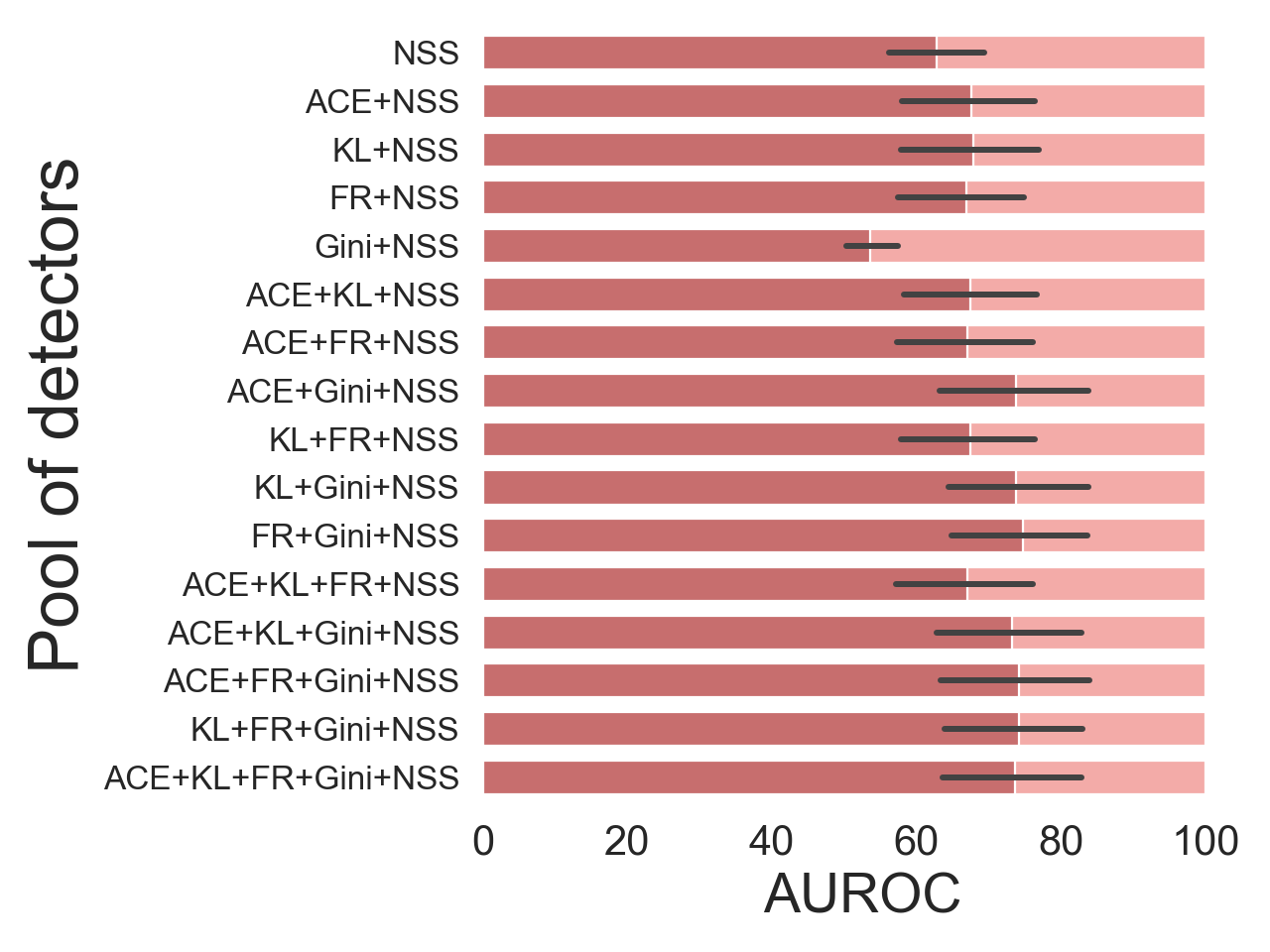}
		\vspace{-1.5\baselineskip}
		\caption{Attacks created with L$_1$}
    \label{fig:a_NSS1}

	\end{subfigure}
	    \hfill
		\begin{subfigure}[b]{.25\columnwidth}
		\centering
		\includegraphics[width=\columnwidth]{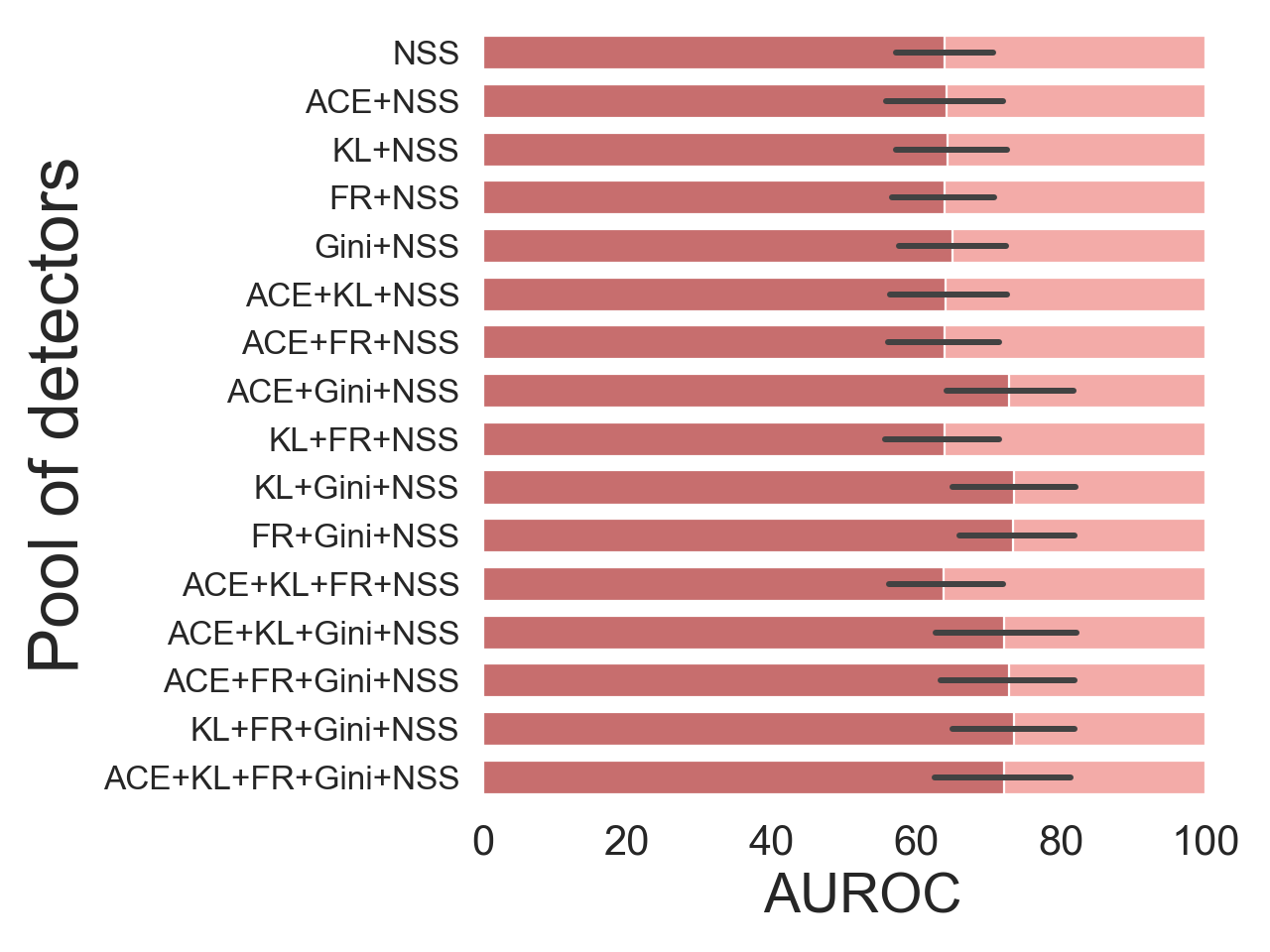}
		\vspace{-1.5\baselineskip}
		\caption{Attacks created with L$_2$}
      \label{fig:a_NSS2}

	\end{subfigure}
	 \hfill
	\begin{subfigure}[b]{.25\columnwidth}
	    \centering
	    \includegraphics[width=\columnwidth]{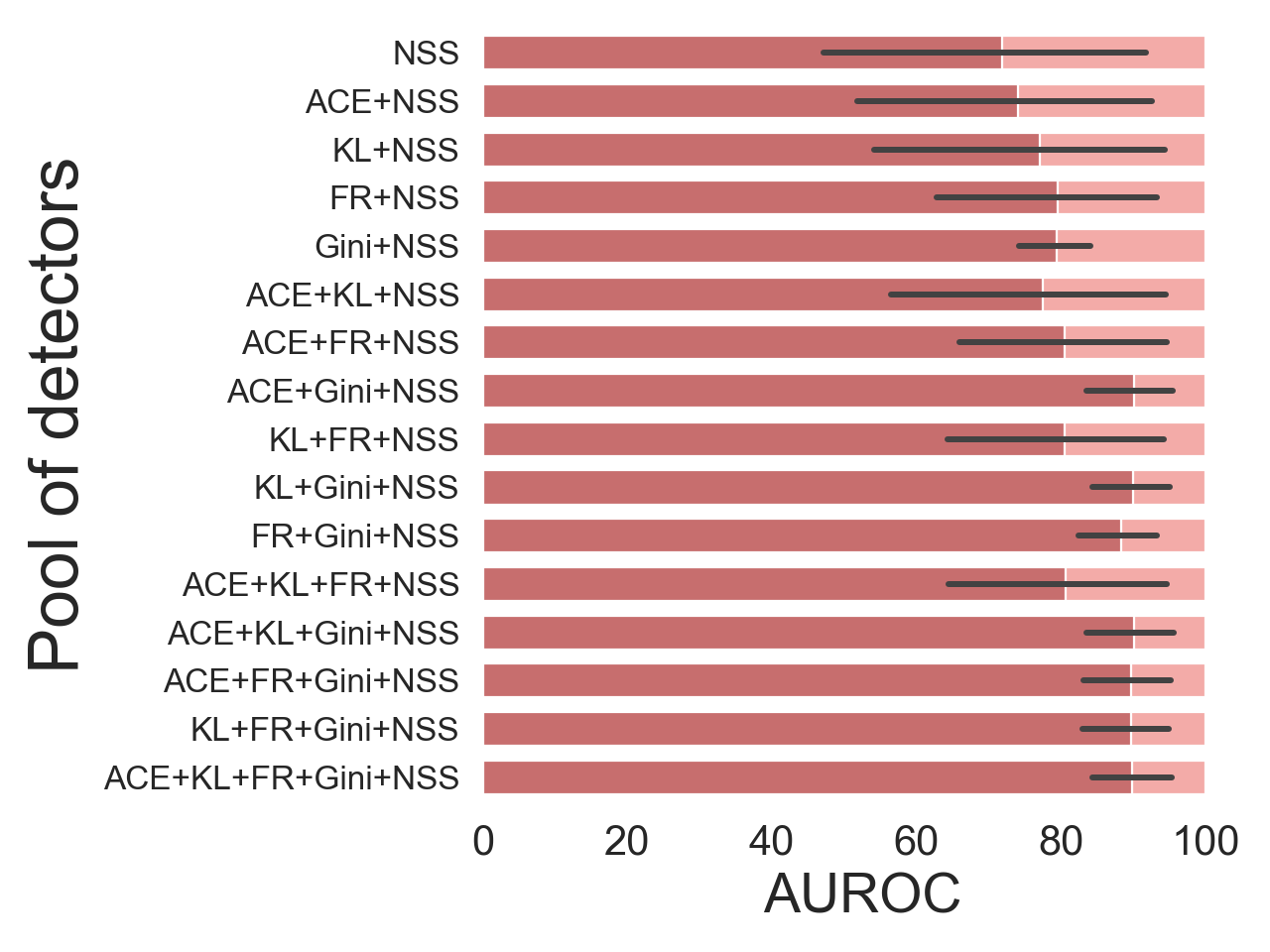}
		\vspace{-1.5\baselineskip}
		\caption{Attacks created with L$_\infty$}
    \label{fig:a_NSSinf}	
 \end{subfigure}
	\caption{Ablation study: {\mead} setting.}
	\label{fig:ablation2}
\end{figure*}

As a final set of experiments, we examine how the combinations of detectors affect the aggregator's performance. Differently from the ablation study in~\cref{app:res_th_framework}, the set of attacks is always fixed as in~\cref{tab:attacks}. Therefore, we consider all the possible combinations ($2^5=32$) of the detectors discussed throughout this paper. In~\cref{fig:ablation2}, the set of detectors to aggregate is shown on the vertical axis, and the average AUROC is shown on the horizontal axis. The results are computed over all the attacks grouped w.r.t. the L$_p$ norm used. \Cref {fig:ablation2} shows that the Gini detector is a necessary but not sufficient detector to include in the pool:  the combination of the Gini detector with even just one other detector yielded unexpected results, despite its low effectiveness when considered in isolation (cf.~\cref{fig:a_Gini1,fig:a_Gini2,fig:a_Giniinf}).
This suggests that having \textit{mutually-exclusive detector} (i.e., a detector that is designed to recognize a specific typology of attack) in our aggregator would be actually beneficial for the detection task. Note that, in this sense, parallelism can be created with the work in~\cite{AbdelnabiF21} where the authors propose a new training approach based on \textit{adversarially-disjointed} models with minimal transferability of the attacks. 

\subsection{Evaluation of the Proposed Solution in the Single-Armed Setting}
\label{sec:res_non_simult}
In these experiments, we move from the multi-armed adversarial attack scenario if~\cite{GranesePRMP2022ECMLPKDD}, to the single-armed setting where the different detectors are aggregated to detect one single attack at a time. Note that, in this case, we do not impose any constraints on the relationship between detectors and attacks. As we follow what is usually done in the literature. As a result, we may come across scenarios where none of the detectors perform optimally in recognizing certain attacks. As well as it is possible to encounter situations where multiple detectors are equally optimal for specific types of attacks.
We refer to~\cref{tab:single_setting} for the complete set of results. 
Crucially, these experiments show that aggregation of the detectors can also improve the performance of detecting specific attacks. In particular, we would like to draw attention to the fact that we outperform NSS in the vast majority of the cases. Moreover, we achieve a maximum gain of 82.8 percentage points in terms of {\auc} (cf. SA attack) and 97.6 percentage points in terms of {\fpr} (cf. FGSM with $\varepsilon =0.5$ attack). On the other side, the competitor outperforms our proposed method only in a few cases, achieving a maximum gain of 5.9 percentage points in terms of {\auc} and 27.4 percentage points in terms of {\fpr} (cf. FGSM with $\varepsilon$=0.03125 attack in both the cases), and these gains are much lower than those obtained by the proposed method.
\begin{table*}[!htbp]
\centering
\caption{The proposed method and NSS in the non-simultaneous setting. The column names ACE, KL, FR, and Gini denote the loss function used to craft the attacks. HOP, DeepFool, CW2, and STA attacks have already been considered individually in ~\cref{tab:final_table}.}
\ra{1.3}
\resizebox{\columnwidth}{!}{%
\begin{tabular}{@{}r|bcbcbcb@{}}
\toprule
& \multicolumn{7}{c}{CIFAR10}\\
\cline{2-8}
& \multicolumn{7}{c}{Ours {\auc}~({\fpr})~~--~~NSS {\auc}~({\fpr})}\\
\cline{2-8}
& \multicolumn{1}{c}{ACE} & \phantom{abc} & \multicolumn{1}{c}{KL} & \phantom{abc} & \multicolumn{1}{c}{FR} & \phantom{abc} & \multicolumn{1}{c}{Gini}\\
\cline{2-2}\cline{4-4}\cline{6-6}\cline{8-8}
\underline{PGD1}\\
 $\varepsilon =$ 5 & 
\textbf{ 66.2 (83.5)}  -- 49.9 (93.5)  &&
\textbf{ 64.2 (85.6)} -- 49.6 (93.0)  &&
\textbf{ 63.1 (86.8)} -- 49.9 (93.3)  &&
\textbf{ 80.7 (58.4)} -- 50.3 (93.2) \\
 $\varepsilon =$ 10 &
\textbf{ 62.6 (87.5) }-- 56.9 (88.4) && 
\textbf{ 62.2} (88.4) -- 56.6 \textbf{(88.3}) &&
\textbf{ 63.1 (86.5) }-- 57.0 (88.1) &&
\textbf{ 86.9 (46.0)} -- 57.1 (88.8) \\
 $\varepsilon =$ 15 &
 \textbf{74.2 (81.3)} -- 63.1 (83.0) &&
\textbf{ 75.2 (80.5)} -- 62.8 (83.1) &&
 \textbf{75.3 (79.5)} -- 63.2 (82.5) &&
\textbf{ 90.0 (37.0)}  -- 63.5 (84.0) \\
 $\varepsilon =$ 20 & 
 \textbf{86.8 (65.3)} -- 68.5 (77.1)  &&
 \textbf{87.5 (63.1)} -- 68.1 (77.3)  &&
 \textbf{86.9 (63.2)} -- 68.7 (76.4)  &&
 \textbf{91.7 (31.1)}  -- 69.9 (77.6) \\
 $\varepsilon =$ 25 &
 \textbf{93.9 (38.4) }-- 73.1 (71.1)  &&
 \textbf{94.3 (36.1)} -- 72.7 (71.8)  &&
 \textbf{93.7 (41.0)} -- 73.4 (70.9)  &&
 \textbf{92.3 (28.9)} -- 75.0 (71.4)  \\
 $\varepsilon =$ 30 & 
 \textbf{97.1 (12.2)} -- 77.1 (64.5)  &&
 \textbf{97.2 (12.6)} -- 76.8 (65.1)  &&
 \textbf{96.8 (15.9)} -- 77.4 (65.2)  &&
 \textbf{92.6 (27.9)} -- 78.6 (67.3)  \\
 $\varepsilon =$ 40 &
 \textbf{98.9 (1.0)} -- 83.5 (52.7) && 
 \textbf{99.0 (1.0)} -- 83.3 (53.5) && 
 \textbf{98.8 (1.0)} -- 83.6 (52.7) && 
 \textbf{92.7 (27.4)} -- 80.1 (64.9) \\
 \midrule
 \underline{PGD2}\\                                                 
 $\varepsilon =$ .125 & 
 \textbf{67.9 (81.0)} -- 49.5 (93.8) &&
 \textbf{65.4 (84.1}) -- 49.1 (93.5) &&
\textbf{ 63.9 (86.5)} -- 49.6 (93.5) &&
 \textbf{80.6 (58.4)} -- 49.5 (94.3) \\
 $\varepsilon =$ .25 &
 \textbf{62.3 (87.4)} -- 55.9 (89.1)  &&
 \textbf{62.1 (88.0)} -- 55.6 (89.2)  &&
 \textbf{62.6 (87.6)} -- 55.8 (89.4)  &&
 \textbf{86.7 (46.4)} -- 55.9 (89.8) \\
 $\varepsilon =$ .3125  &
 \textbf{66.4 (86.2) }-- 59.4 (86.5) &&
 \textbf{67.0 (85.8)} -- 59.0 (86.6) &&
 \textbf{67.8 (84.8)} -- 59.3 (86.6) &&
 \textbf{88.4 (42.2)} -- 59.3 (87.7) \\
 $\varepsilon =$ .5 &
 \textbf{86.4 (66.9)} -- 68.3 (77.4) &&
 \textbf{87.2 (64.4)} -- 68.0 (77.4) &&
 \textbf{86.7 (63.9)} -- 68.4 (77.2) && 
 \textbf{91.4 (31.4)} -- 69.0 (78.7) \\
 $\varepsilon =$ 1 & 
 \textbf{98.9 (0.9)} -- 84.4 (50.6) &&
 \textbf{99.0 (0.9)} -- 84.3 (50.5) && 
 \textbf{98.8 (0.9) }-- 84.7 (50.7) && 
 \textbf{92.5 (27.2)} -- 79.3 (66.8)\\
 $\varepsilon =$ 1.5 & 
 \textbf{99.2 (0.9)} -- 92.8 (28.7) &&
 \textbf{99.3 (0.9)} -- 92.7 (28.9) && 
 \textbf{99.3 (0.7)} -- 93.0 (27.3) && 
 \textbf{92.5 (27.2)} -- 79.5 (66.5) \\
 $\varepsilon =$ 2 &
 \textbf{99.3 (0.8)} -- 96.8 (13.9) &&
 \textbf{99.3 (0.8)} -- 96.9 (13.1) &&
 \textbf{99.3 (0.9)} -- 95.9 (17.2) &&
 \textbf{92.5 (27.2)} -- 79.5 (66.5) \\
 \midrule
 \underline{PGDi}\\               
 $\varepsilon =$ .03125 &
 \textbf{99.1 (0.9)} -- 92.3 (31.0) &&
 \textbf{99.1 (0.9)}  -- 92.1 (31.9) &&
 \textbf{99.0 (0.9)} -- 92.2 (30.7)  &&
 \textbf{94.8 (21.5)} -- 89.0 (44.0) \\
 $\varepsilon =$ .0625 &
 \textbf{99.3 (0.8)} -- 99.1 (3.3) &&
 \textbf{99.3 (0.8)} -- 99.1 (3.3) &&
 \textbf{99.3 (0.8)} -- 99.1 (3.6) &&
 97.4 \textbf{(8.0) }-- \textbf{98.1} (8.1)\\
 $\varepsilon =$ .125 &
 99.3 (0.7) -- \textbf{99.7 (0.6)} &&
 99.3 (0.9) -- \textbf{99.7 (0.6)} &&
 99.3 (0.8)  -- \textbf{99.6 (0.6)} && 
 97.3 (7.2) -- \textbf{99.6 (0.6)}\\
 $\varepsilon =$ .25 &
 99.3 (0.7) -- \textbf{99.7 (0.6)} && 
 99.3 (0.9) -- \textbf{99.7 (0.6)} && 
 99.3 (0.8)  -- \textbf{99.7 (0.6)} && 
 97.1 (7.3) -- \textbf{99.6 (0.6)}\\
 $\varepsilon =$ .3125 &
 99.3 (0.9) -- \textbf{99.7 (0.6)} &&
 99.3 (0.8) -- \textbf{99.7 (0.6)} &&
 99.3 (0.8) -- \textbf{99.7 (0.6)} &&
 97.1 (7.4) -- \textbf{99.7 (0.6)}\\
 $\varepsilon =$ .5 &
 99.3 (0.8) -- \textbf{99.7 (0.6)} &&
 99.3 (0.8) -- \textbf{99.7 (0.6)} &&
 99.3 (0.8)  -- \textbf{99.7 (0.6)} &&
 97.1 (7.3) -- \textbf{99.6 (0.6)}\\
 \midrule
 \underline{FGSM}\\
 $\varepsilon =$ .03125 &
 89.2 (47.6) -- \textbf{94.1 (26.7)} &&
 91.3 (40.6) -- \textbf{94.0 (27.0)} &&
 92.6 (34.1) -- \textbf{96.8 (15.0)} &&
 90.7 (42.7) -- \textbf{96.6 (15.3) }      \\
 $\varepsilon =$ .0625 &
 96.4 (18.5) -- \textbf{99.4 (1.3)} &&
 96.2 (18.8) -- \textbf{99.4 (1.4)} && 
 97.6 (10.3) -- \textbf{99.6 (0.6)} &&
 97.4 (11.9) -- \textbf{99.6 (0.6)} \\
 $\varepsilon =$ .125 &
 99.3 (3.4) -- \textbf{99.7 (0.6)} &&
 99.1 (4.3) -- \textbf{99.7 (0.6)} &&
 99.3 (2.5) -- \textbf{99.5 (0.6)} &&
 99.3 (2.4) -- \textbf{99.5 (0.6)}\\
 $\varepsilon =$ .25 &
 \textbf{99.8 (0.6)} -- 99.7 \textbf{(0.6)} &&
 \textbf{99.7 }(0.8) -- \textbf{99.7 (0.6)} &&
 \textbf{99.6} (1.1) -- 97.9 \textbf{(0.6)} &&
\textbf{ 99.6 }(1.1)  -- 97.7 \textbf{(0.6)} \\
 $\varepsilon =$ .3125 &
 \textbf{99.7 }(0.9) -- \textbf{99.7 (0.6)} &&
 \textbf{99.7 }(0.9) -- \textbf{99.7 (0.6)} &&
 \textbf{99.5} (1.5) -- 95.8 \textbf{(0.6) }&& 
 \textbf{99.5 }(1.5) -- 95.6 \textbf{(0.6)}\\
 $\varepsilon =$ .5 & 
 99.0 (4.9) -- \textbf{99.7 (0.6)} && 
 99.2 (2.7) -- \textbf{99.7 (0.6)} &&
 \textbf{99.2 (2.4)} -- 84.9 (100.0) &&
 \textbf{99.2 (2.4) }-- 84.8 (100.0) \\
 \midrule
 \underline{BIM}\\
 $\varepsilon =$ .03125 & 
 \textbf{98.3 (4.6)} -- 90.3 (37.7) &&
 \textbf{98.3 (4.4)} -- 90.2 (38.1) && 
 \textbf{97.8 (7.1)} -- 90.5 (37.0) &&
 \textbf{92.2 (32.6)} -- 88.2 (45.1) \\
 $\varepsilon =$ .0625 &
\textbf{ 99.4 (0.8)} -- 98.2 (7.5) &&
 \textbf{99.3 (0.9)} -- 98.2 (7.5) &&
 \textbf{99.4 (0.8) }-- 98.3 (7.3) &&
 96.6 (13.1) --\textbf{ 97.3 (12.9)}       \\
 $\varepsilon =$ .125 &
 99.3 (0.9) -- \textbf{99.6 (0.7)} &&
 99.3 (0.9) -- \textbf{99.7 (0.7)} && 
 99.3 (0.8) -- \textbf{99.6 (0.7)} && 
 97.8 (6.9) -- \textbf{99.3 (1.9)}\\
 $\varepsilon =$ .25 &
 99.3 (0.8) -- \textbf{99.7 (0.6)} &&
 99.3 (0.9) -- \textbf{99.7 (0.6)} &&
 99.3 (0.8) -- \textbf{99.7 (0.6)} &&
 97.4 (7.2) -- \textbf{99.6 (0.6)}\\
 $\varepsilon =$ .3125 &
 99.3 (0.9) -- \textbf{99.7 (0.6) }&&
 99.3 (0.8) -- \textbf{99.7 (0.6)} &&
 99.3 (0.9) -- \textbf{99.7 (0.6)} &&
 97.1 (7.4) -- \textbf{99.7 (0.6)}\\
 $\varepsilon =$ .5 &
 99.3 (0.8) -- \textbf{99.7 (0.6)} &&
 99.3 (0.8) -- \textbf{99.7 (0.6)} &&
 99.3 (0.8) -- \textbf{99.7 (0.6)} &&
 96.3 (7.3) -- \textbf{99.7 (0.6)}         \\
 \midrule
 \underline{SA} \\
 $\varepsilon =$ .125 &
 \textbf{91.2 (39.6)} -- 9.4 (99.9) &&
\textbf{ 91.2 (39.6)} -- 9.4 (99.9)  &&
\textbf{ 91.2 (39.6)} -- 9.4 (99.9) &&
\textbf{ 91.2 (39.6)} -- 9.4 (99.9)        \\
 \midrule
 \underline{CWi}\\
 $\varepsilon =$ .3125 & 
\textbf{ 80.7 (60.9) }-- 64.6 (89.8) &&
\textbf{ 80.7 (60.9)} -- 64.6 (89.8) &&
\textbf{ 80.7 (60.9)} -- 64.6 (89.8) &&
\textbf{ 80.7 (60.9)} -- 64.6 (89.8) \\
 \bottomrule
\end{tabular}
}
\label{tab:single_setting}
\end{table*}

\subsection{Evaluation of the Proposed Solution in the Setting of~\cite{GranesePRMP2022ECMLPKDD} and Adaptive Attacks}
\label{app:adaptive}
\begin{table*}[htbp!]
\centering
\caption{The proposed method against the adaptive-attacks under {\mead}. In the following setting, we attack each detector and the classifier once at a time. $\alpha$ is the parameter to control the losses.}
\ra{1.3}
\resizebox{\columnwidth}{!}{%
\begin{tabular}{@{}r|bbcbbcbbcbbcbb@{}}\toprule
& \multicolumn{14}{c}{CIFAR10}  \\
\cmidrule{2-15}
& \multicolumn{2}{c}{$\alpha=0$} & \phantom{abc}& \multicolumn{2}{c}{$\alpha=.1$} &
  \phantom{abc} & \multicolumn{2}{c}{ $\alpha=1$} & \phantom{abc}& \multicolumn{2}{c}{$\alpha=5$} & \phantom{abc} &  \multicolumn{2}{c}{$\alpha=10$}\\ \cmidrule{2-3}
\cmidrule{5-6} \cmidrule{8-9} \cmidrule{11-12} \cmidrule{14-15}
  & \auc & \fpr  && \auc & \fpr && \auc & \fpr && \auc & \fpr && \auc & \fpr\\ 
 \midrule

\textbf{Norm L$_1$}\\

\underline{PGD1$^\star$}\\
$\varepsilon=5$ & 
\textbf{62.1} & \textbf{87.1} &&
61.3 & 88.6 && 
61.2 & 89.3 && 
63.1 & 89.2 && 
62.6 & 91.3
\\
$\varepsilon=10$ &
\textbf{56.8} & 90.6 &&
53.1 & 94.5 && 
54.4 & 93.9 && 
60.0 & 91.0 && 
60.6 & 91.9
\\
$\varepsilon=15$ & 
\textbf{69.3} & \textbf{84.4} &&
51.5 & 96.5 && 
54.7 & 94.6 && 
64.1 & 88.1 && 
65.7 & 87.7
\\
$\varepsilon=20$ & 
\textbf{78.7} & \textbf{73.1} &&
53.4 & 96.8 && 
55.9 & 94.9 && 
66.7 & 84.1 && 
69.4 & 82.7
\\
$\varepsilon=25$ & 
\textbf{87.1} & \textbf{50.8} &&
54.0 & 97.2 && 
56.7 & 94.6 && 
67.8 & 82.7 && 
71.1 & 79.0 
\\
$\varepsilon=30$ &
\textbf{90.3} & \textbf{35.4} &&
54.5 & 97.1 && 
56.6 & 94.4 && 
68.9 & 81.1 && 
71.9 & 78.4 
\\
$\varepsilon=40$ & 
\textbf{92.1} & \textbf{22.7} &&
54.4 & 97.0 && 
57.7 & 93.6 && 
69.4 & 79.7 && 
72.9 & 74.2 
\\
\midrule

\textbf{Norm L$_2$}\\

\underline{PGD2$^\star$}\\
$\varepsilon=0.125$ & 
\textbf{63.9} & \textbf{85.4} &&
61.4 & 88.0 && 
62.4 & 88.8 && 
63.7 & 88.5 && 
63.9 & 89.9
\\
$\varepsilon=0.25$ & 
\textbf{57.1} & \textbf{90.5} &&
52.9 & 94.2 && 
55.0 & 93.6 && 
60.6 & 89.7 && 
61.5 & 90.3 
\\
$\varepsilon=0.3125$ & 
\textbf{61.0} & \textbf{88.9} &&
51.6 & 95.7 && 
54.1 & 94.7 && 
62.2 & 87.8 && 
63.7 & 87.9 
\\
$\varepsilon=0.5$ & 
\textbf{79.4} & \textbf{73.2}&&
52.8 & 96.8 && 
55.3 & 94.3 && 
66.2 & 84.6 && 
68.8 & 81.5 
\\
$\varepsilon=1$ & 
\textbf{91.4} & \textbf{26.4}&&
52.7 & 96.8 && 
57.3 & 93.4 &&
69.0 & 78.3 && 
72.1 & 74.4 
\\
$\varepsilon=1.5$ & 
\textbf{91.9} & \textbf{24.2} &&
53.9 & 96.1 && 
57.9 & 91.4 && 
70.5 & 73.7 && 
74.1 & 68.1 
\\
$\varepsilon=2$ & 
\textbf{91.9} & \textbf{24.1} &&
54.6 & 94.6 && 
59.3 & 88.5 && 
72.3 & 67.8 && 
75.6 & 62.7 
\\



\midrule

\textbf{Norm L$_\infty$}\\
\underline{PGDi$^\star$, FGSM$^\star$, BIM$^\star$}\\\
$\varepsilon=0.03125$ & 
\textbf{82.3} & \textbf{59.7}&&
45.3 & 96.2 && 
46.0 & 96.4 && 
54.5 & 91.4 && 
57.4 & 89.3 
\\
$\varepsilon=0.0625$ & 
\textbf{92.0} & \textbf{29.6} &&
44.3 & 96.2 && 
49.8 & 93.8 && 
59.7 & 82.4 && 
64.3 & 76.4 
\\
$\varepsilon=0.5$ & 
\textbf{94.6} & \textbf{9.7} &&
62.1 & 81.3 && 
54.9 & 81.9 && 
66.1 & 60.8 && 
68.9 & 57.9 
\\

\underline{PGDi$^\star$, FGSM$^\star$, BIM$^\star$, SA}\\
$\varepsilon=0.125$ & 
\textbf{88.9} & \textbf{40.8} &&
48.6 & 90.7 &&
54.9 & 85.0 &&
61.9 & 73.1 &&
66.3 & 67.5 
\\

\underline{PGDi$^\star$, FGSM$^\star$, BIM$^\star$, CWi}\\
$\varepsilon=0.3125$ &
\textbf{80.0} & \textbf{61.1} &&
56.6 & 82.0 &&
56.3 & 79.6 &&
66.1 & 66.1 && 
69.2 & 64.4 
\\

\bottomrule
\end{tabular}
}
\label{tab:table_rebuttal_1}
\end{table*}
\begin{table*}[htbp!]
\centering
\caption{The proposed method against the adaptive-attacks under {\mead}. In the following setting, we attack all the detectors and the classifier together at the time. $\alpha$ is the parameter to control the losses.}
\ra{1.3}
\resizebox{\columnwidth}{!}{%
\begin{tabular}{@{}r|bbcbbcbbcbbcbb@{}}\toprule
& \multicolumn{14}{c}{CIFAR10}  \\
\cmidrule{2-15}
& \multicolumn{2}{c}{$\alpha=0$} & \phantom{abc}& \multicolumn{2}{c}{$\alpha=.1$} &
  \phantom{abc} & \multicolumn{2}{c}{ $\alpha=1$} & \phantom{abc}& \multicolumn{2}{c}{$\alpha=5$} & \phantom{abc} &  \multicolumn{2}{c}{$\alpha=10$}\\ \cmidrule{2-3}
\cmidrule{5-6} \cmidrule{8-9} \cmidrule{11-12} \cmidrule{14-15}
  & \auc & \fpr  && \auc & \fpr && \auc & \fpr && \auc & \fpr && \auc & \fpr\\ 
 \midrule

\textbf{Norm L$_1$}\\

\underline{PGD1$^\star$}\\
$\varepsilon=5$ & 
\textbf{62.1} & \textbf{87.1} &&
61.2 & 90.4 && 
63.6 & 86.8 && 
65.8 & 83.9 && 
66.3 & 83.2
\\
$\varepsilon=10$ &
\textbf{56.8} & 90.6 &&
50.5 & 96.4 && 
55.9 & 91.6 && 
60.1 & 88.1 && 
61.1 & 87.2 
\\
$\varepsilon=15$ & 
\textbf{69.3} & \textbf{84.4} &&
47.3 & 97.6 && 
53.8 & 92.3 && 
62.0 & 84.9 && 
63.7 & 83.7
\\
$\varepsilon=20$ & 
\textbf{78.7} & \textbf{73.1} &&
47.1 & 97.9 && 
54.2 & 92.5 && 
64.2 & 82.8 && 
66.8 & 79.1
\\
$\varepsilon=25$ & 
\textbf{87.1} & \textbf{50.8} &&
47.8 & 98.0 && 
55.0 & 92.1 && 
66.5 & 79.5 && 
68.8 & 77.2
\\
$\varepsilon=30$ &
\textbf{90.3} & \textbf{35.4} &&
48.8 & 98.0 && 
55.8 & 91.3 && 
67.4 & 78.5 && 
70.4 & 75.0
\\
$\varepsilon=40$ & 
\textbf{92.1} & \textbf{22.7} &&
49.1 & 98.0 && 
56.8 & 90.5 && 
68.6 & 77.4 && 
72.5 & 71.6
\\
\midrule

\textbf{Norm L$_2$}\\

\underline{PGD2$^\star$}\\
$\varepsilon=0.125$ & 
\textbf{63.9} & \textbf{85.4} &&
62.4 & 88.5 && 
65.0 & 86.2 && 
66.9 & 82.9 && 
67.2 & 81.1
\\
$\varepsilon=0.25$ & 
\textbf{57.1} & \textbf{90.5} &&
51.2 & 96.0 && 
56.3 & 91.7 && 
60.6 & 87.2 && 
61.6 & 86.8
\\
$\varepsilon=0.3125$ & 
\textbf{61.0} & \textbf{88.9} &&
56.0 & 94.6 && 
57.9 & 93.6 && 
65.3 & 86.4 && 
66.7 & 86.6
\\
$\varepsilon=0.5$ & 
\textbf{79.4} & \textbf{73.2}&&
46.8 & 97.8 &&
54.6 & 91.3 && 
64.5 & 82.4 && 
66.8 & 79.5
\\
$\varepsilon=1$ & 
\textbf{91.4} & \textbf{26.4}&&
47.2 & 98.0 && 
57.8 & 89.4 && 
69.9 & 73.8 && 
73.1 & 71.7
\\
$\varepsilon=1.5$ & 
\textbf{91.9} & \textbf{24.2} &&
47.5 & 97.6 && 
59.9 & 86.9 && 
73.2 & 68.7 && 
76.5 & 63.1
\\
$\varepsilon=2$ & 
\textbf{91.9} & \textbf{24.1} &&
49.0 & 97.0 && 
62.8 & 83.3 && 
75.6 & 63.7 && 
79.5 & 56.6
\\



\midrule

\textbf{Norm L$_\infty$}\\
\underline{PGDi$^\star$, FGSM$^\star$, BIM$^\star$}\\\
$\varepsilon=0.03125$ & 
\textbf{82.3} & \textbf{59.7}&&
40.2 & 98.0 && 
47.6 & 95.5 && 
60.6 & 86.2 && 
65.0 & 81.8
\\
$\varepsilon=0.0625$ & 
\textbf{92.0} & \textbf{29.6} &&
37.9 & 98.0 && 
47.0 & 95.9 && 
61.9 & 82.1 && 
65.8 & 77.1
\\
$\varepsilon=0.25$ &
\textbf{95.9} & \textbf{8.8} &&
36.5 & 96.4 && 
47.4 & 97.7 && 
62.5 & 92.6 && 
65.4 & 90.8
\\
$\varepsilon=0.5$ & 
\textbf{94.6} & \textbf{9.7} &&
36.7 & 96.2 && 
46.0 & 97.7 && 
61.6 & 96.1 && 
66.0 & 94.8
\\

\underline{PGDi$^\star$, FGSM$^\star$, BIM$^\star$, SA}\\
$\varepsilon=0.125$ & 
\textbf{88.9} & \textbf{40.8} &&
38.5 & 95.9 &&
46.8 & 95.4 &&
60.1 & 85.0 &&
61.9 & 83.2\\

\underline{PGDi$^\star$, FGSM$^\star$, BIM$^\star$, CWi}\\
$\varepsilon=0.3125$ &
\textbf{80.0} & \textbf{61.1} &&
37.2 & 95.3 &&
46.7 & 97.4 &&
60.9 & 92.4 && 
64.1 & 90.1\\

\bottomrule
\end{tabular}
}
\label{tab:table_rebuttal_2}
\end{table*}

\begin{figure}
	\centering
		\begin{subfigure}[b]{0.49\columnwidth}
		\centering
		\includegraphics[width=\columnwidth]{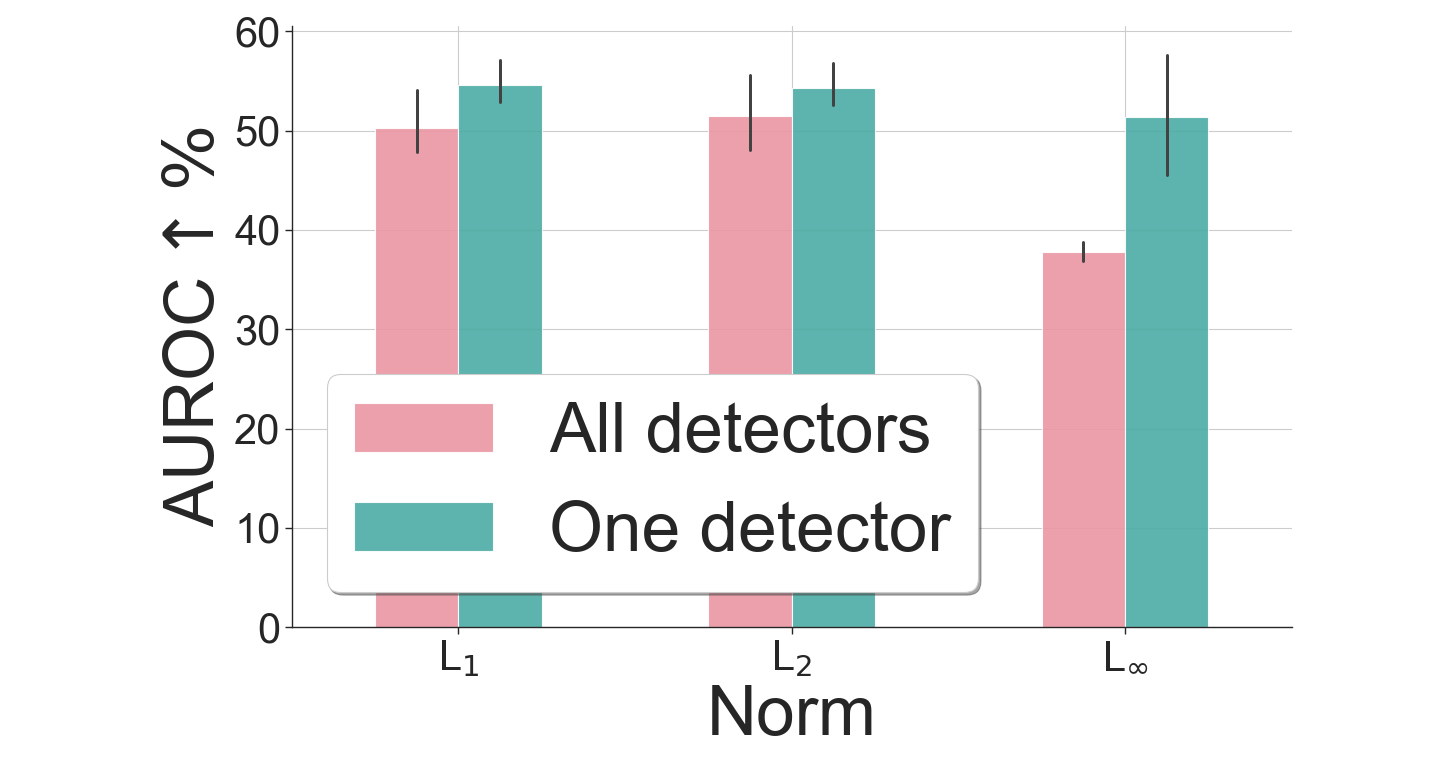}
		\vspace{-1.5\baselineskip}
		\caption{Analysis \auc}
		\label{fig:adaptive_auroc}
	\end{subfigure}
 \hfill
		\begin{subfigure}[b]{0.49\columnwidth}
		\centering
		\includegraphics[width=\columnwidth]{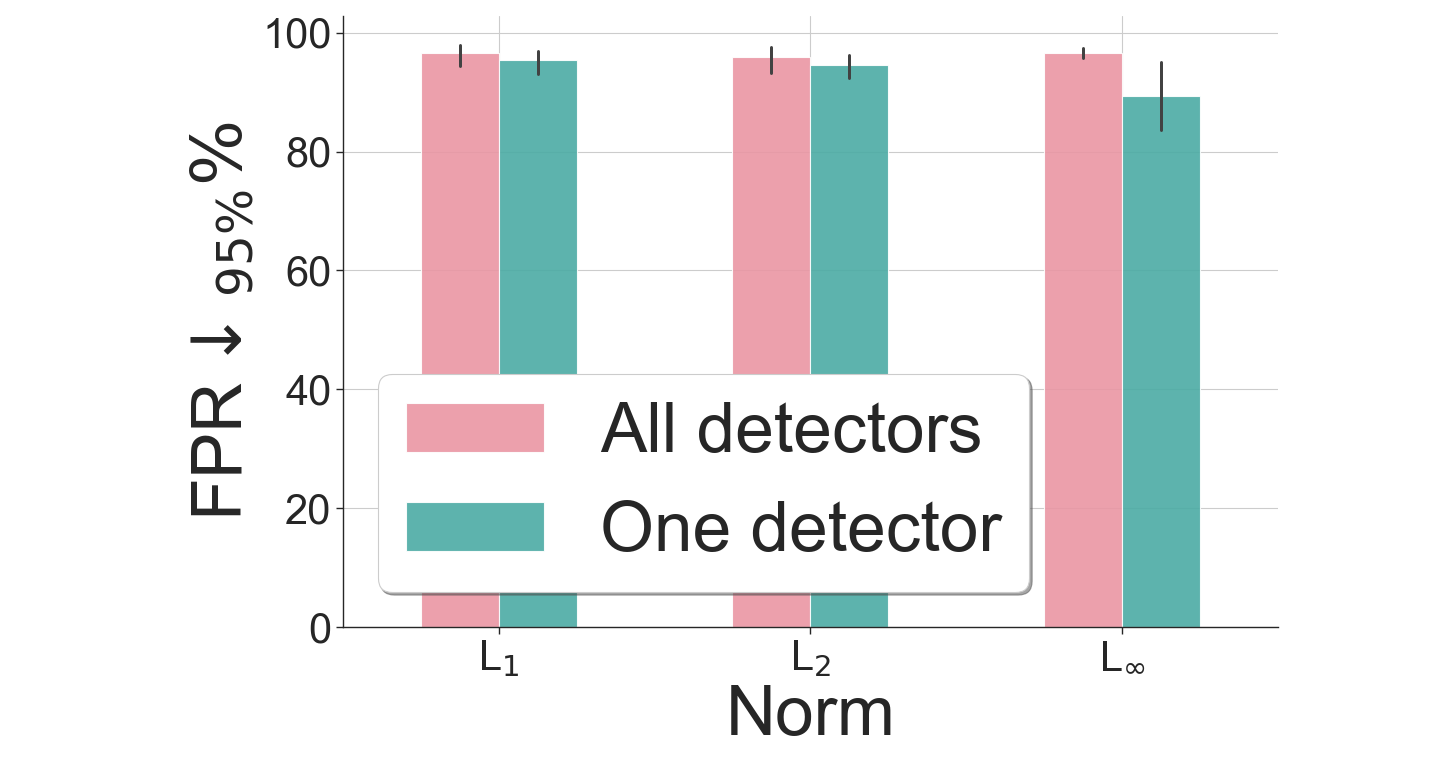}
		\vspace{-1.5\baselineskip}
		\caption{Analysis  \fpr}
		\label{fig:adaptive_fpr}
	\end{subfigure}
	\caption{Our method against the adaptive-attacks under {\mead}. We consider the worst case scenario in~\cref{tab:table_rebuttal_1,tab:table_rebuttal_1}, i.e., when $\alpha=0.1$.}
	\label{fig:adaptive}
\end{figure}
We present a new experimental setting to address the case in which also the detectors are attacked at the same time as the target classifier, taking the cue from~\cite{Bryniarski2021,Carlini017,TramerCBM20,CarliniMagnet}. 
It is important to note that, in the spirit of the {\mead} framework, we are not simply considering a scenario in which a \textit{single} adaptive attack is perpetrated on the classifier and detectors, but rather \underline{multiple} adaptive attacks are concurrently occurring. This scenario has not yet been considered in~\cite{GranesePRMP2022ECMLPKDD}, so we are the first to deal with such a setting. We extend
the framework to include two main cases: \textit{(i)} for attacks on the classifier and the single detectors individually; \textit{(ii)} for attacks on the classifier and all the detectors simultaneously.

The tables with the complete results are \cref{tab:table_rebuttal_1,tab:table_rebuttal_2}, where $\alpha$ is the coefficient that controls the gradient's speed of the attack against the detectors. We try many different values $\alpha=\left\{.1, 1, 5, 10\right\}$. The case where $\alpha$ is equal to 0 is added for completeness, and it corresponds to the case where only the target classifier is attacked.
We report in~\cref{fig:adaptive} the comparison of the results between case \textit{(i)} and case \textit{(ii)} on CIFAR10 and $\alpha=0.1$, as this corresponds to the case with the worst performances. As can be seen, the performances of our aggregator improve when the detectors are attacked singularly. This is particularly interesting for the setting we are dealing with. Indeed, our method is not a new supervised adversarial detection method but a framework to aggregate detectors, in this case, applied to the adversarial detection problem. Hence, it does not propose solving the problem of finding a new robust method for adaptive attacks but rather creating a mixture of experts based on the proposed sound mathematical framework. Thus, an attacker to successfully fool our method needs to have the \textit{complete access to all} the underlying detectors and also \textit{an up-to-the-date knowledge of the detectors employed} as the defender can always include a new detection mechanism to the pool of the detectors.

To give more insights on the proposed aggregator under this setting, we train a \textit{stronger} version of the four shallow detectors where the detectors at training time have seen the corresponding adaptive attacks generated through the PGD algorithm. We report the results in~\cref{tab:adaptive_strong} where we focus on the group of simultaneous attacks with L$_\infty$ norm and $\varepsilon=0.25$ as this represents the worst result of our method in~\cref{tab:table_rebuttal_2}.
If our method was only good as the best among the detectors, we should expect similar results in~\cref{tab:adaptive_strong}. In this case, the only solution would be to train a better detector. \textbf{However, the strength of the aggregator is not just mimicking the performance of its parts but rather creating a mixture of experts based on the proposed sound mathematical framework.} Therefore, we should expect better performances. Indeed, this consistently holds as the method performs much better than the best detector.
\begin{table}[!htbp]
\vspace{5mm}
\centering
\caption{Comparison between the proposed method and the single detectors (\textit{stronger} version) against the adaptive-attacks. Norm L$_\infty$ and $\varepsilon=0.25$ (i.e., attacks PGDi$^\star$, FGSM$^\star$, BIM$^\star$).}
\ra{1.3}
\resizebox{.5\columnwidth}{!}{%
\begin{tabular}{@{}r|b|b|b|b|b@{}}\toprule
CIFAR10 & \multicolumn{1}{c}{Ours} &
\multicolumn{1}{c}{ACE} & \multicolumn{1}{c}{KL}  & \multicolumn{1}{c}{FR} & \multicolumn{1}{c}{{Gini}} \\ \midrule
\auc & \textbf{54.6} & 35.7 & 30.6 & 26.3 & 36.2 \\
\fpr & \textbf{73.0} & 96.5 & 97.0 & 97.4 & 99.6 \\
\bottomrule
\end{tabular}
}
\label{tab:adaptive_strong}
\end{table}

\subsection{AutoAttack~\cite{Croce020a} in the Proposed Solution}
\label{app:autoattack}
We present an application of AutoAttack~\cite{Croce020a}, a state-of-the-art evaluation tool for robustness, redesigned for adversarial detection evaluation and adapted to our simultaneous attacks framework.
In its original version, AutoAttack evaluates the accuracy of robust classifiers. In so doing, ~\cite{Croce020a} proposes a multiple attacks framework to ensure that at least one attack succeeds in producing an adversarial example for each natural one. In their context, it does not matter which attack will succeed since any successful attack would undermine the accuracy of the target classifier in the same way. In our case, the number of different successful attacks for each natural sample will affect the detection quality since a detector is successful only if it can detect all of them. Because of the above-mentioned differences, it is impossible to deploy it directly in our framework without any modifications. A modified version of AutoAttack adapted to the evaluation of our proposed method, has been implemented, and the results are presented below. While AutoAttack suggests using different attack strategies, in our case, we combine different attack strategies matched with different losses to make the pool of attacks more strong and more diversified.
\begin{table}
\centering
\caption{The proposed method on AutoAttack ({\mead} setting). The attacks are APGD-CE, APGD-DLR, FAB, SA.}
\ra{1.3}
\resizebox{0.29\columnwidth}{!}{%
\begin{tabular}{@{}r|bb@{}}\toprule
& \multicolumn{2}{c}{CIFAR10}  \\
\cmidrule{2-3}
& \multicolumn{2}{c}{Ours} \\ \cmidrule{2-3}
  & \auc & \fpr  \\ 
 \midrule

\textbf{Norm L$_1$}\\

\underline{}\\
$\varepsilon=5$ & 
57.1 & 88.4
\\
$\varepsilon=10$ &
67.1 & 75.7
\\
$\varepsilon=15$ & 
72.2 & 66.7
\\
$\varepsilon=20$ & 
72.7 & 65.2
\\
$\varepsilon=25$ & 
72.8 & 65.6
\\
$\varepsilon=30$ &
73.4 & 64.0
\\
$\varepsilon=40$ & 
73.6 & 64.0
\\
\midrule

\textbf{Norm L$_2$}\\

\underline{}\\
$\varepsilon=0.125$ & 
67.4 & 81.0
\\
$\varepsilon=0.25$ & 
58.0 & 89.0
\\
$\varepsilon=0.3125$ & 
58.1 & 88.8
\\
$\varepsilon=0.5$ & 
69.4 & 74.7
\\
$\varepsilon=1$ & 
75.1 & 61.6
\\
$\varepsilon=1.5$ & 
76.1 & 60.7
\\
$\varepsilon=2$ & 
76.1 & 60.5
\\

\midrule

\textbf{Norm L$_\infty$}\\
\underline{}\\\
$\varepsilon=0.03125$ & 
75.7 & 61.0
\\
$\varepsilon=0.0625$ & 
76.0 & 60.7
\\
$\varepsilon=0.125$ & 
76.8 & 60.3
\\
$\varepsilon=0.25$ &
76.8 & 60.0
\\
$\varepsilon=0.3125$ &
78.6 & 57.6
\\
$\varepsilon=0.5$ & 
76.1 & 60.3
\\
\bottomrule
\end{tabular}
}
\label{tab:autoattack}
\end{table}

\vfill

\end{document}